\documentclass[a4paper, 10 pt, conference]{ieeeconf}
\IEEEoverridecommandlockouts  

\usepackage{cite}
\usepackage{algorithmic}
\usepackage{textcomp}

\usepackage{mathtools}
\usepackage{amsmath}
\usepackage{hyperref}
\usepackage{graphicx}
\usepackage{subfigure}
\usepackage{amsfonts,amssymb} 
\usepackage{caption} 
\usepackage{subfiles} 

\usepackage{color,soul} 
\renewcommand\hl[1]{#1} 

\title{\LARGE \bf
Controlling Pivoting Gait using Graph Model Predictive Control}

\author{Ang Zhang $^{1}$ ,  Keisuke Koyama$^{1}$, Weiwei Wan$^{1}$, Kensuke Harada$^{1}$
\thanks{$^{1}$Department of Engineering Science, Osaka University, Osaka, Japan
        {\tt\small zhang at hlab.sys.es.osaka-u.ac.jp}}%
}

\begin{document}

\maketitle

\begin{abstract}

\hl{Pivoting gait is efficient for manipulating a big and heavy object with relatively small manipulating force, in which a robot iteratively tilts the object, rotates it around the vertex, and then puts it down to the floor. 
However, pivoting gait can easily fail even with a small external disturbance due to its instability in nature. 
To cope with this problem, we propose a controller to robustly control the object motion during the pivoting gait by introducing two gait modes, i.e., one is the double-support mode, which can manipulate a relatively light object with faster speed, and the other is the quadruple-support mode, which can manipulate a relatively heavy object with lower speed. 
To control the pivoting gait, a graph model predictive control is applied taking into account of these two gait modes. 
By adaptively switching the gait mode according to the applied external disturbance, a robot can stably perform the pivoting gait even if the external disturbance is applied to the object. 
}
\end{abstract}


\section{Introduction}
\label{sec:introduction}
Although current robots mostly manipulate objects by once picking them up \cite{berscheid2020self,gualtieri2018pick,xu2020planning}, a pick-and-place manipulation is energy consuming and is not adequate for manipulating a large and heavy object since the grasped object has to be completely lifted up. 
On the other hand, a human can effectively select an adequate manipulation strategy taking into account both features of the task and physical parameters of the object. For example, when a human moves a large and heavy object like furniture, 
\hl{a human may once tilt the object, rotates it around an axis through the contact point between the object and the floor, }
and then puts it down to the floor. Such manipulation style of a large object is called the pivoting manipulation. 
Since the manipulated object is not completely constrained by the robot, the pivoting manipulation is classified as a style of the nonprehensile manipulation \cite{lynch1999dynamic}. So far, several different styles of nonprehensile manipulation has been proposed, such as rolling \cite{li1990motion}, pushing\cite{lynch1996stable, harada2007real, zhou2019pushing}, and pivoting\cite{doshi2019hybrid, lynch1999toppling, mason1993dynamic}. 


\begin{figure}
\centering     
\subfigure[Exchange of gait modes in the pivoting gait.]{\label{fig:intro1}\includegraphics[width=0.95\linewidth]{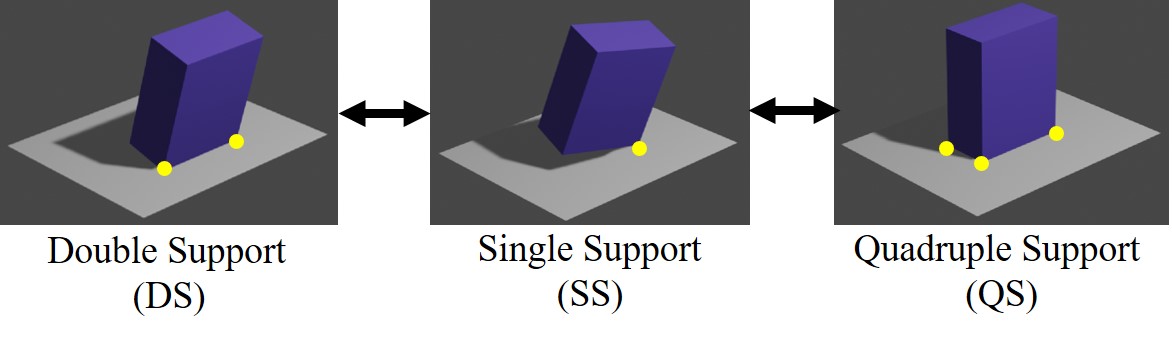}}
\subfigure[The pivoting gait runs in the DS mode in (1). A 1kg bottle is placed on the top of the object in (2). The pivoting gait in the DS mode fails after the placement in (3). A switch of gait mode to the QS mode enables the pivoting gait against the disturbance from a 2kg bottle in (4). ]{\label{fig:intro2}\includegraphics[width=0.85\linewidth]{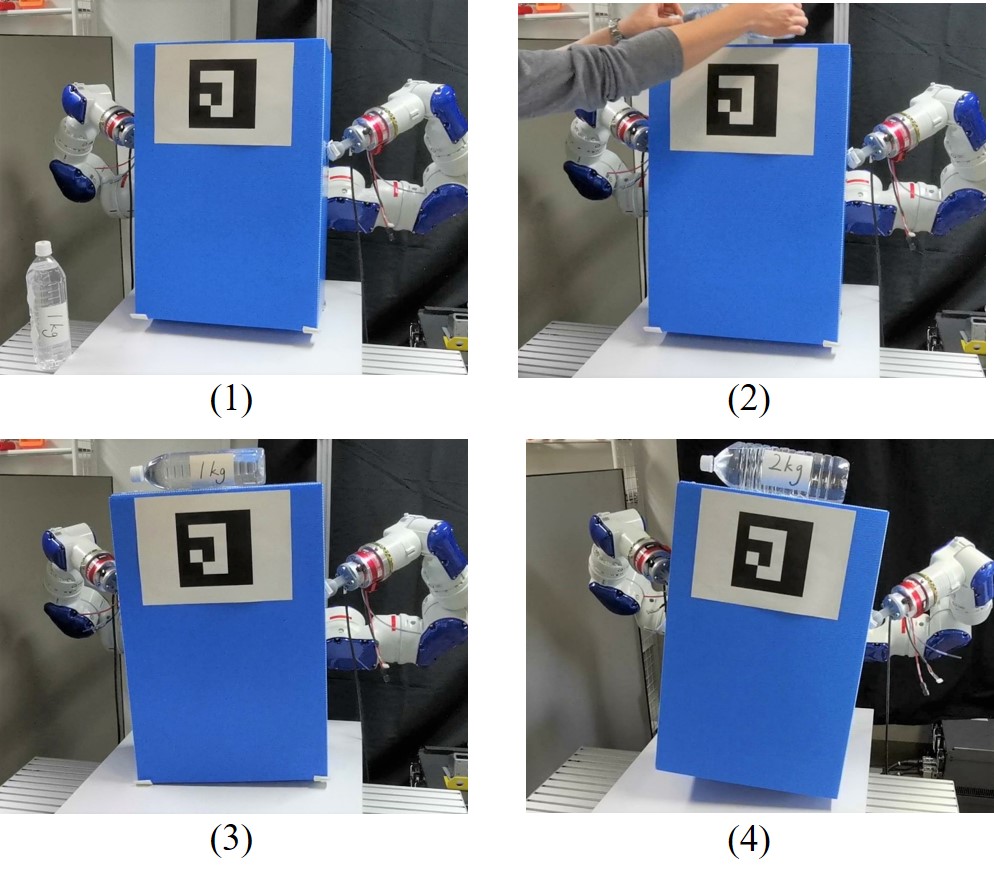}}
\caption{Gait modes and the robot performs the pivoting gait.}
\label{fig:intro}
\end{figure}

\hl{If pivoting manipulation involves the change of rotation vertices, it is called the pivoting gait}\cite{onyshko1980mathematical} due to the correspondence between the feet of a legged robot and the vertices of a manipulated object. Pivoting gait is effective especially when moving a large and heavy object since the object's weight is mostly supported by the contact with the floor\cite{yoshida2010planning}. 
\hl{Though there are some research works about feedback control of pivoting manipulation}\cite{yoshikawa2000dynamic, hou2019reorienting}
\hl{, there is no research about improving the robustness of pivoting motion by designing the gait mode.}

Similar to bipedal walking, pivoting gait includes single support (SS) where the object rotates around a supporting vertex \hl{which corresponds to the pivoting manipulation}, and double support (DS) where to change the supporting vertex, the object once contacts the floor with a supporting edge including two vertices. In addition, we define quadruple support (QS), where, to change its supporting vertex, the object once contacts the floor with a supporting surface including more than three vertices.
The way of changing supporting vertices results in different behaviors of pivoting gait and we propose two gait modes: the DS and the QS modes.
If a robot pivots an object with the QS mode, a robot can manipulate relatively heavy object with moving slowly. This is because the face contact is statically stable if the vertical projection of the center of gravity (CoG) is included in the face. On the other hand, if a robot pivots an object with the DS mode, a robot can manipulate an object with moving relatively fast. 
\hl{One difficulty of controlling pivoting gait lies in the change of contact modes.} 

\hl{Although adaptively using multiple gait modes can probably enable a robot to realize the robust pivoting gait}, there is a challenge on how to select an adequate gait mode. 

In this article, we propose a graph Model Predictive Controller (MPC) to control the pivoting gait. Two gait modes are designed and put in a graph. The graph selects the gait mode and outputs a reference trajectory of the object to MPC. Then, MPC tracks the reference and generates the desired position and force of the end effector. 
Besides, we use vision systems and force sensors to monitor the execution of the pivoting gait. If a disturbance is detected, the robot tries to change the gait mode by referring to the graph. The MPC compares the current state with the reference and realizes a feedback control.   
\hl{Simulation and experiments show the DS mode is fast while the QS mode is stable. In addition, the ability to switch gait modes improves the robustness of the control system and the robot can successfully achieve a stable pivoting gait under unexpected event like uncertainty in object mass and perturbation.}

The contributions of this work are:
\begin{itemize}
 \setlength{\parskip}{0cm} 
 \setlength{\itemsep}{0cm} 
  \item MPC is proposed to predict the future dynamic of the object during the pivoting gait.
  \item Two gait modes are designed to robustly perform the pivoting gait according to different purposes, i.e., the DS mode for fast \hl{walking} and the QS mode for walking stably. 
  \item A graph MPC is developed to select the sequence of modes. The merit of graph MPC is that by designing cyclic motions in a graph, we can select gait modes and realize the feedback control in real-time. 
  \item We strengthen the control system with state feedback by adding vision and force sensors to resist disturbances. 
\end{itemize}

In this paper, after introducing the related works in Section \ref{sec2:related}, we formulate the pivoting gait and predictive controller in Section \ref{sec3:formulation}. Section \ref{sec4:graph} provides the design of graph. 
In Section \ref{sec5:simulation}, simulation and experiments show that the proposed feedback control system is able to pivot an object to track a referenced trajectory while being free to switch gait modes, robust against external perturbations and uncertainty in object's weight. Section \ref{sec:6conclusion} summarizes the results and describes future work.

\section{Related works}
\label{sec2:related}

\subsection{Nonprehensile Manipulation}
Nonprehensile manipulation attracts more and more attentions nowadays as it enables manipulating the object with fewer degrees of freedom\cite{srinivasa2005control, amanhoud2019dynamical, ruggiero2018nonprehensile} where it includes the manipulation styles like throwing\cite{satici2016coordinate}, catching\cite{batz2010dynamic, cigliano2015robotic}, batting\cite{huang2012adding,mulling2013learning}, pushing\cite{harada2003whole,polverini2020multi}, sliding\cite{ramirez2011nonprehensile}, rolling\cite{donaire2016passivity} \hl{and pivoting}\cite{maeda2004motion,hou2018fast,cruciani2017hand}. In nonprehensile manipulation, the object is manipulated without satisfying the form or force closure \cite{serra2019control} which indicates that the object is sensitive to the environmental dynamics. 

Among several styles of nonprehensile manipulation, pivoting is a promising strategy to manipulate a heavy object. Aiyama et al.\cite{aiyama1993pivoting} originally proposed the pivoting gait in which they show how heavy objects can be effectively manipulated. 
Yoshida et al. \cite{yoshida2010pivoting} proposed a method for planning the pivoting gait by a humanoid robot. 
\hl{More recently,} Murooka et al. \cite{murooka2017feasibility} \hl{studied whole-body manipulation for humanoid robot to achieve pivoting task and explored simultaneous planning and estimation for humanoid pivoting tasks} \cite{murooka2018simultaneous}. Shi et al. \cite{shi2020aerial} \hl{proposed an aerial pivoting framework to pivot object by aerial robots.} Hou et al. \cite{hou2020robust} \hl{investigated the planar pivoting
problem, in which a pinched object is reoriented to a desired pose through swing motion.}

\hl{Pivoting gait is a process of repeating pivoting manipulation around the left and right rotation vertices. We design two gait modes for pivoting gait. To best of our knowledge, there is no research about feedback control of pivoting gait with the ability to switch gait modes.}

\subsection{Model Predictive Control}
Recently, MPC has been widely used in 
chemistry process \cite{testud1978model}, 
power system \cite{geyer2008hybrid}, 
solar technology \cite{camacho2008hybrid,garcia2009sliding} and flight control \cite{slegers2006nonlinear}. 
In robotics research, MPC is frequently used in bipedal walking \cite{faraji2014versatile} as MPC can effectively predict future dynamic behavior and cope with constraints on the state and the input \cite{wang2009model,zhang2019humanoid,neves2016energy}. 
Naveau et al. \cite{naveau2016reactive} modified the MPC schemes formulated as an optimization problem to include various gait modes.
Valenzuela et al. \cite{valenzuela2016mixed} introduce integer variables to represent the active contact modes and compute the mode sequences using mixed-integer nonlinear programming. 
Graph-based method also provides a way to plan a sequence of motion. Woodruff et al. \cite{woodruff2017planning} proposed a graph search algorithm to plan through a sequence of manipulation primitives describing different contact states. 
Tazaki et al. \cite{tazaki2006graph} proposed a graph-based MPC which draws possible modes in a graph and simulates bipedal walking at different speeds. Murooka et al. \cite{murooka2021humanoid} \hl{proposed loco-manipulation planning for humanoid robots based on graph search.}

While this research inspires from the similarity between the pivoting gait manipulation and humanoid's bipedal gait, they are essentially different due to the following two factors. Firstly, the formulation of pivoting gait is more complex than biped gait since a force-controlled dual-arm manipulator controls the contact mode of the grasped object. Secondly, the change of gait mode can be used to control the stability of the grasped object under gravity. This research applies the MPC to the pivoting gait where an impedance-controlled dual-arm manipulator is controlled to change the gait mode with predicting the object's future dynamics. Different from bipedal walking, we newly design a QS gait mode by changing the object's supporting vertices. The QS mode provides a stable motion of walking especially when perturbation happens.  

In this article, we investigate real-time control method for pivoting gait by considering two gait modes. A graph MPC is proposed to select the proper gait modes and realize feedback control by using the vision and force information. 

\section{Formulation of Model Predictive Control}
\label{sec3:formulation}

\begin{figure}
\centering
\includegraphics[width=0.35\textwidth]{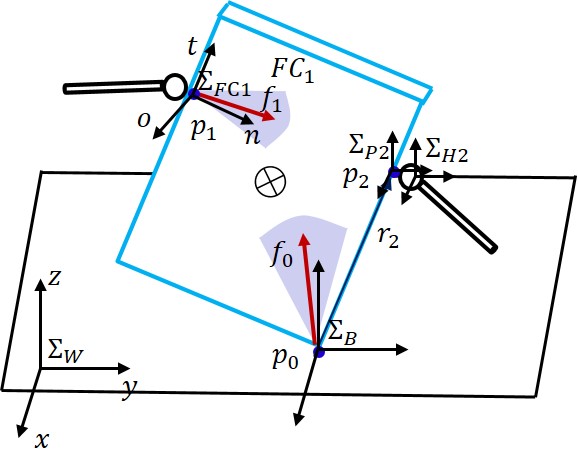}
\caption{Example of frames and contact forces}
\label{fig:friction}
\end{figure}

The target system of this work consists of two robot arms and a rigid object with a polygonal shape, see Fig. \ref{fig:friction}. 

\subsection{Nomenclature}
We describe here the notation used in this paper:
\begin{itemize}
 \setlength{\parskip}{0cm} 
 \setlength{\itemsep}{0cm} 
  \item $\Sigma_*$: Reference frames at $*=W, B, Hi, Pi$ and $FCi$ representing the coordinate frames attached to the ground, object, center of $i$-th spherical hand ($i=1,2$), 
  contact point on the object surface with the $i$-th hand, friction cone constraint at the contact point, respectively, where $n$ and $t, o$ denote the axes in the normal and two tangent directions, respectively. 
  \item $p_*$, $R_*$, $\Psi_*$, $\omega_*$ : The position vector, rotation matrix, Euler angle vector and angular velocity vector of $\Sigma_*$ with respect to $\Sigma_W$, respectively, where $*=B, Hi, Pi, FCi$ and $\Psi_* = [\phi_*, \psi_*, \theta_*]^ \mathrm{T}$.
  \item $X_k$, $Y_k$, $U_k$ : \hl{The states, outputs and free variables in MPC taking place in the prediction horizon} $n_p$ where $X_k = [x_k, \cdots, x_{k+n_p -1]}$ and the definitions of $Y_k$ and $U_k$ are similar with $X_k$. 
  \item $p_{BPi}$ : The vector directing from the origin of $\Sigma_B$ to the origin of $\Sigma_{Pi}$.
  \item $f_i$ : Force applied at the $i$-th contact point.
  \item $m_o$, ${\cal I}_o$ : object's mass and inertia matrix. 
  \item $g$ : Gravity force vector.
\end{itemize}

\noindent
For simplicity, we rewrite $p_{Pi}$ as $p_i$. We express the contact between the object and the ground during the single support phase by $i=0$. In addition, we assume the point contact with friction at each contact point when an object's vertex contacts the ground. 

\subsection{Kinematics}
\label{3.2formulation}

Pivoting gait is a manipulation style of an object iterating the steps where once raising the object up on a vertex, rotating it around that vertex and putting it down. 
Due to the contact with the ground at a fixed vertex, the object motion is constrained as 
\begin{equation}
SD_{B0} \left[\begin{matrix}
   \dot{p}_B \\
   \omega_B 
  \end{matrix}
  \right]    = S \left[\begin{matrix}
   \dot{p}_{0} \\
   \omega_{0} 
  \end{matrix}
  \right]  = o ,
  \label{eq:obj-sup}
\end{equation}
where $S$ is a selection matrix which selects the linear velocity and $D_{Bi}$ transforms the linear/angular velocity from $\Sigma_B$ to $\Sigma_W$,  
\begin{equation}
    S = [ I_3 \quad O_3],
    \label{selection matrix}
\end{equation}
\begin{equation}
    D_{Bi} =  \left[\begin{matrix}
   I_3 & -[(R_B  \prescript{B}{}{p_{BPi}})\times]\\
   O_3 & I_3
  \end{matrix}
  \right],
  \label{eq:hand-obj}
\end{equation}
where $[*\times]$ denotes the skew-symmetric matrix of a vector $*$ equivalent to the cross product operation, and $I_3$ and $O_3$ denote the $3 \times 3$ identity and zero matrices, respectively. 

Due to the contact with the $i$-th EEF ($i = 1, 2$), the object motion is constrained as  
\begin{equation}
    S D_{Bi}\left[\begin{matrix}
   \dot{p}_B \\
   \omega_B 
  \end{matrix}
  \right] = S D_{Hi}\left[\begin{matrix}
   \dot{p}_{Hi} \\
   \omega_{Hi} 
  \end{matrix}
  \right] = 
   \dot{p}_{i} .
  \label{finger vel}
\end{equation}

\begin{figure*}
\centering
\includegraphics[width=0.8\textwidth]{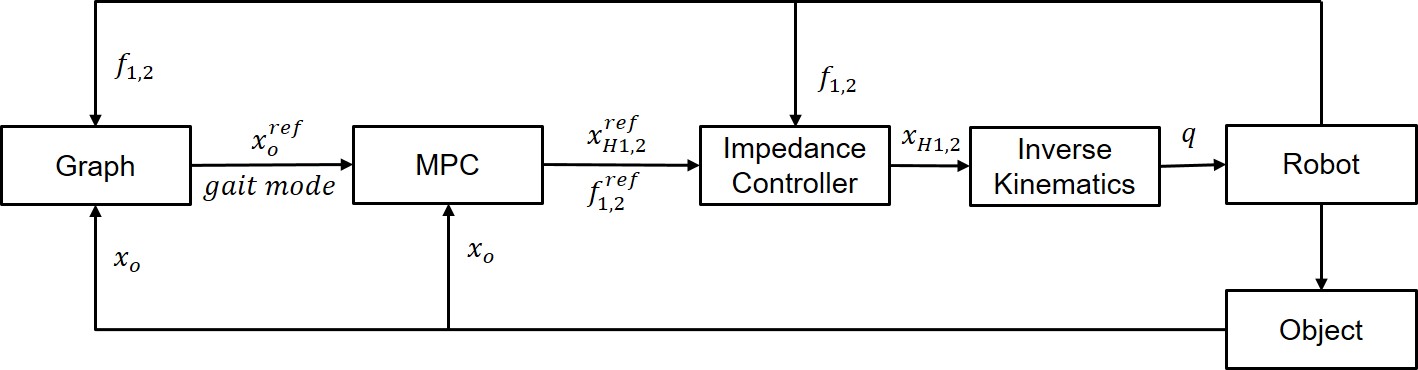}
\caption{This diagram describes the feedback loop used to control pivoting gait. According to $x_0$, the state of the object captured by the camera and $f_{1,2}$, the force of robot hands which are collected by force sensors, the graph selects a walking mode and outputs $x_o^{ref}$, a reference trajectory of the object. Then the MPC generates both $x_{H1,2}^{ref}$, desired hand trajectories and $f_{1,2}^{ref}$, desired forces. An impedance controller is implemented to handle the force control and outputs $x_{H1,2}$, hand trajectories. After doing inverse kinematics, the joint configuration $q$ is sent to the robot and then the robot manipulates the object.}
\label{fig:scheme}
\end{figure*}

Given a reference trajectory of the object's rotation, we apply the MPC to generate the motion of EEF and the target force applied by EEF, see Fig. \ref{fig:scheme}. 
With the knowledge of future dynamics, the MPC tries to find a way to pivot the object to the reference trajectory by solving an optimization problem.
In this research, we consider the dynamics related to the rotation of the object to formulate the controller since rotational dynamics are dominant to the gravitational stability of the object. 
In addition, we control the object motion in the single support phase where quadruple and double support phases are used to connect two single support phases. Hence, we formulate the state equation in the single support phase. 

Define the state vector as $x = [\Psi_B^T \quad \omega_B^T] ^ \mathrm{T}$. 
The relationship between angular velocity $\omega_B$ and the velocity of Euler angles $\dot{\Psi}_B$ is given by 
\begin{equation}
     \dot{\Psi}_B = W \omega_B ,
\end{equation}
where
\begin{equation}
 W = \left[\begin{matrix}
   1 & \sin\phi_B  \tan\theta_B & \cos\phi_B  \tan\theta_B \\
   0 & \cos\phi_B & -\sin\phi_B\\
   0 & \sin\phi_B \sec\theta_B & \cos\phi_B \sec\theta_B
  \end{matrix}
  \right].      
\end{equation}
From this equation, $k+1$-th step of the state can be predicted by
\begin{equation}
    \left[\begin{matrix}
    \Psi_B \\
    \omega_B
    \end{matrix}
    \right]_{k+1} =  \left[\begin{matrix}
   I_3 & W \\
   O_3 & I_3
  \end{matrix}
  \right] \left[\begin{matrix}
    \Psi_B \\
    \omega_B
    \end{matrix}
    \right]_k + \left[\begin{matrix}
    W T^2 /2  \\
   T
  \end{matrix}
  \right] [\dot{\omega}_B]_k ,
  \label{predict eq}
\end{equation}
where $T$ denotes the sampling time. 

\subsection{Dynamics}
\label{4model predictive control}

The object is accelerated by the force applied by two EEFs. Euler's equation of the object's rotational motion can be obtained as 
\begin{equation}
    r_1 \times f_1 + r_2 \times f_2 + r_{com} \times m_o g = {\cal I}_o\dot{\omega}_B + \omega_B \times {\cal I}_o \omega_B, 
    \label{rot dynamic}
\end{equation}
where $r_i=p_i-p_0$\ $(i=1,2)$, $r_{com}=p_{com}-p_0$ and $p_{com}$ denotes the position vector of object's CoG. Substituting \eqref{rot dynamic} into \eqref{predict eq} and define the force applied by EEF to be the input $u_k = [f_1 \quad f_2]^T_k$, we obtain the state equation as follows: 
\begin{equation}
    x_{k+1} =  A x_k + B u_k + D ,
    \label{eq:prediction}
\end{equation}
where $A, B, D$ are coefficient matrices defined as,
\begin{equation}
    A =  \left[\begin{matrix}
    I_3 & W  \\
    O_3 & I_3 
  \end{matrix}
  \right] ,
\end{equation}
\begin{equation}
    B = \left[\begin{matrix}
    {\cal I}_o^{-1} W  T^2 /2 [r_1 \times] & {\cal I}_o^{-1} W  T^2 /2 [r_2\times]  \\
    {\cal I}_o^{-1} T [r_2 \times] & {\cal I}_o^{-1} T [r_2 \times] 
  \end{matrix}
  \right] ,    
\end{equation}
\begin{equation}
   D = \left[\begin{matrix}
    {\cal I}_o^{-1} W  T^2 /2 [r_{com} \times] mg  \\
    {\cal I}_o^{-1} T [r_{com} \times] mg 
  \end{matrix}
  \right] .
\end{equation}
The coefficient matrices $B$ and $D$, which include $m_o$ and ${\cal I}_o$, reflect the dynamics of the object.
Here we note that, during the pivoting gait, the supporting vertex of the object expressed by $p_o$ changes between two consecutive single support phases. 


\hl{Considering the prediction horizon $n_p$, we define the states and the free variables taking place in the prediction horizon as}
 $X_k = [x_k, x_{k+1}, \cdots , x_{k + n_p - 1}]^T$ and $U_k = [u_k, u_{k+1}, \cdots , u_{k + n_p - 1}]^T$, respectively. 
According to \eqref{eq:prediction}, we have,
\begin{equation}
    X_{k+1} = A' X_k + B' U_k + D' ,
    \label{eq:predict along horizon}
\end{equation}
\hl{where the coefficient matrices are,}
\begin{equation}
    A' = [A, A^2, ... , A^{n_p}] ,
\end{equation}
\begin{equation}
    B' =  \left[\begin{matrix}
    A^0 B & \cdots & 0  \\
    \cdots & \ddots & \vdots \\
    A^{n_p - 1}B & \cdots & A^0 B
  \end{matrix}
  \right] ,
\end{equation}
\begin{equation}
    D' = \left[\begin{matrix}
    A^0 D & \cdots & 0  \\
    \cdots & \ddots & \vdots \\
    A^{n_p - 1}D & \cdots & A^0 D
  \end{matrix}
  \right] .
\end{equation}

\subsection{Output Equation}
The system outputs the velocity of EEF. 
From \eqref{eq:obj-sup} and \eqref{eq:hand-obj}, $\dot{p}_i$ can be obtained as
\begin{equation}
    \dot{p}_i = SD_{Bi}\left[\begin{matrix} [(R_B \prescript{B}{}{p_{BPi}})\times)] \\ I_3
    \end{matrix}\right]\omega_B .
    \label{eq:pi}
\end{equation}
From this equation, the velocity of EEF can be obtained as
\begin{equation}
   \dot{p}_{Hi} =S(SD_{Hi})^+\dot{p}_i + (I_3 - S(SD_{Hi})^+SD_{Hi})k_v ,
   \label{eq:phi}
\end{equation}
where $*^+$ is the pseudo-inverse of a matrix $*$ and $k_v$ denotes a 6-dimensional vector. 
By setting $y_k = [\dot{p}_{H1}^T \quad \dot{p}_{H2}^T]_k^T$ and $Y_k$ as the combination of $y_k$ over the prediction horizon , we can define the output equation by the form of 
\begin{equation}
    Y_k = C X_k + E,
    \label{eq:output}
\end{equation}
where the coefficient matrices $C$ and $E$ can be easily defined from \eqref{eq:pi} and \eqref{eq:phi}. 

\subsection{Cost Function}
\hl{The cost function used in MPC is defined as} 
\begin{equation}
   J_{mpc} =  \frac{\alpha }{2}\left \| X_{k+1} - X^{ref}_o \right\|^2 +
    \frac{\beta }{2}\left \| U_k  \right\|^2 ,
    \label{cost func}
\end{equation}
where $\alpha, \beta$ are the weights.
$X^{ref}_o$ \hl{is the reference trajectory of the object along the prediction horizon} and it is provided by the graph which we will discuss in Section \ref{sec4:graph}. The first and the second terms in \eqref{cost func} denote the amount of the state error and the amount of the input force of the EEF. 
\hl{Considering} \eqref{eq:predict along horizon}, the cost function can be formulated in the following form,
\begin{equation}
    \label{expand cost}
    J_{mpc} = \frac{\alpha }{2} \left \| A' X_k + B' U_k + D' - X^{ref}_o \right\|^2 + \frac{\beta }{2} \left \| U_k  \right\|^2  . 
\end{equation}

From this cost function, we can formulate the  quadratic programming(QP) problem as
\begin{equation}
   \mathop{\min}\limits_{U_k} \frac{1}{2} U_k^T Q U_k + r^T U_k +s,
\end{equation}
where
\begin{eqnarray}
Q &=& \frac{\alpha}{2}B'^T B' + \frac{\beta}{2}I_{2 n_p},\\
r &=& \alpha B'^T(A' X_k + D' - X_o^{ref}), \\
s &=& \frac{\alpha}{2}(A' X_k + D' - X_o^{ref})^T(A' X_k + D' - X_o^{ref}).
\end{eqnarray}
QP solvers such as qpOASES\cite{ferreau2014} and quadprog can be implemented to solve this optimization problem.

\subsection{Force Constraint}

Suppose the friction coefficient at each contact point is known. The interaction force at the contact location must lie within the friction cone.
In this research, the friction cone has been approximately linearized by using the regular 4-sided polygon. We have
\begin{equation}
\begin{split}
    f_i \in FC_i = \{ f_i : |f_i^t| &\leq \mu f_i^n,\\
    |f_i^o| &\leq \mu f_i^n, \\
    0 &\leq f_i^n \leq f_{max}^{n} \} ,   
\label{coulomb}
\end{split}
\end{equation}
where $f^n _{max}$ is a designed upper bound of the normal contact force. An example of the frame $\Sigma_{FC1}$ with axes $\{n, t, o\}$ at contact point $p_{1}$ is shown in Fig. \ref{fig:friction}. 
Since we set $u_k = [f_1 \quad f_2]^T_k$, \eqref{coulomb} acts as constraints to the input,
\begin{equation}
    -L_{max} \leq H_c u_k \leq L_{max} ,
\end{equation}
where 
\begin{equation}
L_{max}=
\left[\begin{matrix}
\mu f_{max}^{n} \quad  \mu f_{max}^{n} \quad f_{max}^{n} \quad \mu f_{max}^{n} \quad  \mu f_{max}^{n} \quad f_{max}^{n}
\end{matrix}\right]^T ,
\end{equation}
denotes the bound on the normal force component and 
\begin{equation}
    H_{ck} = \left[\begin{matrix}
    R_{FC1}^T &\quad  O_3 \\
    O_3 &\quad R_{FC2}^T
  \end{matrix}
  \right] .
\end{equation}

\hl{Considering the free variables taking place in the prediction horizon, we have}
\begin{equation}
    -L'_{max} \leq H'_c U_k \leq L'_{max} ,
\end{equation}
where $L'_{max}$ is a matrix with $n_p$ elements,
$
    L'_{max} = [L_{max}, \cdots, L_{max}]^T 
$
and 
$
    H'_c = [H_{c1}, \cdots, H_{ck}]
$

\subsection{Impedance Control}
After applying MPC, we obtain the desired interaction force as an input $u_k$ and output EEF trajectory by \eqref{eq:output}. This subsection shows how to apply the desired interaction force and the EEF trajectory to the real robot.
\hl{We apply the impedance control} \cite{fukumoto2018force} \hl{where the EEF is controlled to behave like a mechanical impedance.} The desired impedance can be defined as
\begin{equation}
M_{imp}\ddot{x}_{Hi}+D_{imp}\dot{x}_{Hi}+K_{imp}(x_{Hi} - x_{Hi}^{ref}) = f_{i} - f_{i}^{ref},
\label{impedance0}
\end{equation}

\begin{equation}
D_{imp}\dot{x}_{Hi} = f_{i} - f_{i}^{ref},
\label{impedance}
\end{equation}

where $M_{imp}, D_{imp}$ and $K_{imp}$ denote the mass, damping and stiffness matrices, respectively. $x_{Hi}^{ref}$ and $f_{i}^{ref}$ indicate the reference position and force of the $i$-th hand. These references are provided by MPC where $x_{Hi}^{ref}$ is calculated by \eqref{eq:output} and $f_{i}^{ref}$ are provided by $u_k$, the input of MPC. How to control a robot to realize the desired impedance depends on the types of joint servo controllers. In our case, we use a velocity-controlled industrial robot equipped with a function of impedance control. 

\section{Graph Model Predictive Control}
\label{sec4:graph}

To adaptively select the gait mode, we introduce the graph MPC.
If a disturbance is detected by the vision and force sensors, we update the weight of the edge.  Then, the graph selects the new gait mode and provides a reference trajectory of the object to the MPC. The MPC drives the object from the current state to the reference state and realizes a feedback control. 

\subsection{Graph}

The graph expresses the change of contact states between the object and ground while the cost function, as a weight of each edge, is designed to select a path. By searching for the graph, we can select a gait mode with the lowest cost and outputs a sequence of key poses as a reference to MPC. 

\begin{figure}
\centering
\includegraphics[width=0.65\linewidth]{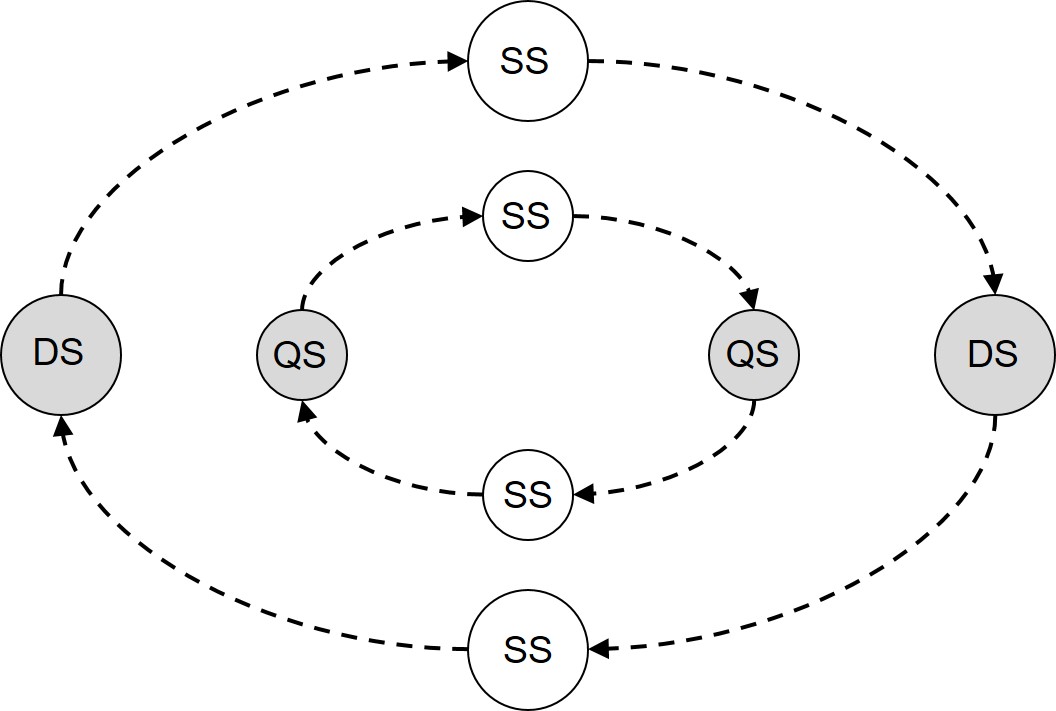}
\caption{A node indicates a supporting state of the object. A gait mode is represented by four nodes. A loop indicates a principle walking motion. The inner and outer loops correspond to QS and DS gait modes, respectively.}
\label{fig:steady state}
\end{figure}

In the graph, each node includes the object's configuration and a supporting state of the object, i.e., SS, DS, and QS phases. 
To compose the graph, we define two types of motions: principal motion and switching motion. The principal motion is cyclic as shown in Fig. \ref{fig:steady state}. 
The inner loop corresponds to a cyclic motion of the QS gait mode where the SS and QS phases alternate. On the other hand, the outer loop corresponds to a cyclic motion of the DS gait mode. Here, the SS phase includes the right-foot and left-foot support phases. The right-foot SS phase comes after the DS or QS phases coming after the left-foot SS phase, and vise versa. On the other hand, the switching motion is a transient motion from one principal motion to the other.

\begin{figure}
\centering
\includegraphics[width=0.45\textwidth]{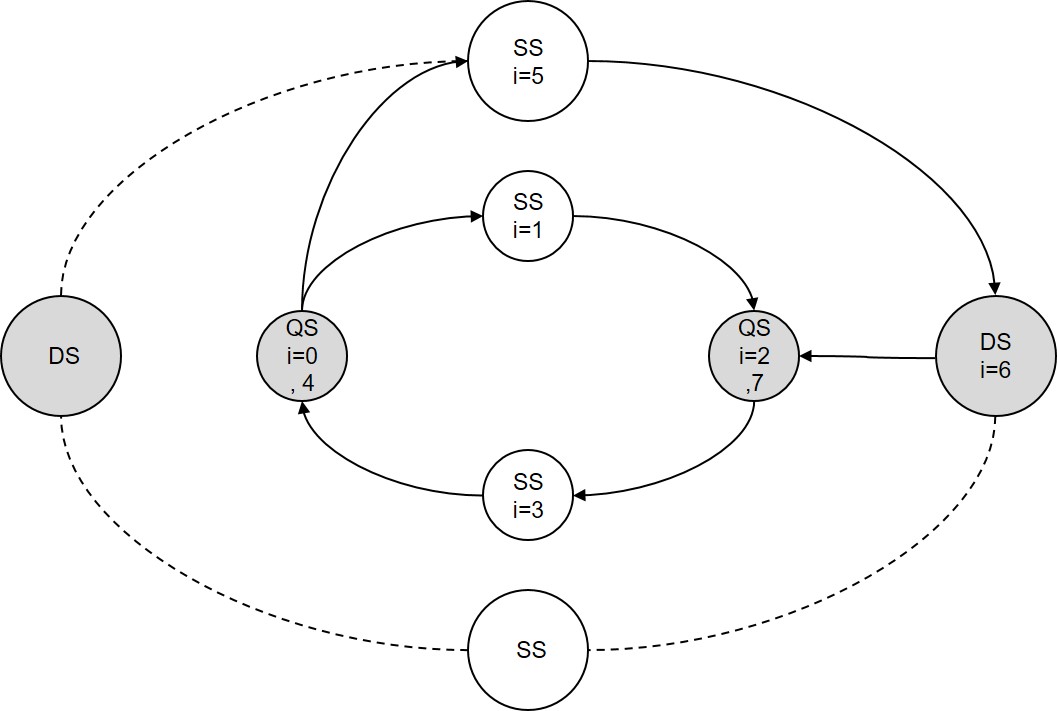}
\caption{Example of a graph. $i$ denotes a sequence of motion. The solution path is drawn by solid arrows while the candidate path is drawn by dashed arrows. The object walks in QS mode firstly, then walks in DS mode, and finally rests in a QS pose.}
\label{fig:graphMPC}
\end{figure}

An overview of the graph including an example of solution path is shown in Fig. \ref{fig:graphMPC}. 
Nodes are connected by edges which are drawn by solid and dashed arrows where solid arrows denote an example of the solution path. 
In addition, the nodes included in the solution path include the number $i$ which implies the sequence.  
In the example shown in Fig. \ref{fig:graphMPC}, the object starts from the initial node ($i=0$), walks in the QS mode firstly (from $i = 0$ to $4$), then switches to the DS mode ($i=5, 6$) and finally rests at QS pose ($i=7$).  

\subsection{Weights of edges: Cost Functions}

Each edge of the graph includes two pieces of information, where one is the transient time to move from one node to the other node, and the other piece is the weight.  
In this subsection, we design a cost function as the edge's weight. To evaluate the desirability of the gait mode, we compare the cost of paths and select a gait mode with the lowest cost.  
We assign the cost $J_{path}$ of a path which includes $n$ edges. It is composed of a cost $J_{s}$ which is related to the states, $J_t$ related to time, and $J_{dst}$ related to external disturbance,
\begin{equation}
    J_{path} = \alpha_g J_{s} + \beta_g J_t + \gamma_g J_{dst} ,
    \label{eq:cost of path}
\end{equation}
where $\alpha_g, \beta_g$ and $\gamma_g$ denote the weights of each cost function. Weights can be set free under different purposes, for example, if a fast walking motion is wanted, we can increase the value of $\beta_g$. The cost functions are defined as follows, $J_s$ indicates the summation of transition costs between nodes. $J_t$ indicates the summation of times taken to complete the transition. $J_{dst}$ is designed for updating the cost once a disturbance is detected.
\begin{equation}
    J_{s} = \sum_{i=0}^{n-1} \left \| x_i - x_{i+1} \right\|^2 ,
    \label{eq:cost of step}
\end{equation}
\hl{where $x_i$ is the reference state of the object saved in node $i$ and the state in each node is designed with respect to the way of support, for example, single support, double support, and quadruple support.} $n$ indicates the number of edges along the path. In this work, we select $n = 4$ because four edges are enough to represent a gait mode. 
\begin{equation}
    J_t = \sum_{i=0}^{n-1} t_{i} ,
\end{equation}
where $t_i$ is the duration of time taken for the transition from $x_i$ to $x_{i+1}$ along the edge. The duration of time is manually selected and the duration of time for the DS mode is selected smaller than that of the QS mode because the DS mode is designed to move faster.
Note that the time duration in graph between $x_i$ and $x_{i+1}$ is $t_i$ which is different from sampling time $T$ between $x_{k}$ and $x_{k+1}$ in MPC.  
A cost function of disturbance $J_{dst}$ is defined as
\begin{equation}
    J_{dst} = \left\{
\begin{array}{rcl}
\delta   &      & \text{if disturbance is detected}\\
0        &      & \text{otherwise.}
\end{array} \right .
\label{eq:cost of dst}
\end{equation}
$\delta$ is a positive value indicating the existence of external disturbance. 
\begin{equation}
    \delta = \left\{
\begin{array}{rcl}
\delta_{ds}   &      & \text{for DS gait mode}\\
\delta_{qs}   &      & \text{for QS gait mode.}
\end{array} \right .
\end{equation}
$\delta$ includes both $\delta_{qs}$ and $\delta_{ds}$. We set $\delta_{ds} >  \delta_{qs}$ because disturbance causes a bigger influence on DS mode. 

Both cameras and force sensors are used to detect the disturbance. Define $\Psi^{thr}$ and $f_i^{thr}$ as the thresholds to detect the occurrence of disturbance. The external disturbance is detected if the following conditions are met 
\begin{equation}
    | \Psi^{cur} - \Psi^{pre} | \geq  \Psi^{thr} ,
    \label{eq:euler diff}
\end{equation}
\begin{equation}
    | f_i^{cur} - f_i^{pre} | \geq  f_i^{thr} ,
    \label{eq:force diff}
\end{equation}
where $\Psi^{cur}$ and $f_i^{cur}$ are the current Euler angles detected by the cameras and force data collected from the force sensor on $i$-th arm, respectively. $\Psi^{pre}$ and $f_i^{pre}$ indicate the data at the previous sampling time. After designing nodes and edges, we can select the gait mode by looking for the optimal path with minimum cost.

\subsection{Selection of gait modes in graph}
\begin{figure}
\centering     
\subfigure[Candidate paths start from the top node. The inner and the outer path represent the QS and DS modes, respectively.]{\label{fig:a}\includegraphics[width=0.4\linewidth]{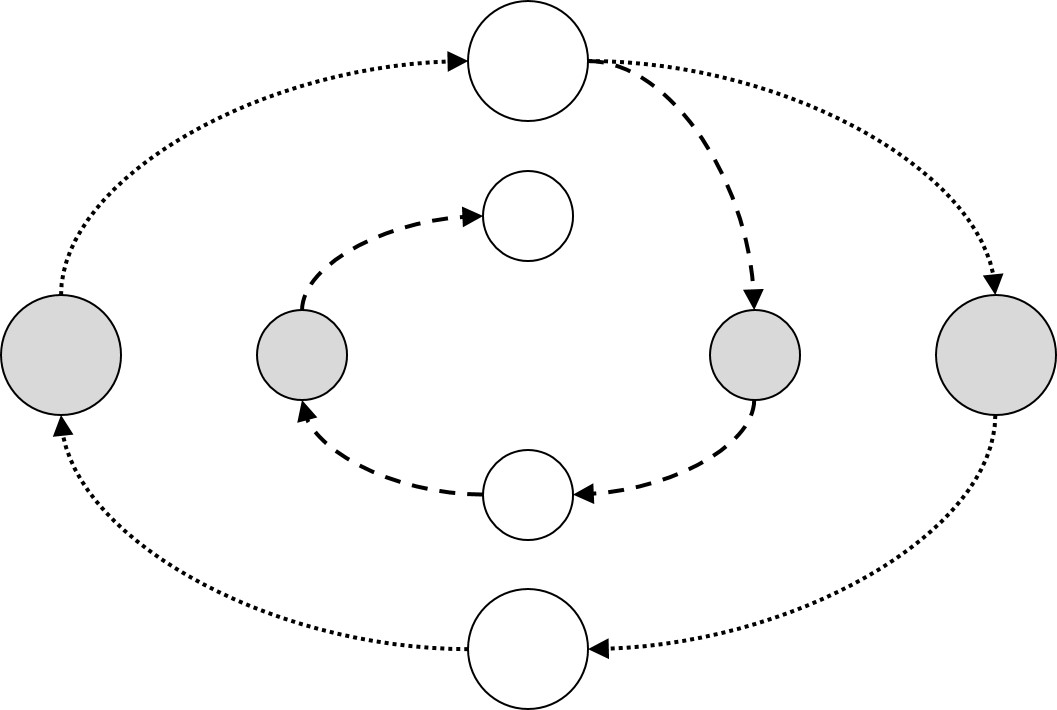}}
~
~
\subfigure[Select the optimal path with the lowest cost. The other candidate path is removed and there is a transition from the top node to the right node along the optimal path.]{\label{fig:b}\includegraphics[width=0.4\linewidth]{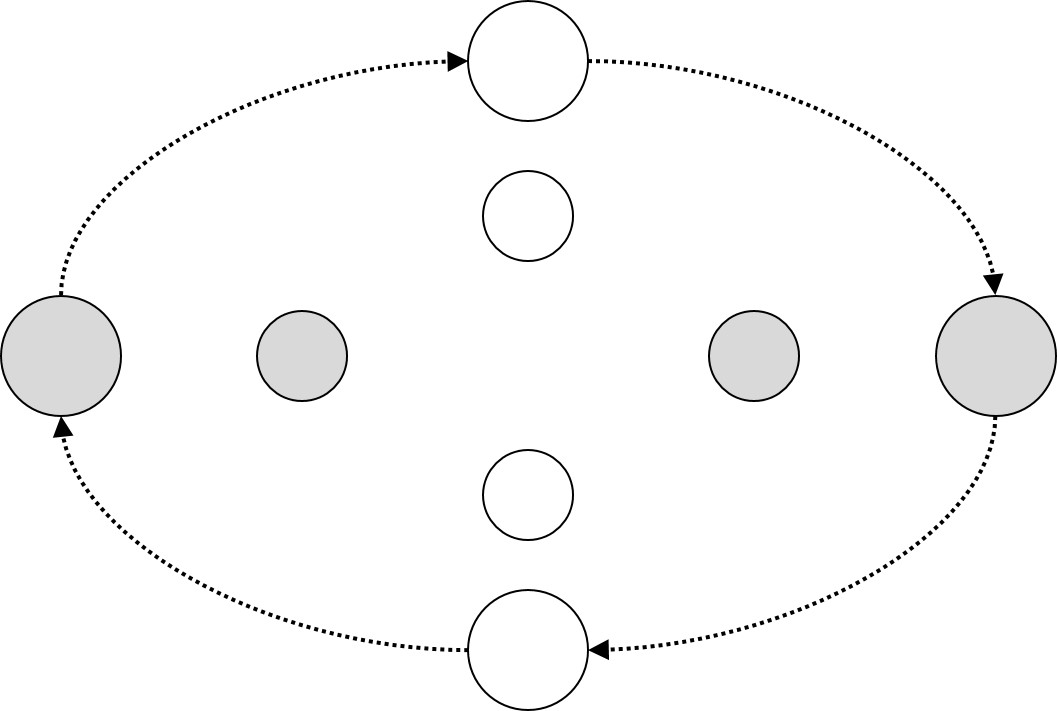}}


\subfigure[If a disturbance is detected, the costs of path \eqref{eq:cost of path} will be updated by $\gamma_{g} \delta_{qs}$ or $\gamma_g \delta_{ds}$.]{\label{fig:c}\includegraphics[width=0.4\linewidth]{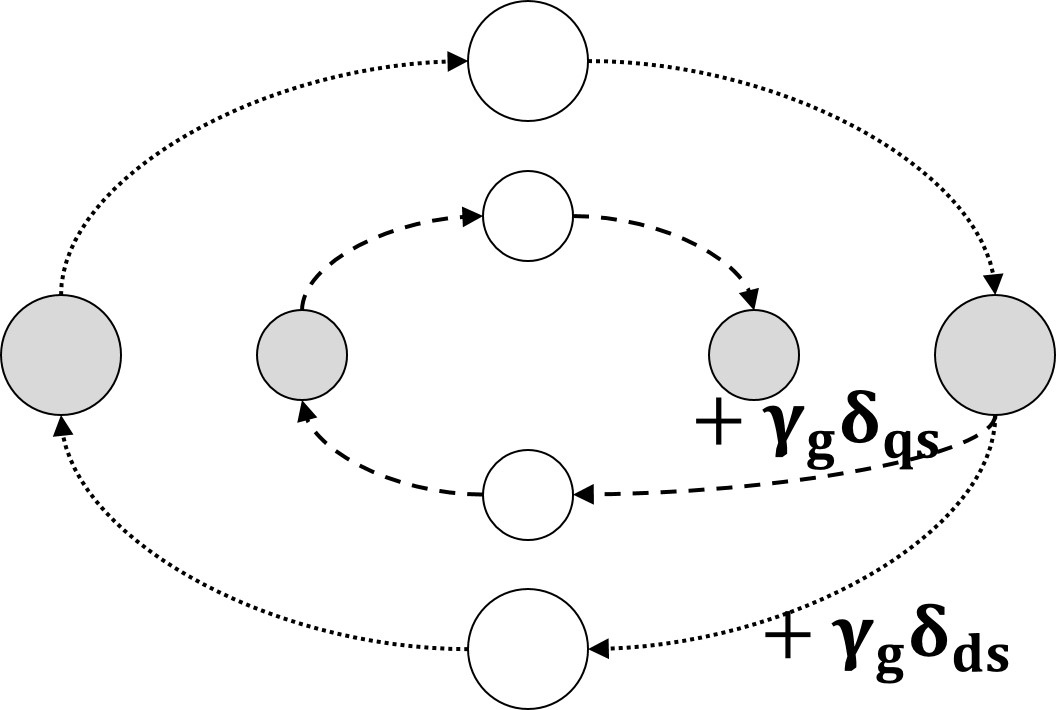}}
~
~
\subfigure[Select the optimal path by comparing the costs. A switch of gait mode from the DS to the QS mode is achieved.]{\label{fig:d}\includegraphics[width=0.4\linewidth]{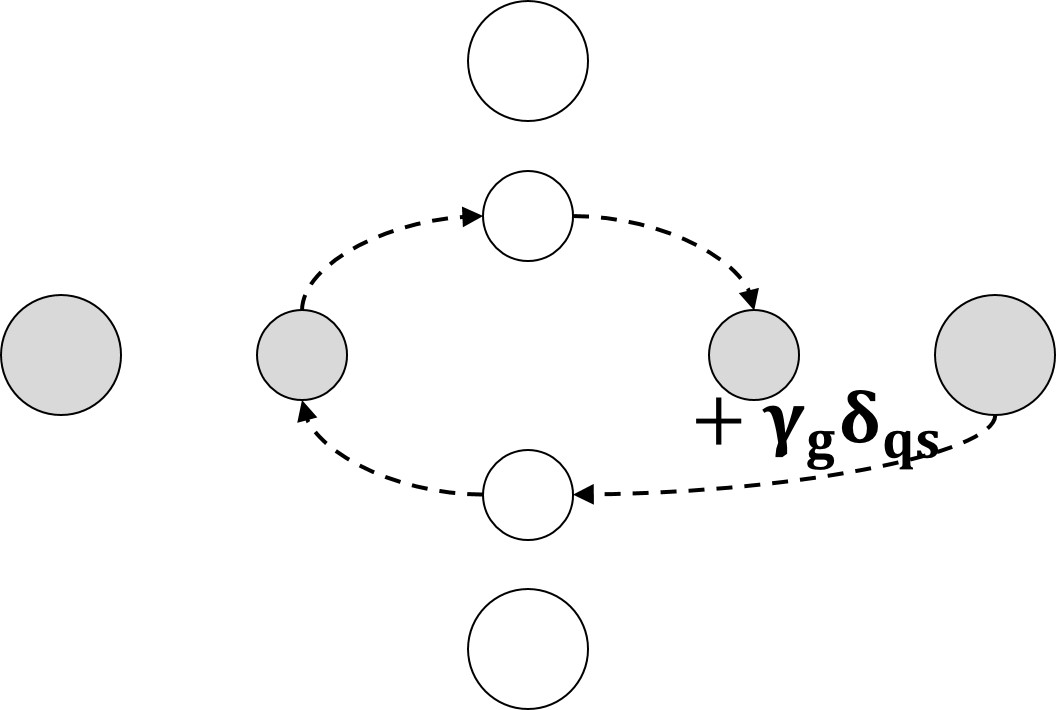}}
\caption{Procedures to select a gait mode in the graph.}
\label{fig:subgraph}
\end{figure}

By comparing the cost function of paths, we can select an optimal path that contains information on gait modes. Fig. \ref{fig:subgraph} showed the procedures of finding the optimal path:

The top node is selected as a starting node, see Fig.\ref{fig:a}. Suppose $n = 4$ in a candidate path, two candidate paths are shown by directed arrows. The inner dashed arrow indicates the QS gait mode while the outer dot arrow indicates the DS gait mode.  

By comparing the cost, $J_{path}$ of candidate paths, only the path with the lowest cost is remained and selected as the optimal path, see Fig.\ref{fig:b}.
Then, a transition happens from the current node to the next node along the optimal path.

If a disturbance is detected, the costs of paths in \eqref{eq:cost of path} will be updated because of the change of $J_{dst}$ in \eqref{eq:cost of dst}. This change can be reflected in the graph, for example, see Fig.\ref{fig:c}. $\gamma_g \delta_{qs}$ and $\gamma_g \delta_{ds}$ are added to the costs of the inner path and outer path, respectively. After comparing the costs of paths, the inner path is selected as the optimal path, see Fig.\ref{fig:d}. A transition from the outer path to the inner path implies that the gait mode is switched from the DS mode to the QS mode.

In this work, the main task of the graph is to select gait modes and generates a reference trajectory of the object corresponding to the selected gait modes. The reference trajectory is then provided to MPC to track. 
If there is a disturbance during motion, the robot will stop its motion and wait for the new command. Meanwhile, edges in the graph are updated where $J_{dst}$ is added to their weights. Considering the change of edges, the graph selects new gait modes and outputs a reference trajectory to MPC. Then, based on the current state and the reference, the MPC generates the trajectory of EEFs and robot moves.     

\section{Simulation and Experiment}
\label{sec5:simulation}

The target object is a box in the size of 0.6$\times$0.4$\times$0.2m and its weight is 1.4kg. \hl{We use a dual-arm robot (Yaskawa Motoman SDA5F) to manipulate the box and change the gait mode during motion.} 

\subsection{Simulation}
\begin{figure}
\centering
\includegraphics[width=0.45\textwidth]{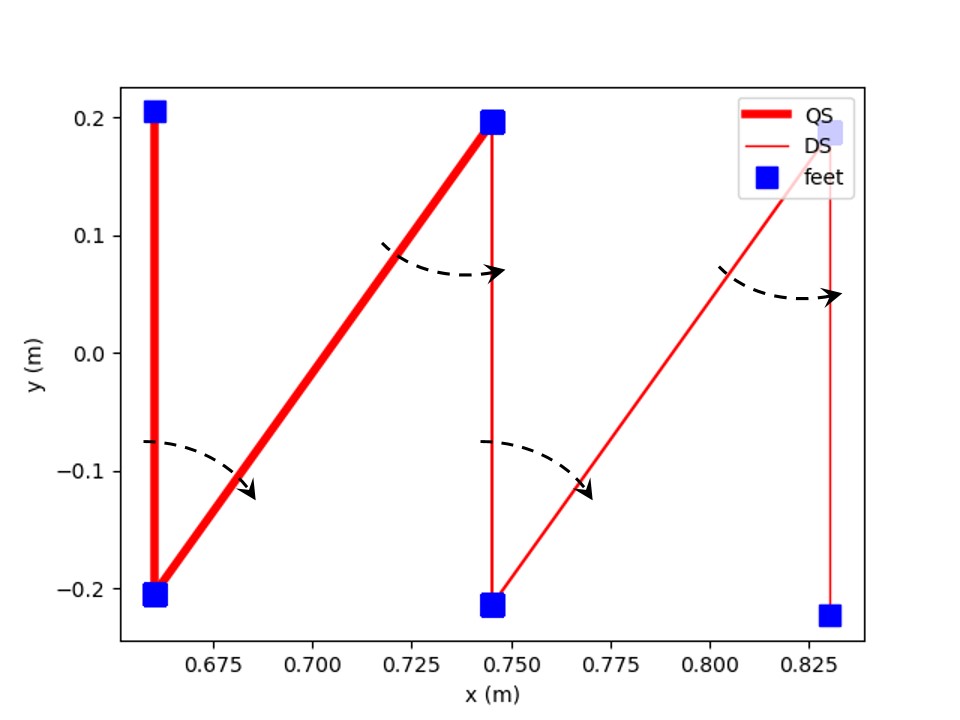}
\caption{Footprints of walking motion. The blue squares are the hind support feet of the object. The red line indicates the edge connecting hind feet where the wider and thinner lines imply QS and DS mode, respectively. The black dashed line implies the swing motion. The object starts to walk in QS mode for one step and then walks forward for 3 steps in DS mode.}
\label{fig:gaitSimu}
\end{figure}

In the simulation, we assume that the object walks stably at the beginning and then walks fast which involves a change of gait mode. This can be simply done by modifying weights in \eqref{eq:cost of path}.
The obtained sequence of modes is the QS mode for the first step and then the DS mode for three steps. 
In Fig.\ref{fig:gaitSimu}, the top view of the gaits generated by MPC is shown where the support feet are marked as blue squares, the edge connecting two support feet is drawn in red lines where the wider line indicates the QS mode and the thinner line indicates the DS mode. We can observe that gait mode changed from QS to DS after the first step. The rotation around the support foot is shown in black dashed arrows and the rotation indicates the yaw angle of the object. 

\begin{figure}
\centering     
\subfigure[]{\label{sim1}\includegraphics[width=0.22\textwidth]{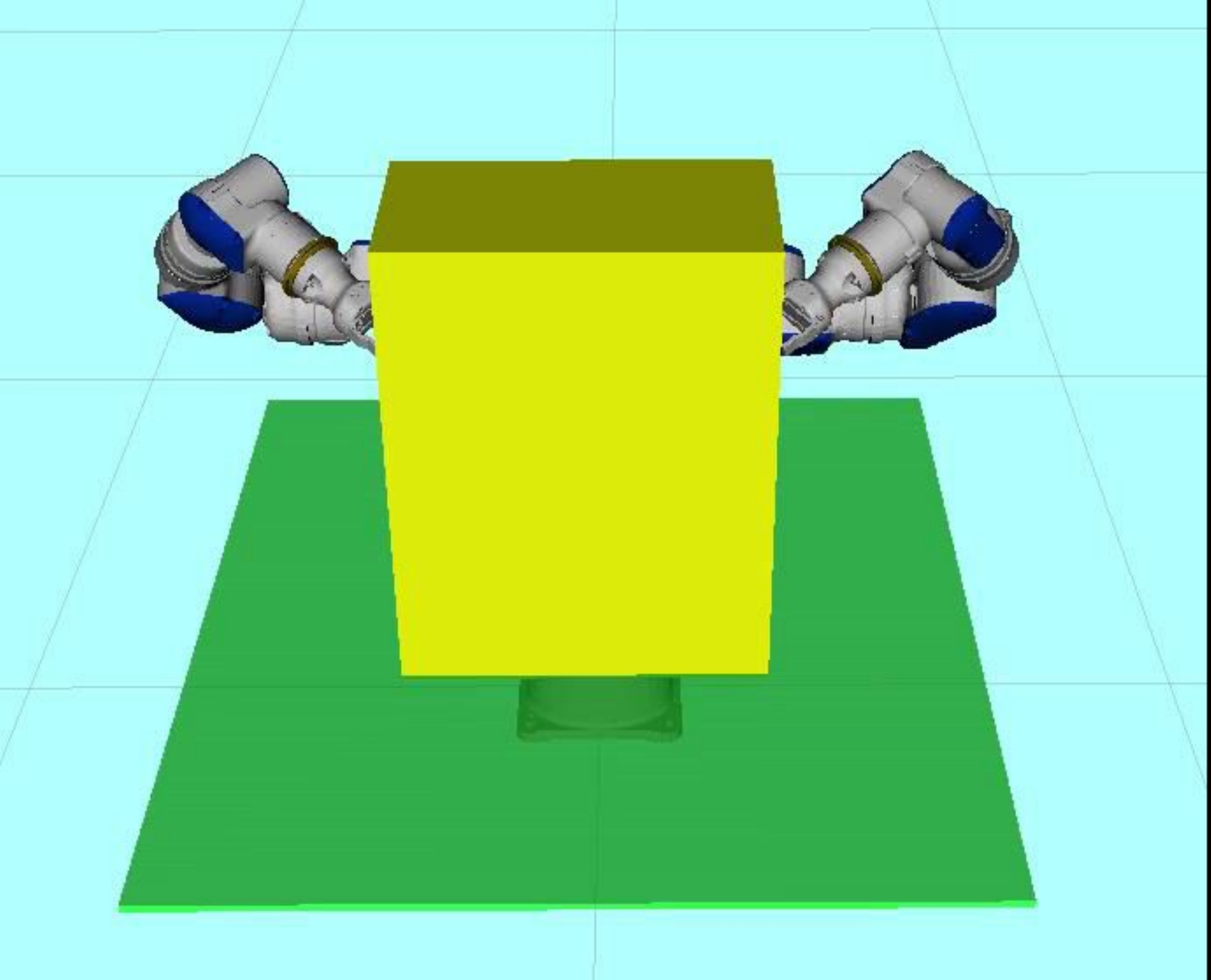}}
\subfigure[]{\label{sim2}\includegraphics[width=0.22\textwidth]{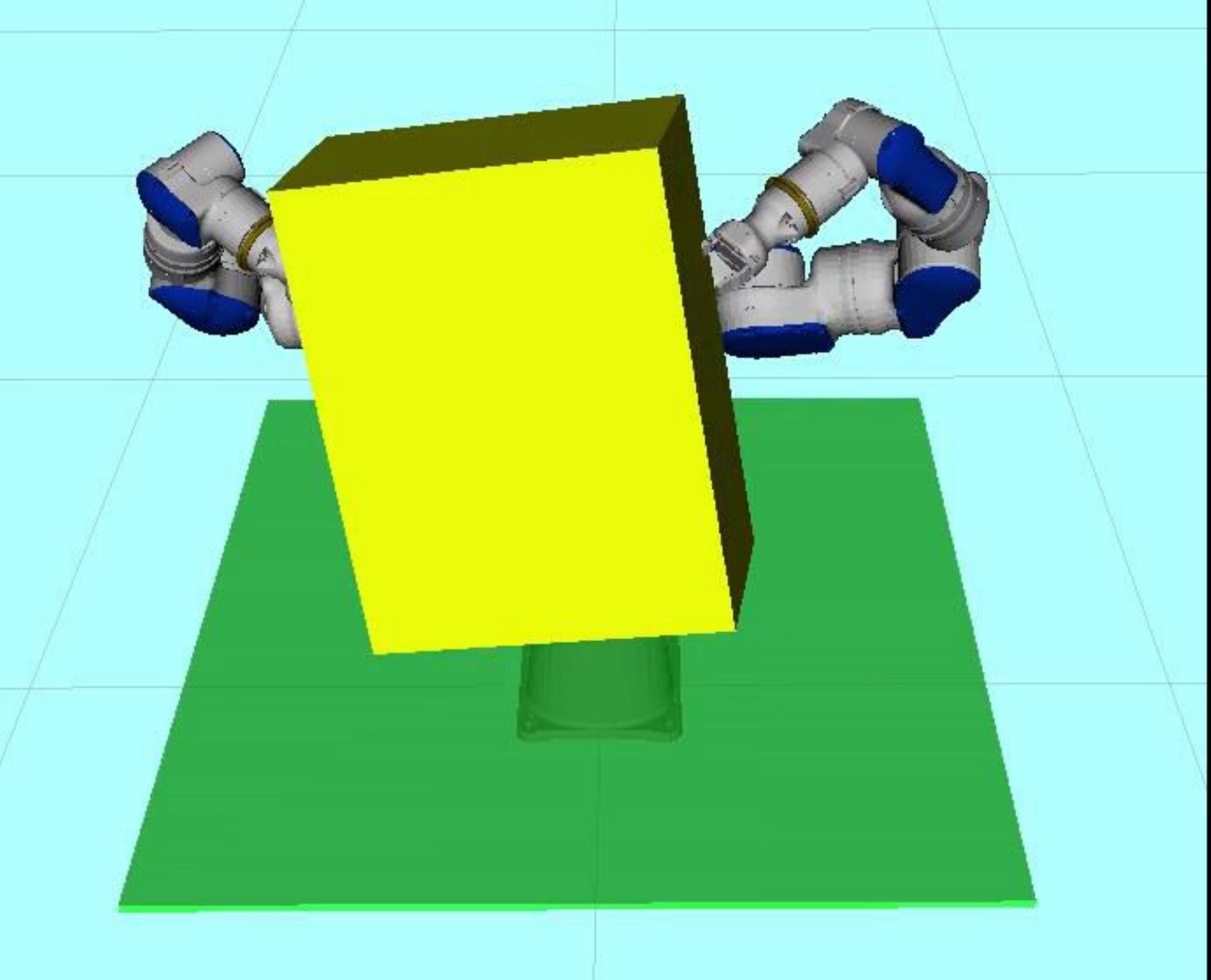}}
\subfigure[]{\label{sim3}\includegraphics[width=0.22\textwidth]{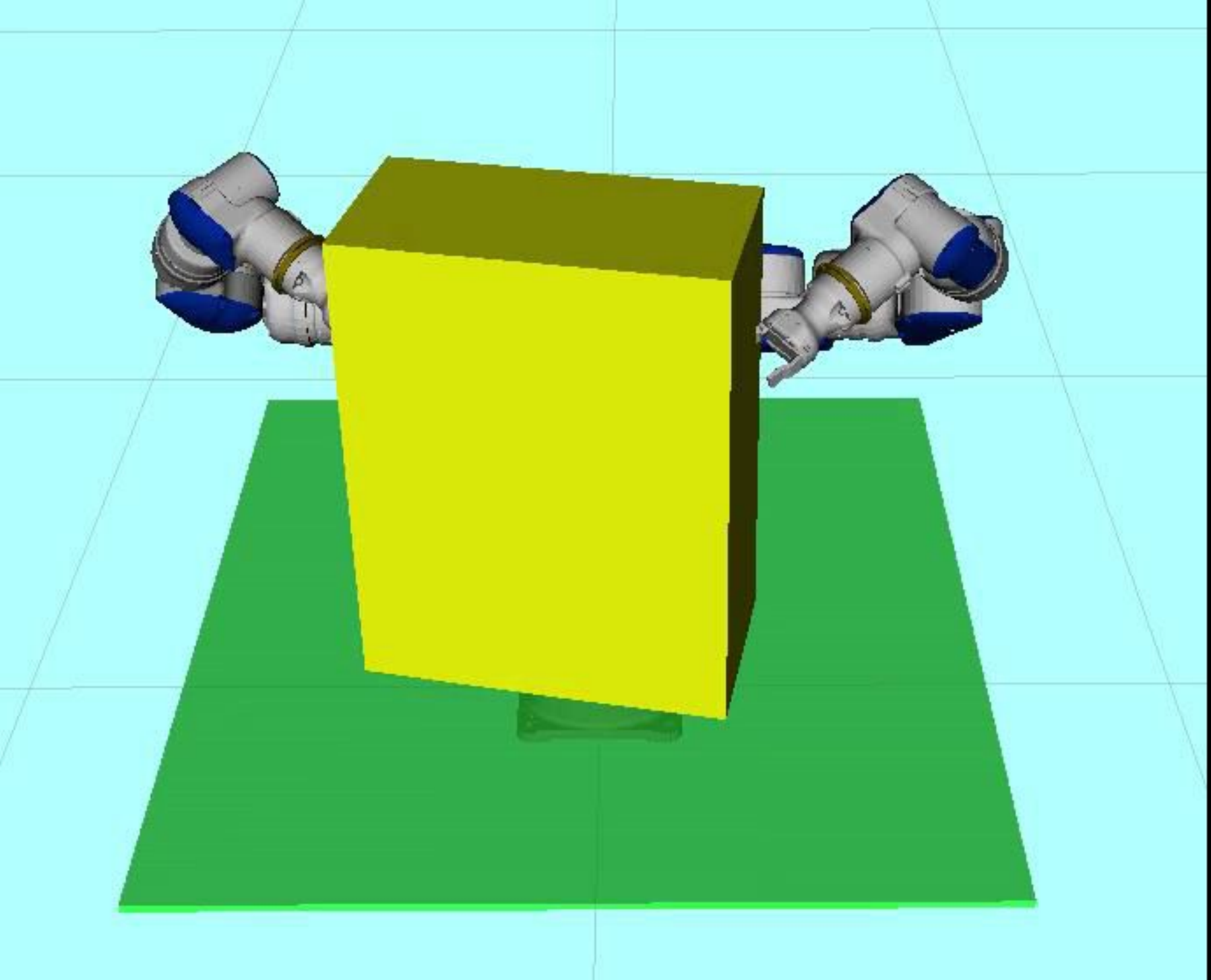}}
\subfigure[]{\label{sim4}\includegraphics[width=0.22\textwidth]{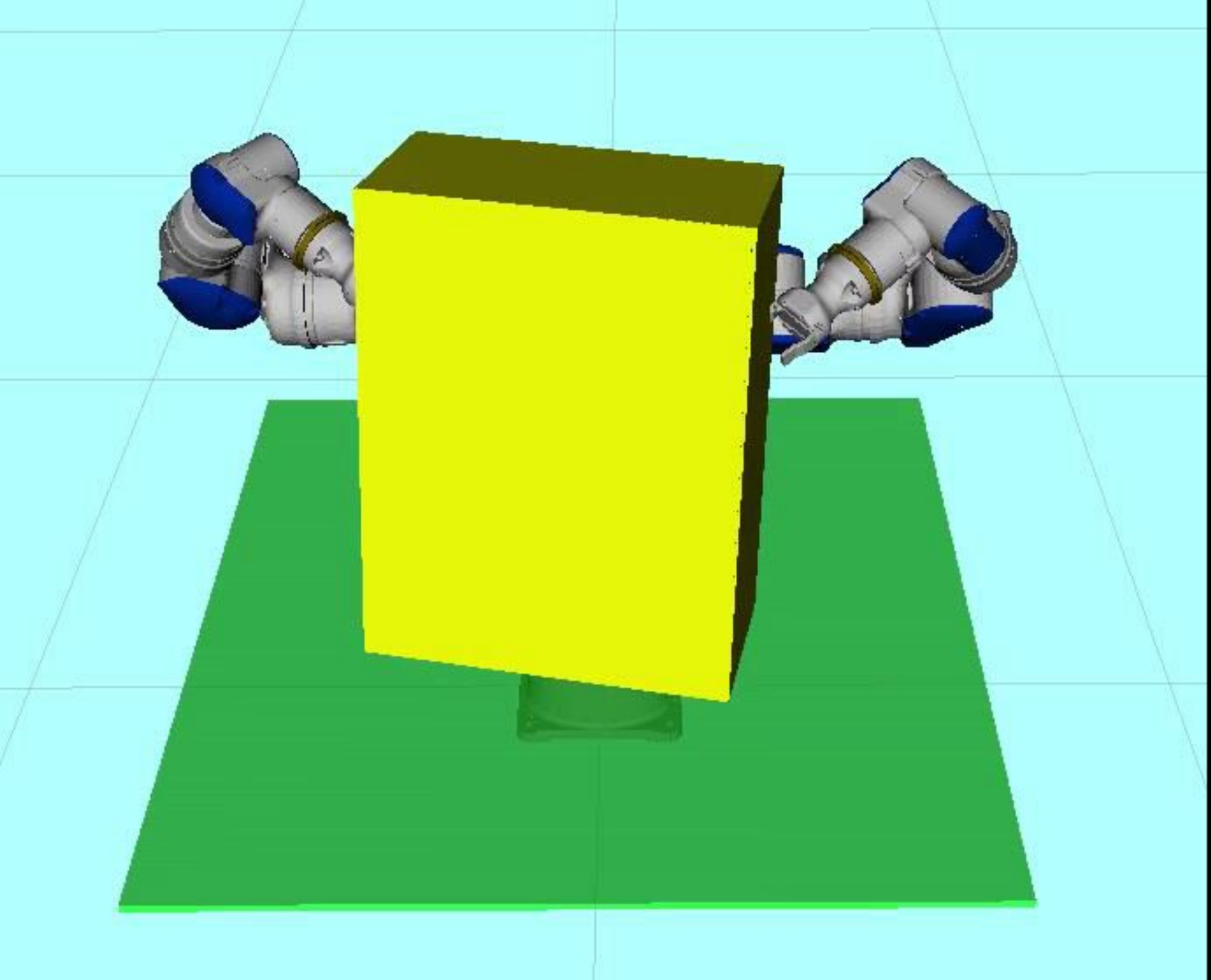}}
\subfigure[]{\label{sim5}\includegraphics[width=0.22\textwidth]{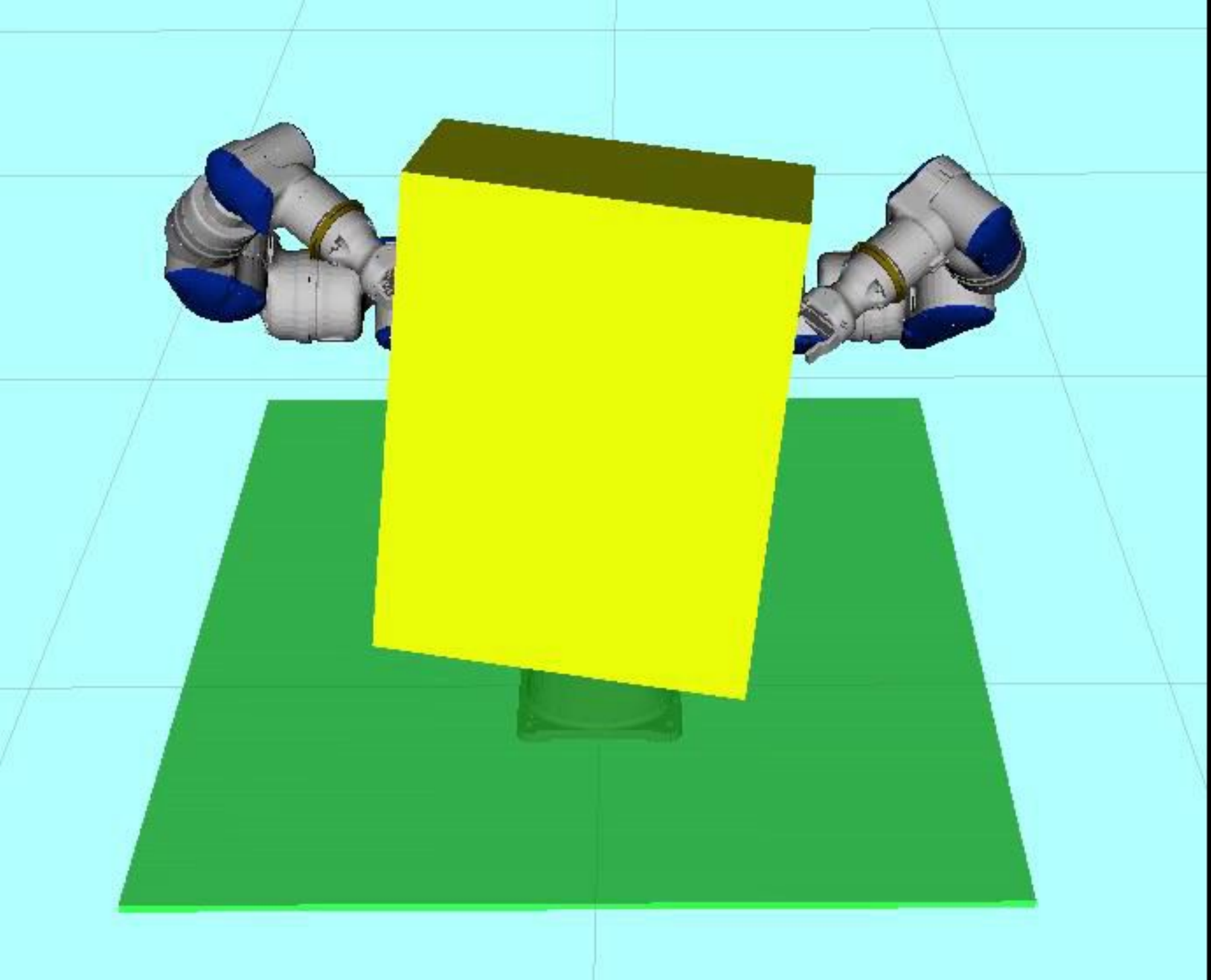}}
\subfigure[]{\label{sim6}\includegraphics[width=0.22\textwidth]{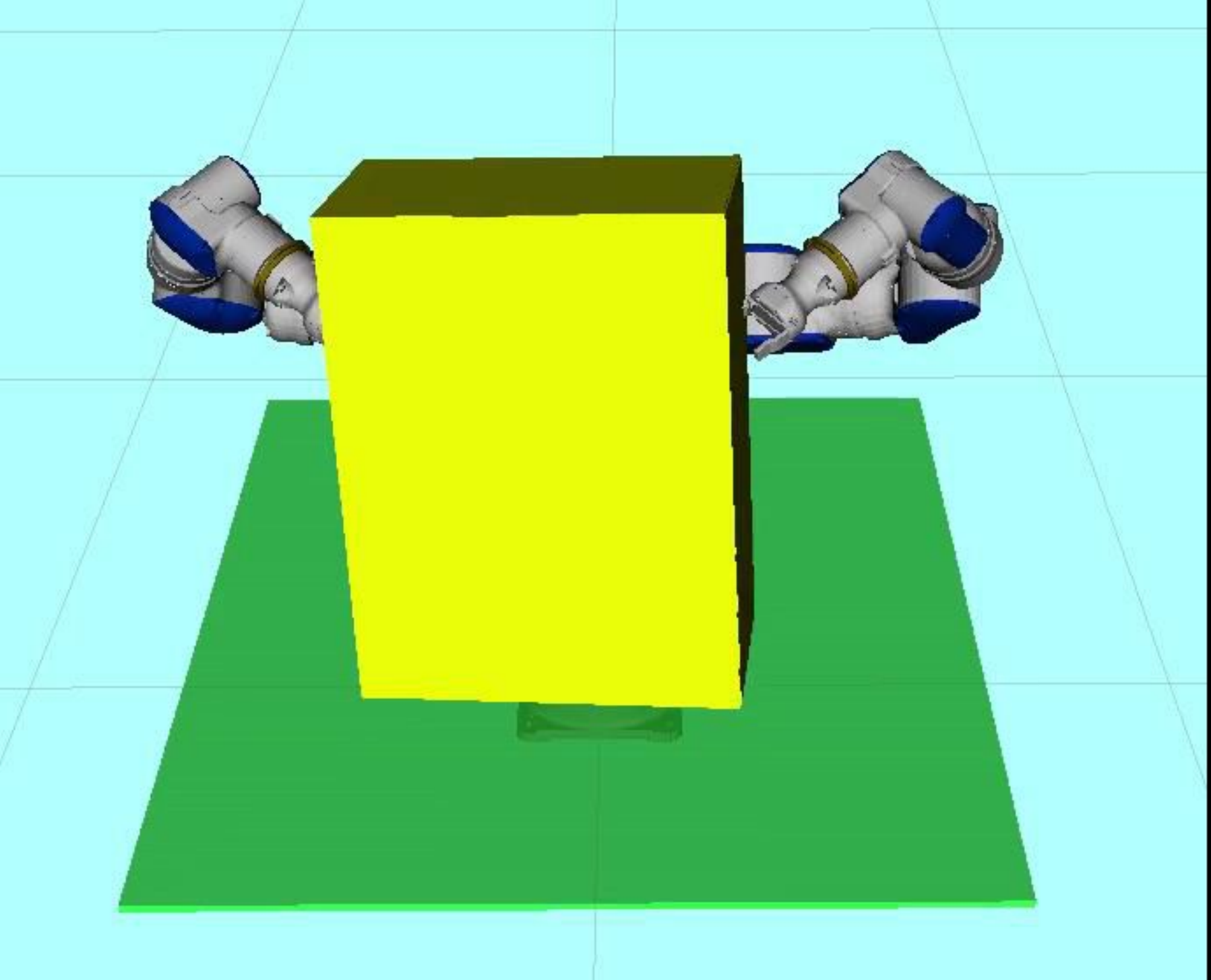}}
\subfigure[]{\label{sim7}\includegraphics[width=0.22\textwidth]{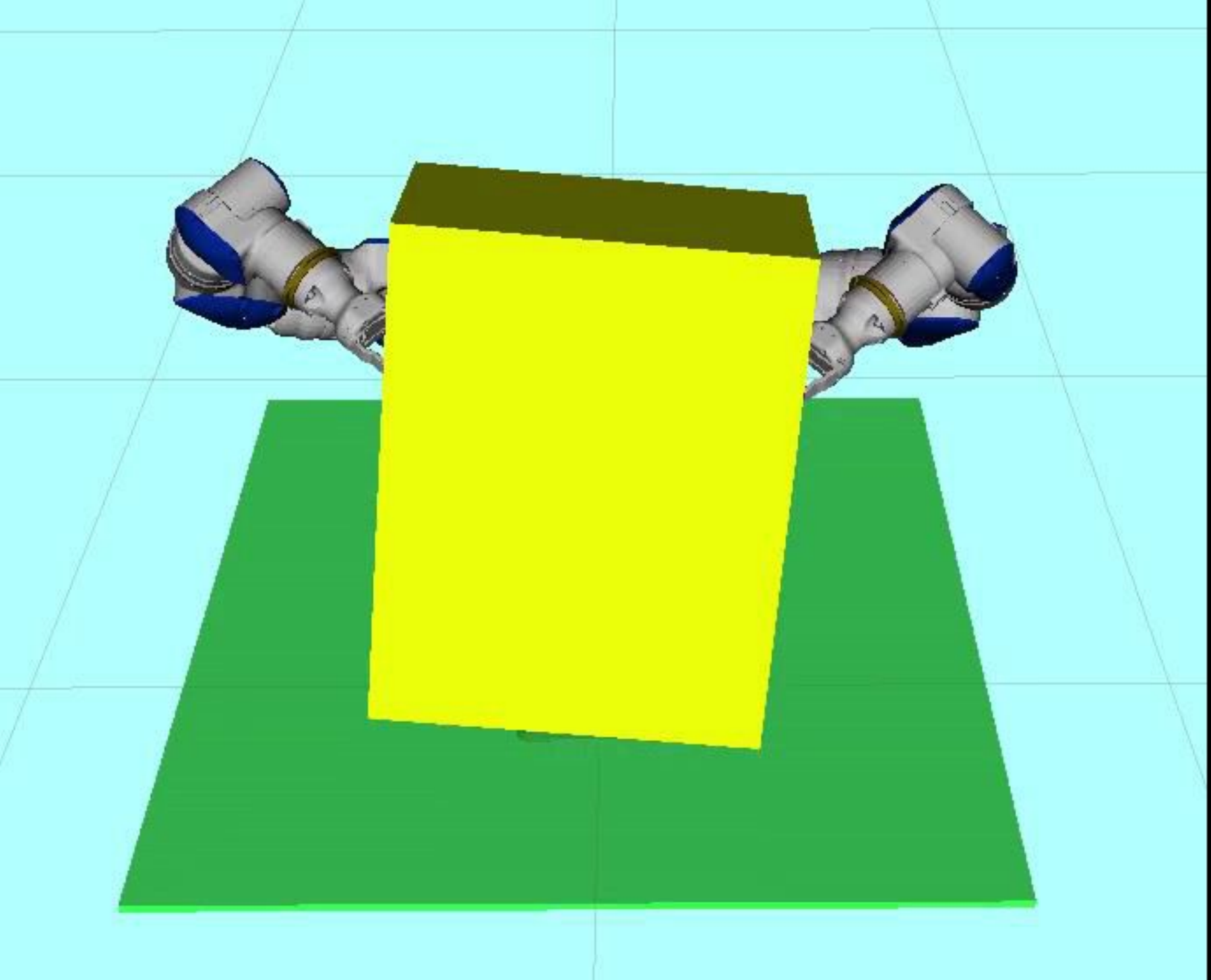}}
\subfigure[]{\label{sim8}\includegraphics[width=0.22\textwidth]{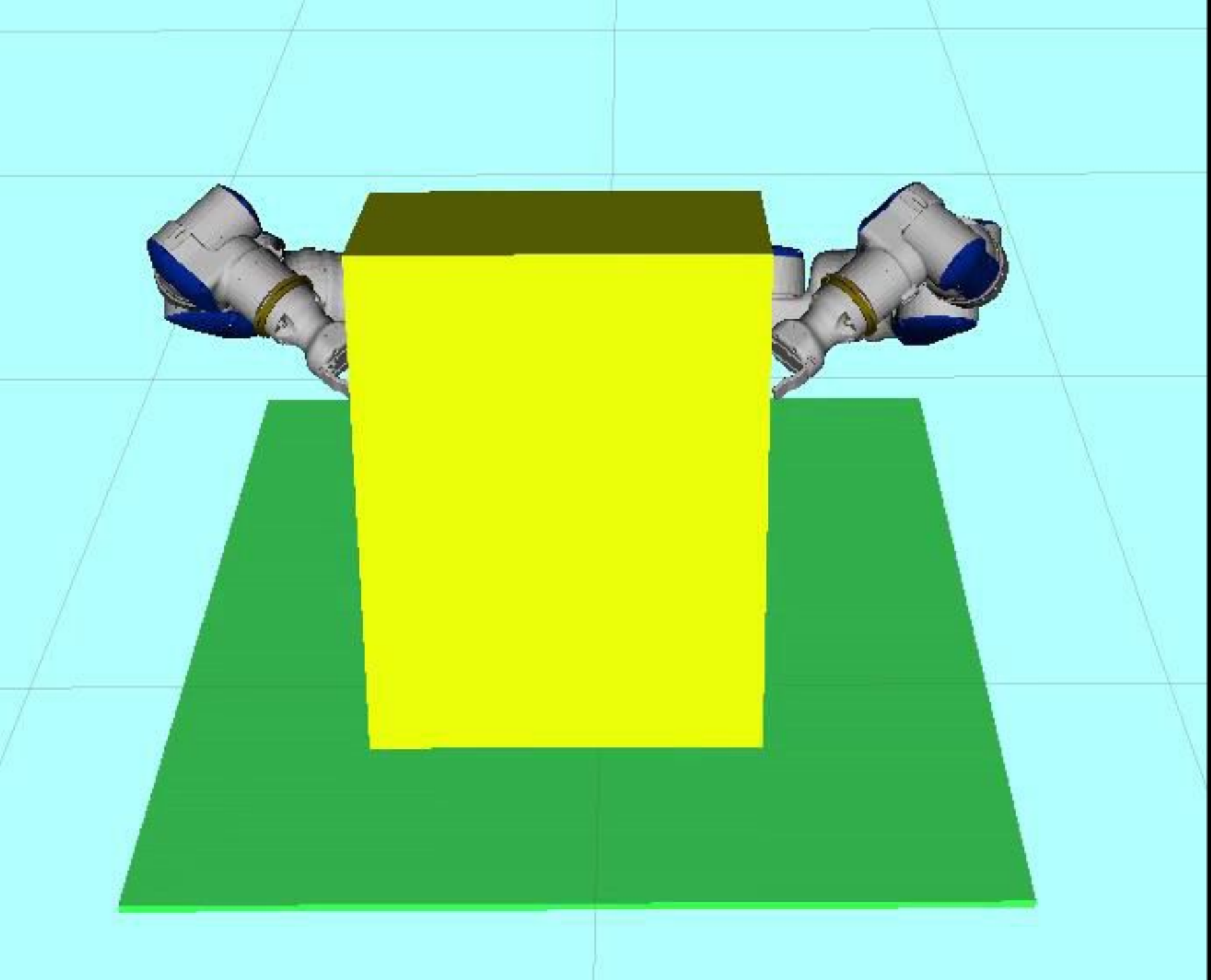}}
\caption{Simulation of pivoting gait. The walking motion starts from QS mode for the first step and then switches into DS mode for 3 steps. }
\label{simulation}
\end{figure}

We simulate the walking motion in RVIZ. In Fig.\ref{sim2}, the object rotates around the vertex (we define it as right foot) near the right EEF of the robot. Then, the gait mode transforms from QS (Fig.\ref{sim3}) to DS (Fig.\ref{sim4}). The supporting foot changes to the left in Fig.\ref{sim5}. Similarly, a foot change happens in Fig.\ref{sim6} and Fig.\ref{sim7} and finally, the object is moved to a rest pose in Fig.\ref{sim8}. \hl{The simulation shows that a switch of gait mode can be realized during the motion of pivoting gait.}  

\subsection{Experiment 1: DS and QS mode}

\begin{figure*}
\centering     
\subfigure[]{\label{ds1}\includegraphics[width=0.16\textwidth]{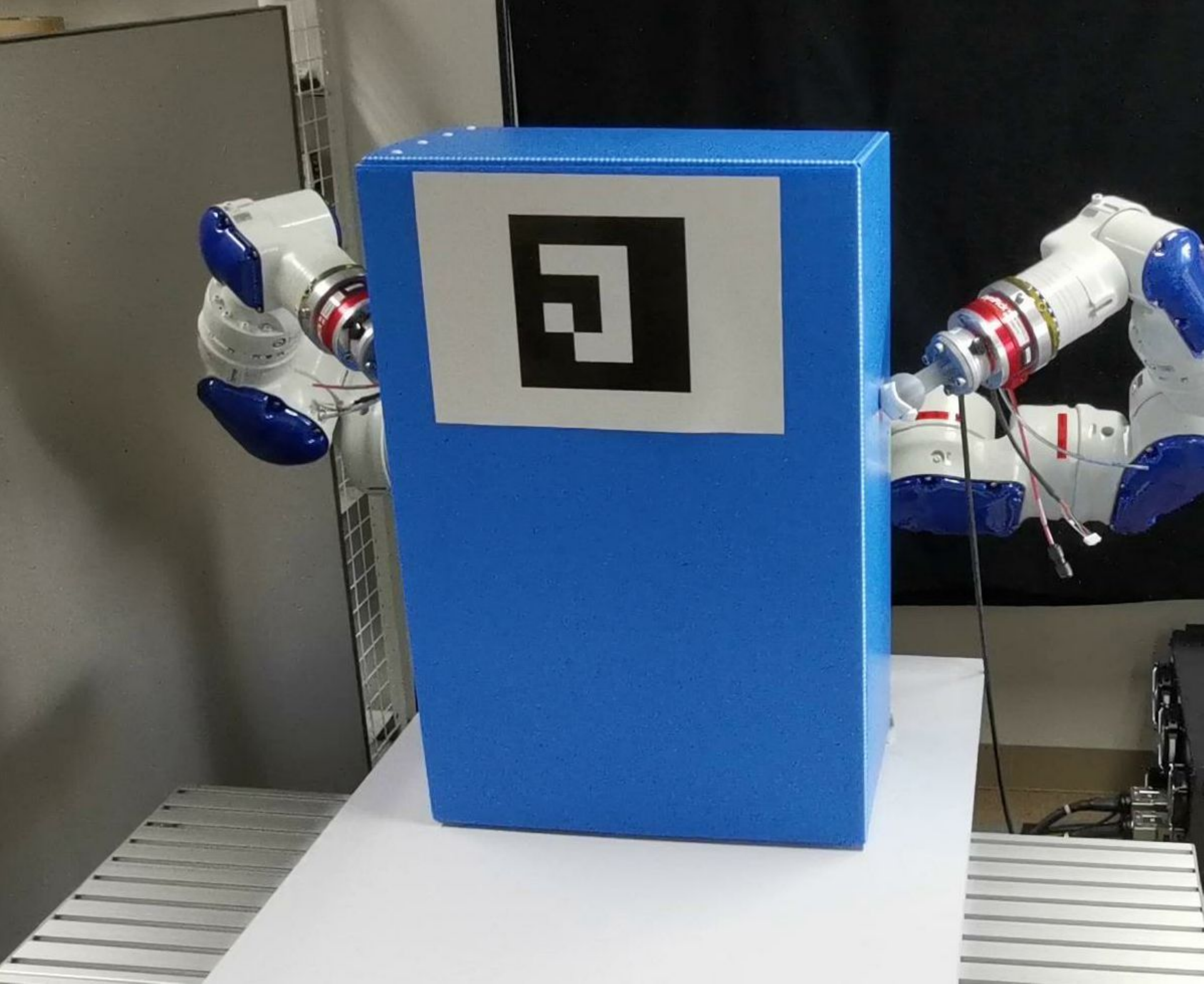}}
\subfigure[]{\label{ds2}\includegraphics[width=0.16\textwidth]{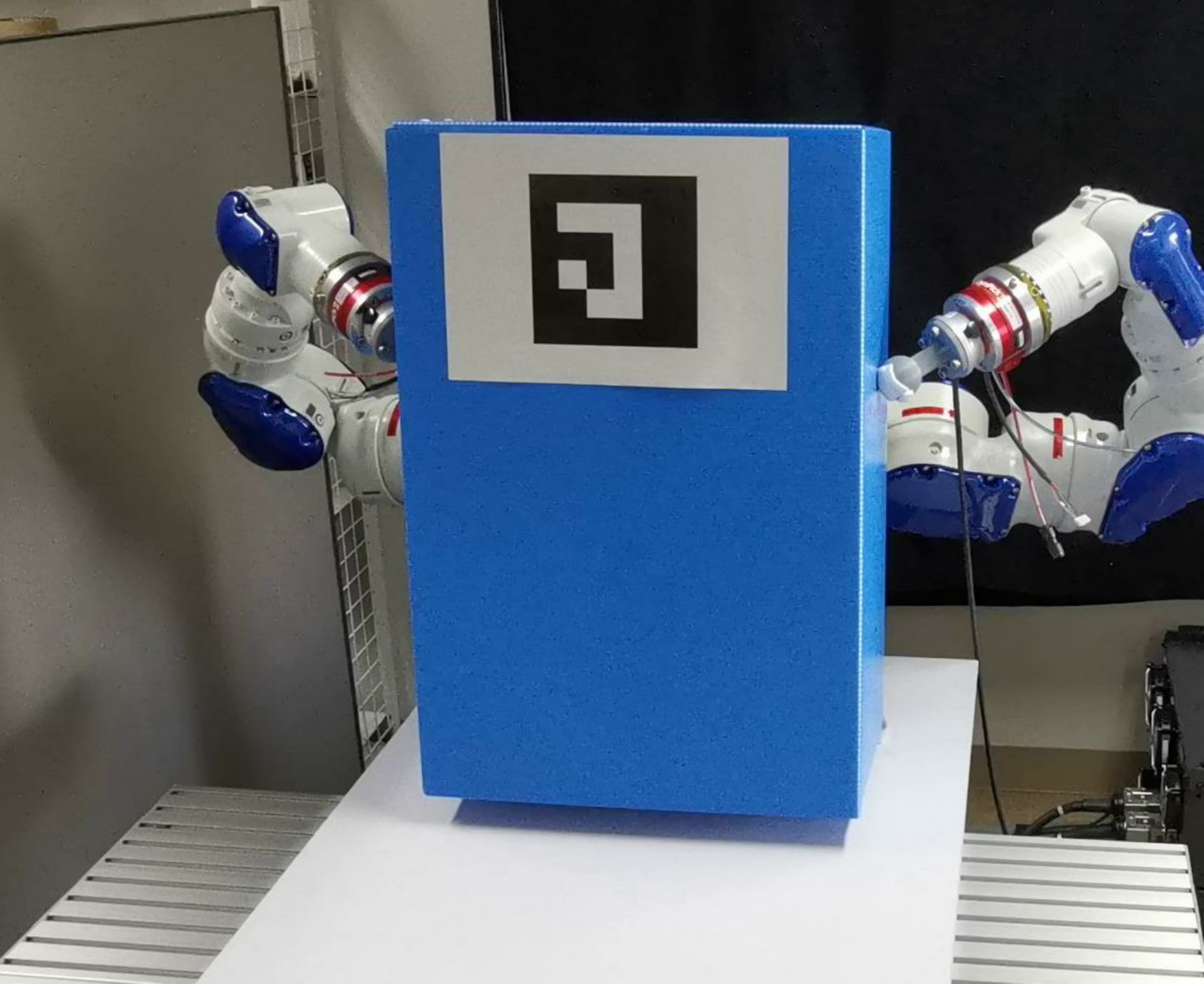}}
\subfigure[]{\label{ds3}\includegraphics[width=0.16\textwidth]{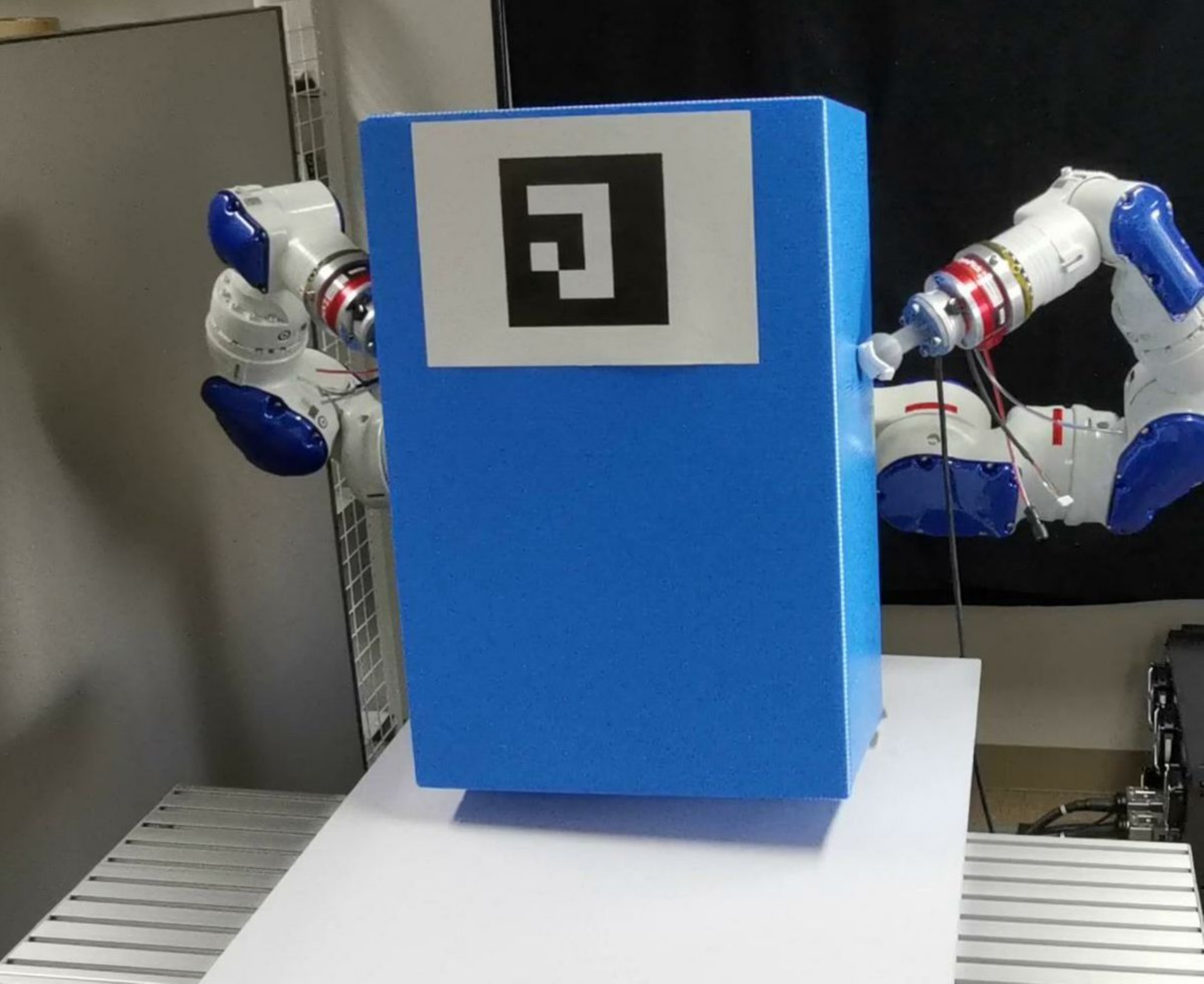}}
\subfigure[]{\label{ds4}\includegraphics[width=0.16\textwidth]{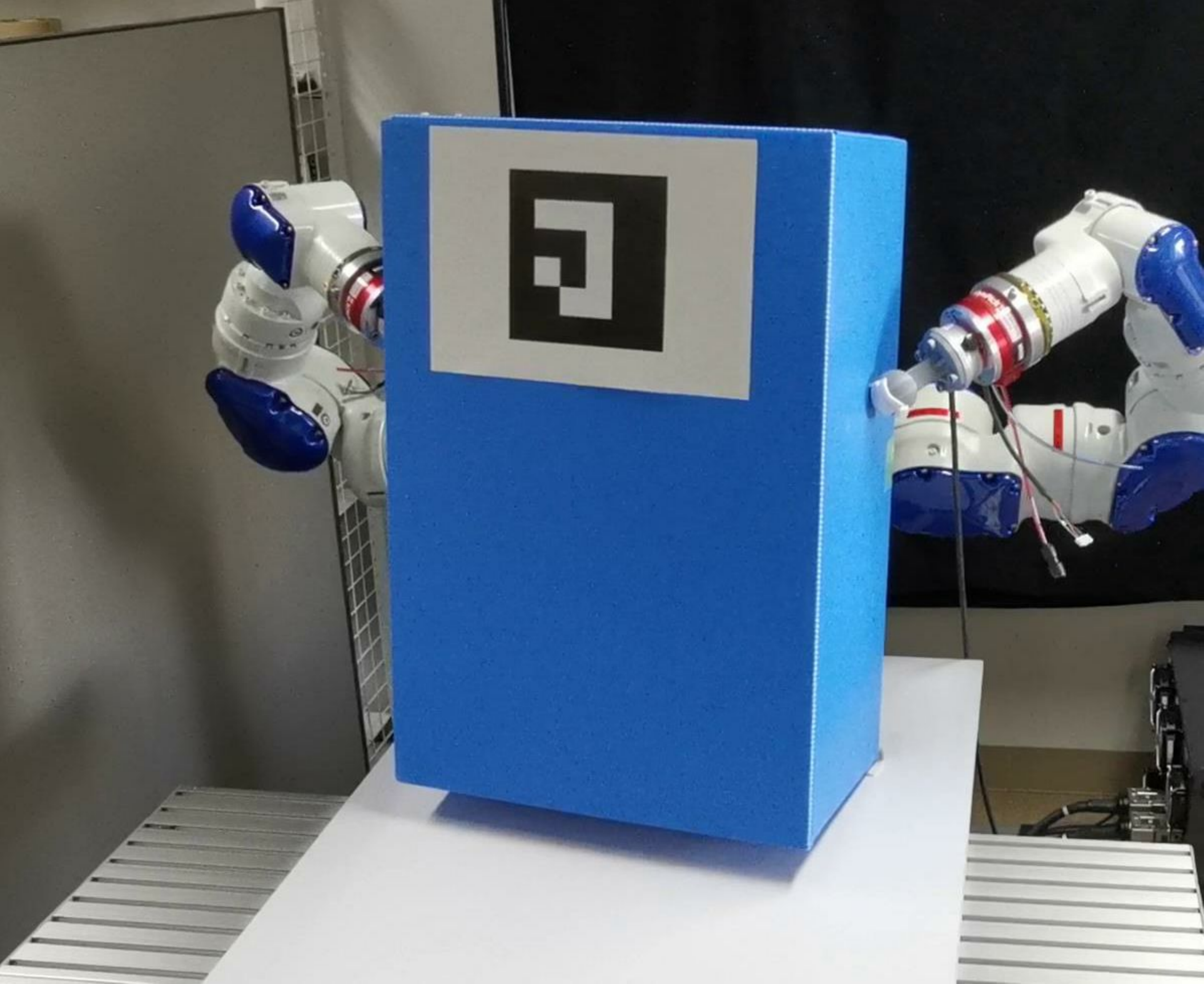}}
\subfigure[]{\label{ds5}\includegraphics[width=0.16\textwidth]{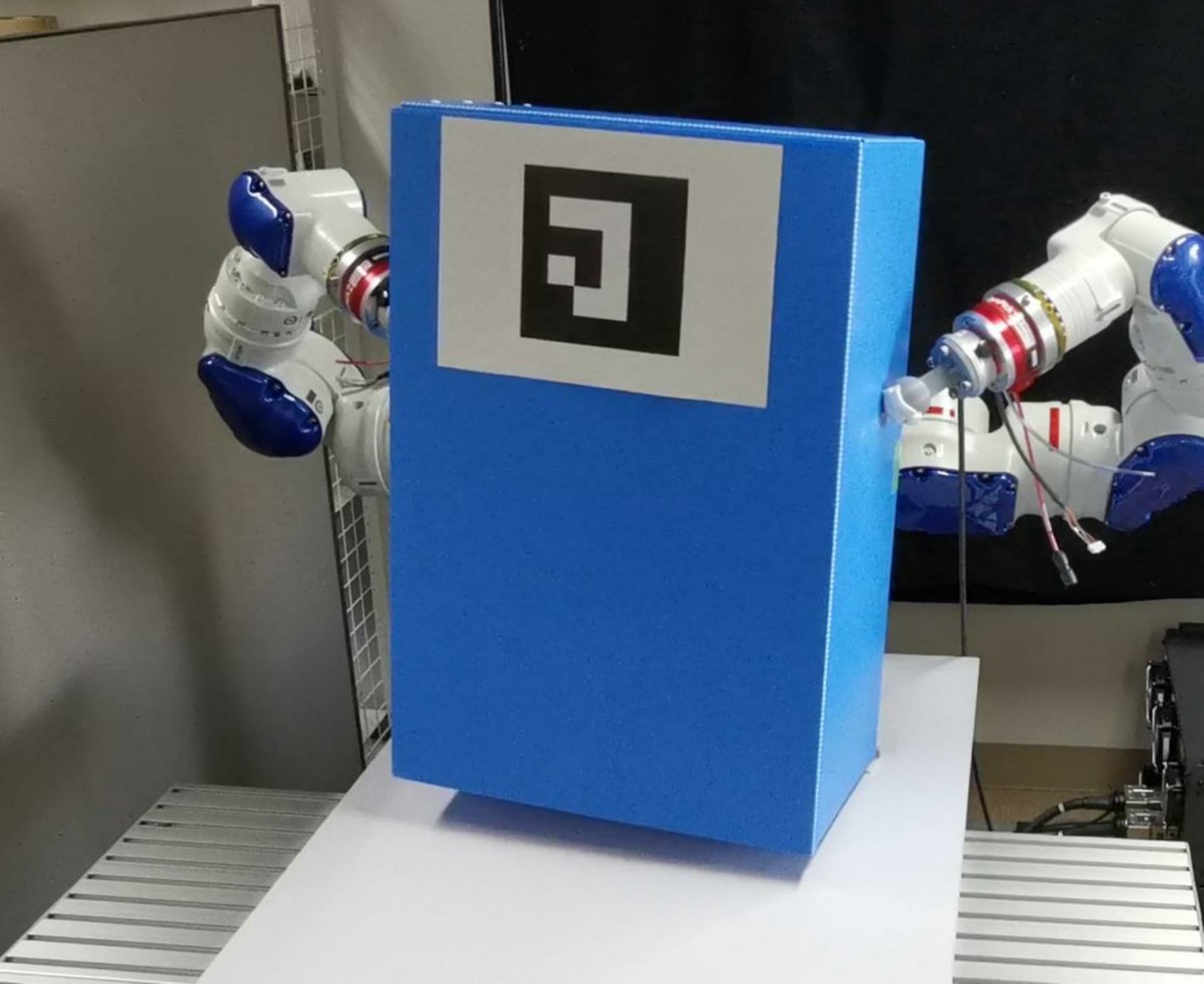}}
\subfigure[]{\label{ds6}\includegraphics[width=0.16\textwidth]{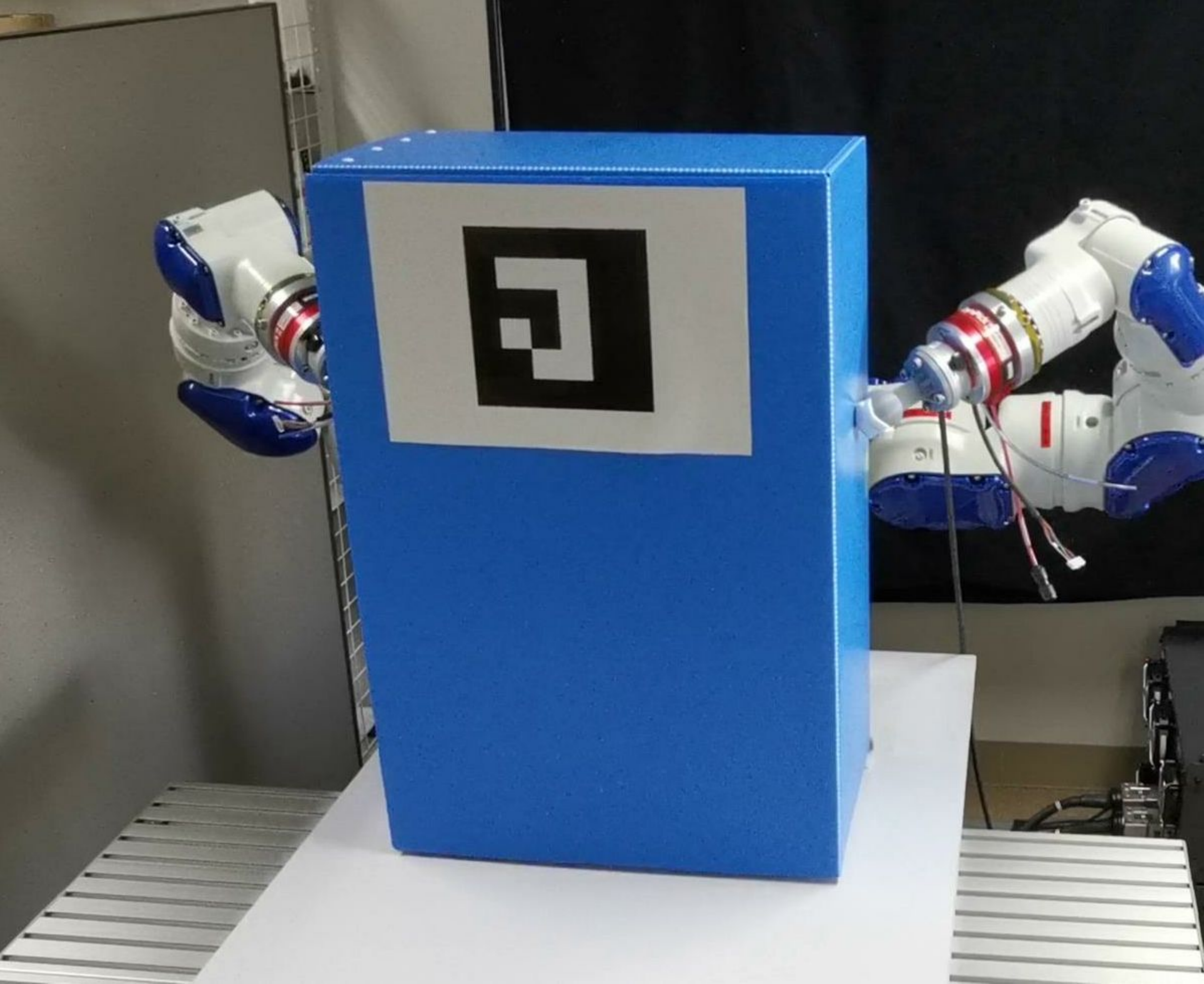}}
\caption{Robot pivots box to walk in the DS mode for two steps. }
\label{fig:ds2steps}
\end{figure*}

We evaluate the performance of the controller design in test scenarios through experiments. Two ball-shaped EEF are designed to keep a point contact between the robot and the box. Force sensors are installed on both sides of the robot's wrists. Impedance control\cite{hogan1984impedance,beltran2020learning} is implemented to control the force between robot's EEF and the box. The robot in use is Yaskawa Motoman SDA5F and the controller for the robot is FS100. In front of the robot, a box is placed on a table at a height of 80cm, see Fig.\ref{ds1}.

\begin{figure}
\centering
\includegraphics[width=0.45\textwidth]{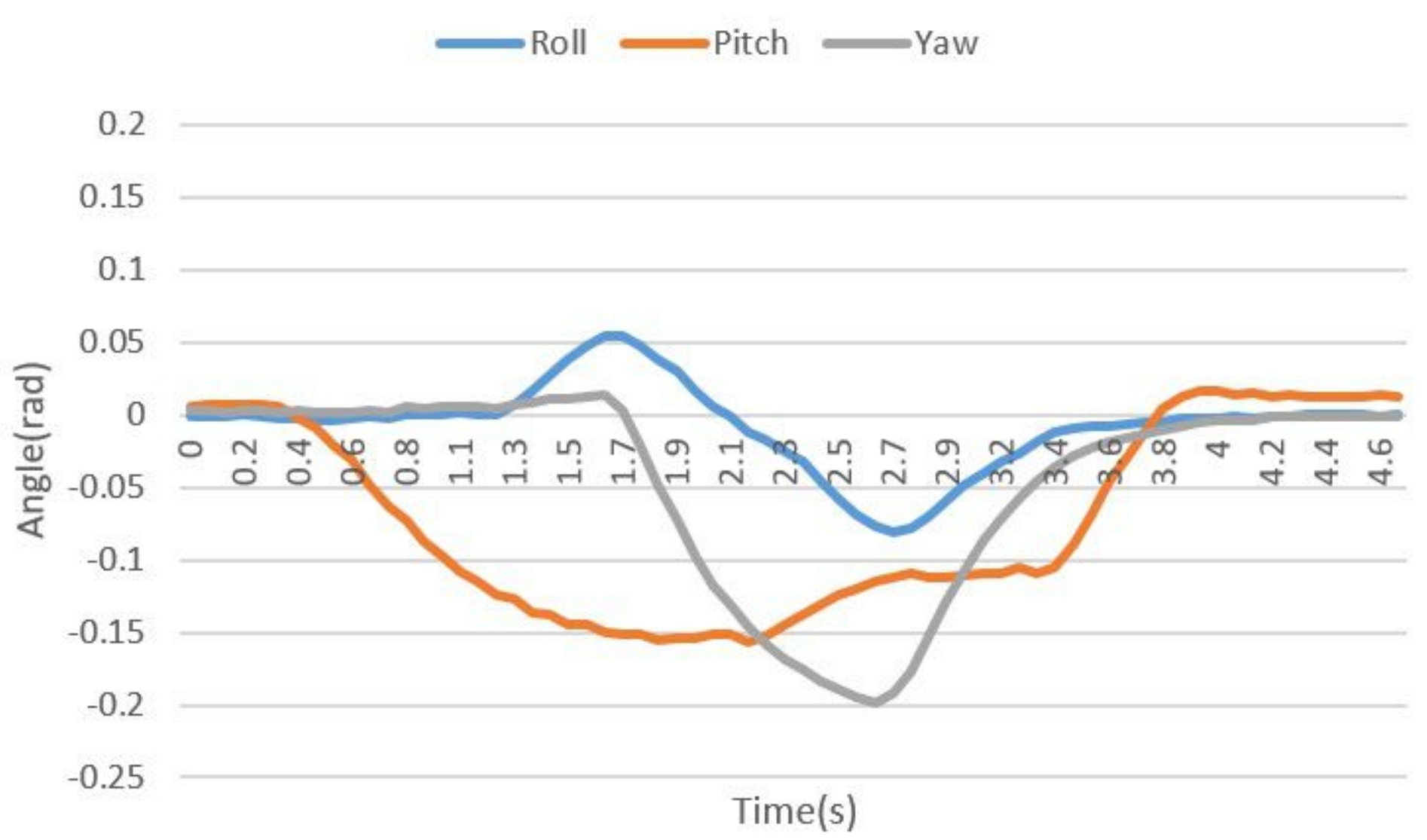}
\caption{Euler angles of the object which walks in the DS mode. At time 2.1s, only the Roll angle goes to zero which indicates the DS pose. }
\label{fig:dsCam}
\end{figure}

\begin{figure*}
\centering     
\subfigure[]{\label{qs1}\includegraphics[width=0.19\textwidth]{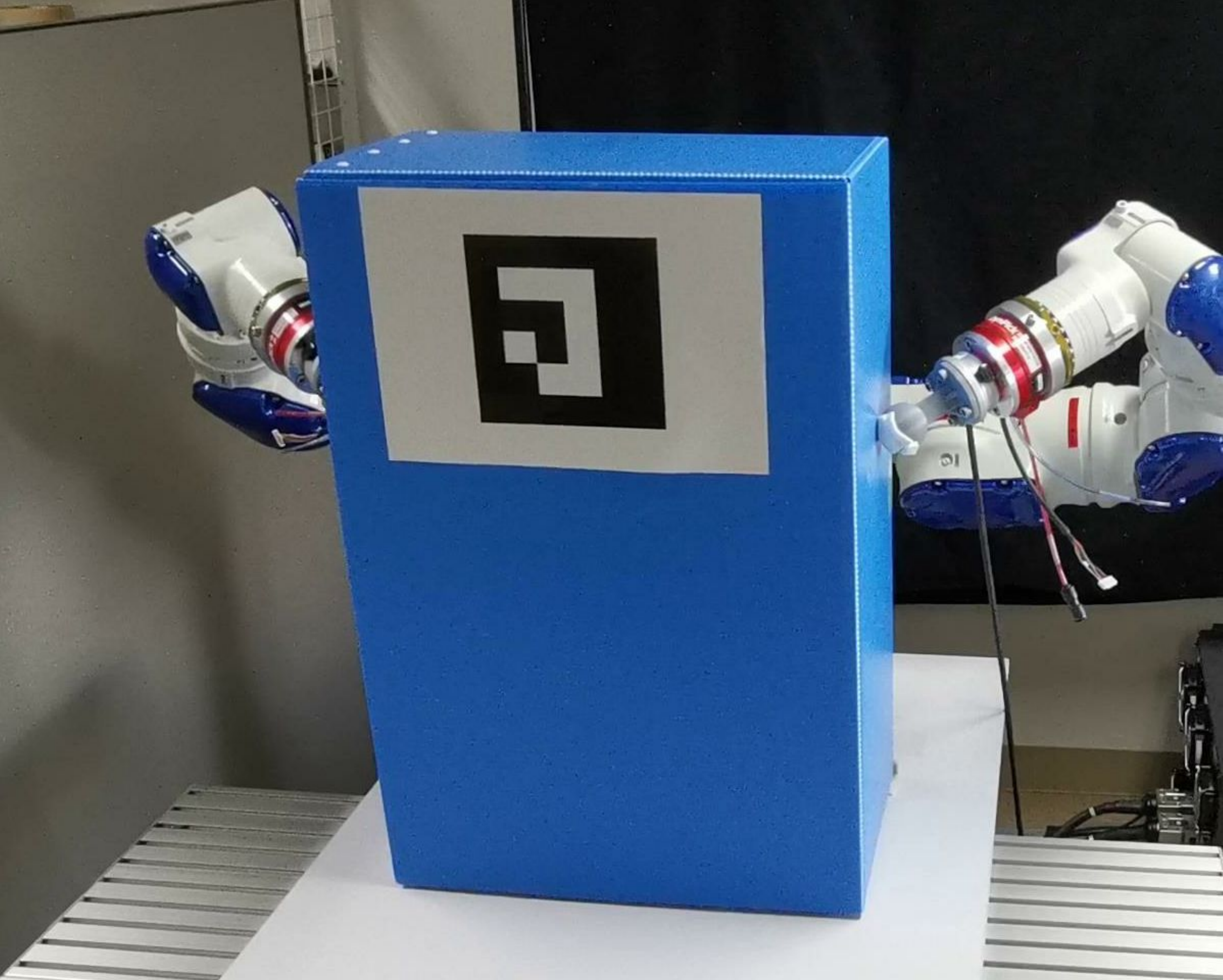}}
\subfigure[]{\label{qs2}\includegraphics[width=0.19\textwidth]{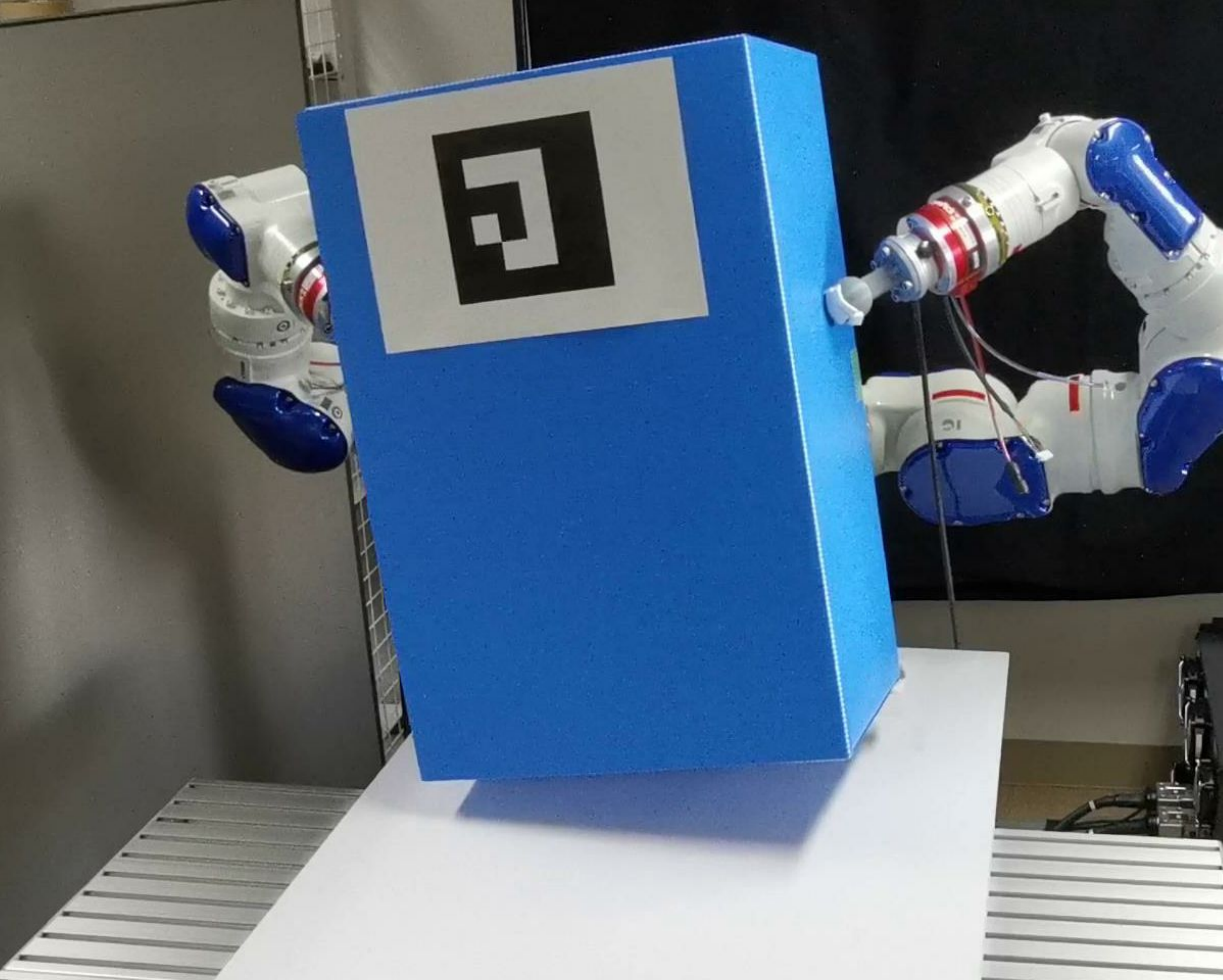}}
\subfigure[]{\label{qs3}\includegraphics[width=0.19\textwidth]{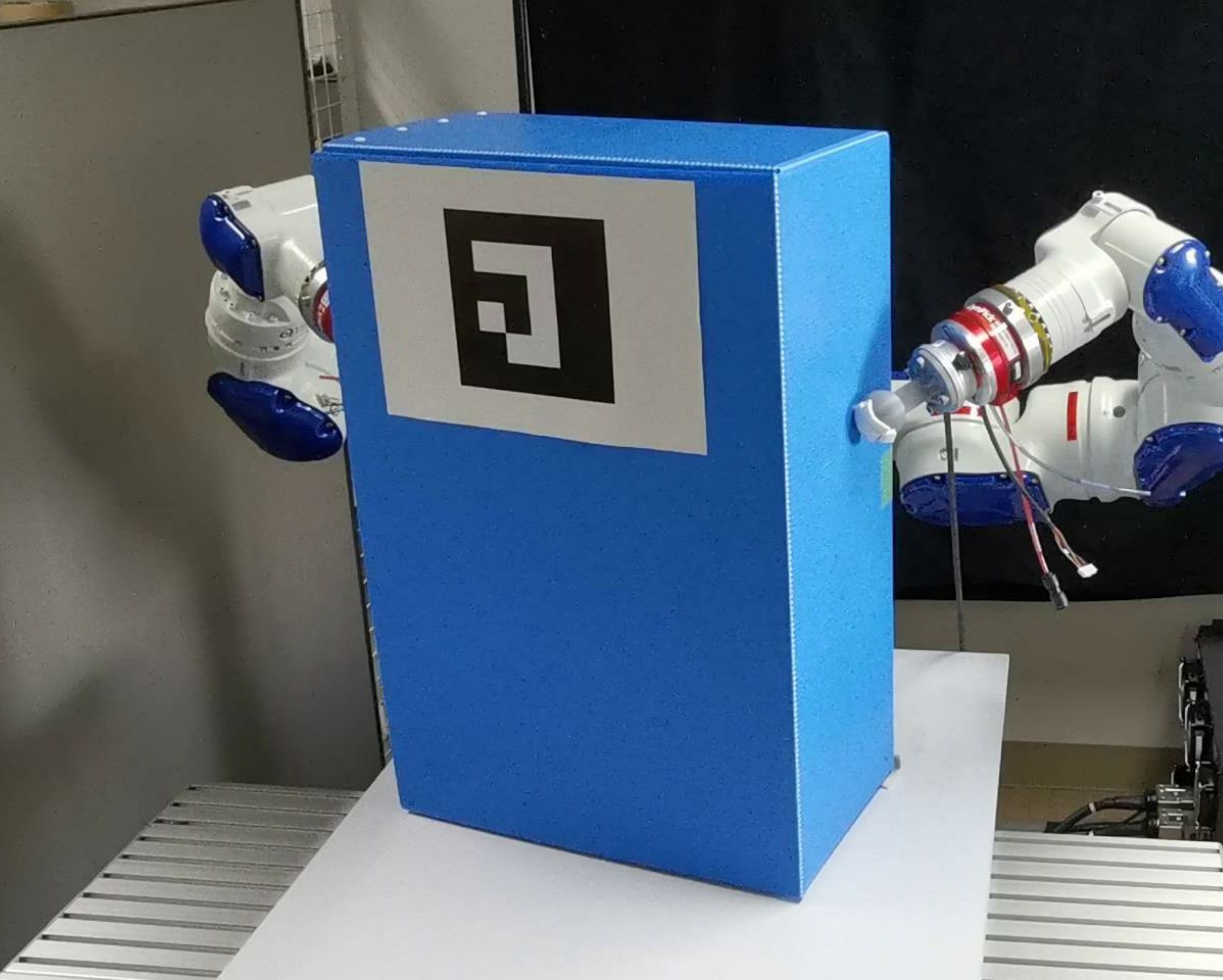}}
\subfigure[]{\label{qs4}\includegraphics[width=0.19\textwidth]{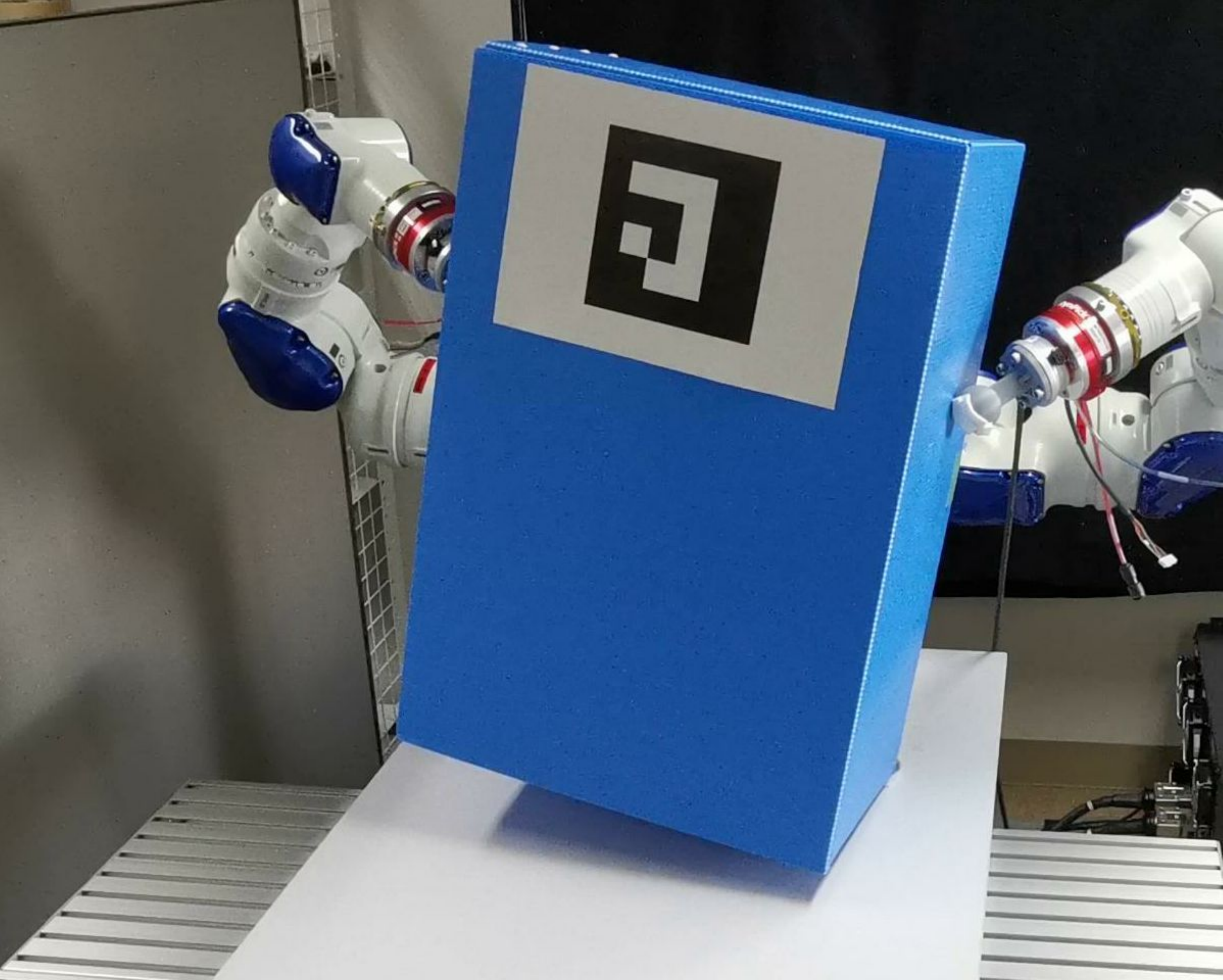}}
\subfigure[]{\label{qs5}\includegraphics[width=0.19\textwidth]{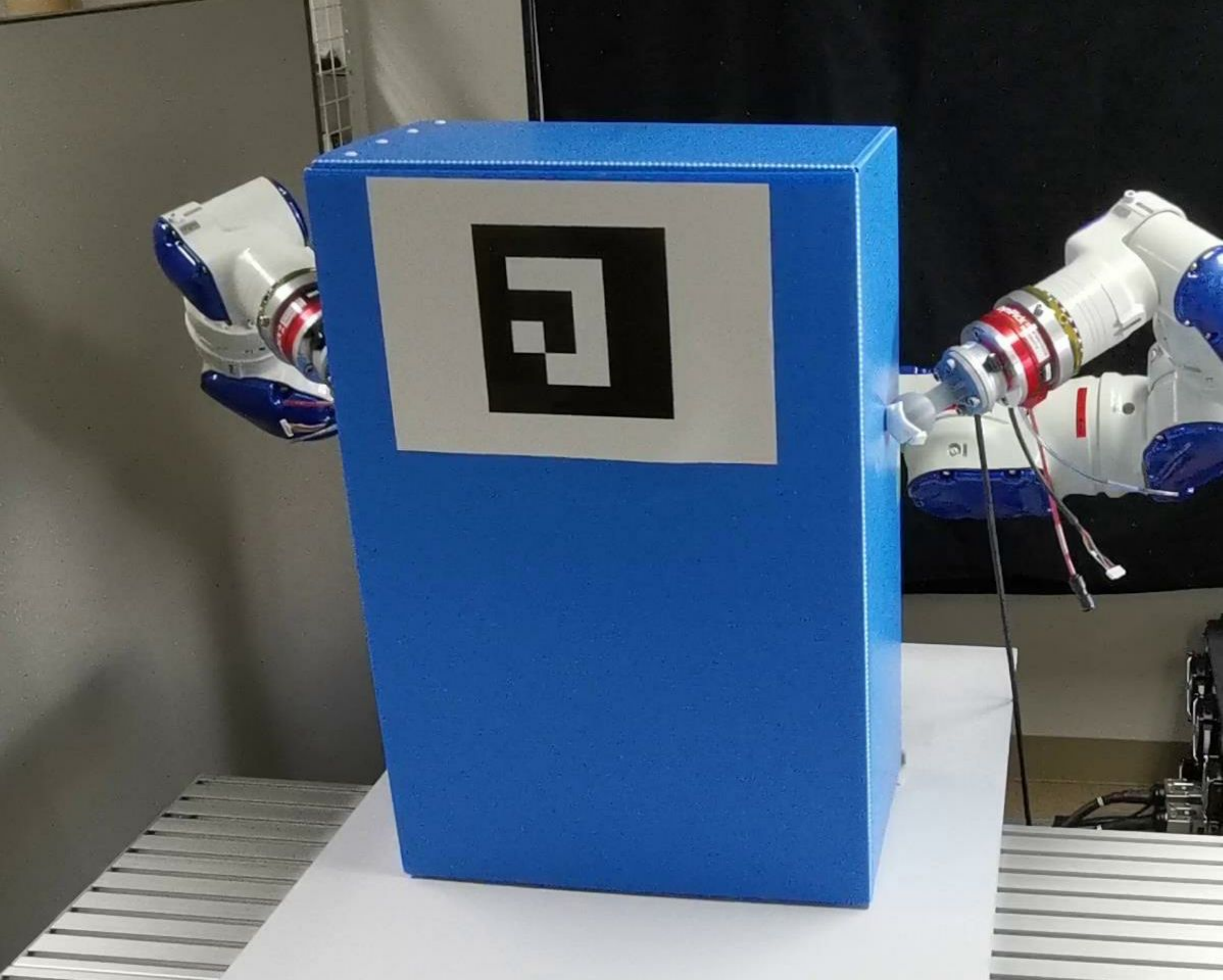}}
\caption{Robot pivots box to walk in the QS mode for two steps. }
\label{fig:qs2steps}
\end{figure*}

\hl{To find the characteristic of the gait modes, the first experiment is running pivoting gait in a single gait mode.} The step length in the DS mode (Fig.\ref{fig:ds2steps}) and that in the QS mode (Fig.\ref{fig:qs2steps}) are same. In the experiment of the DS mode, Fig.\ref{fig:dsCam} shows the Euler angles of the object during walking and Fig.\ref{fig:ds2steps} shows the motion. The object is firstly rotated from the initial pose (Fig.\ref{ds1}) to a DS pose (Fig.\ref{ds2} and at time 1.2s in Fig.\ref{fig:dsCam}). Then the object walks forward (Fig.\ref{ds3}), changes support foot from right to left (Fig.\ref{ds4} and at time 2.1s in Fig.\ref{fig:dsCam}), and finally goes to a rest pose in Fig.\ref{ds6}. 

\begin{figure}
\centering
\includegraphics[width=0.45\textwidth]{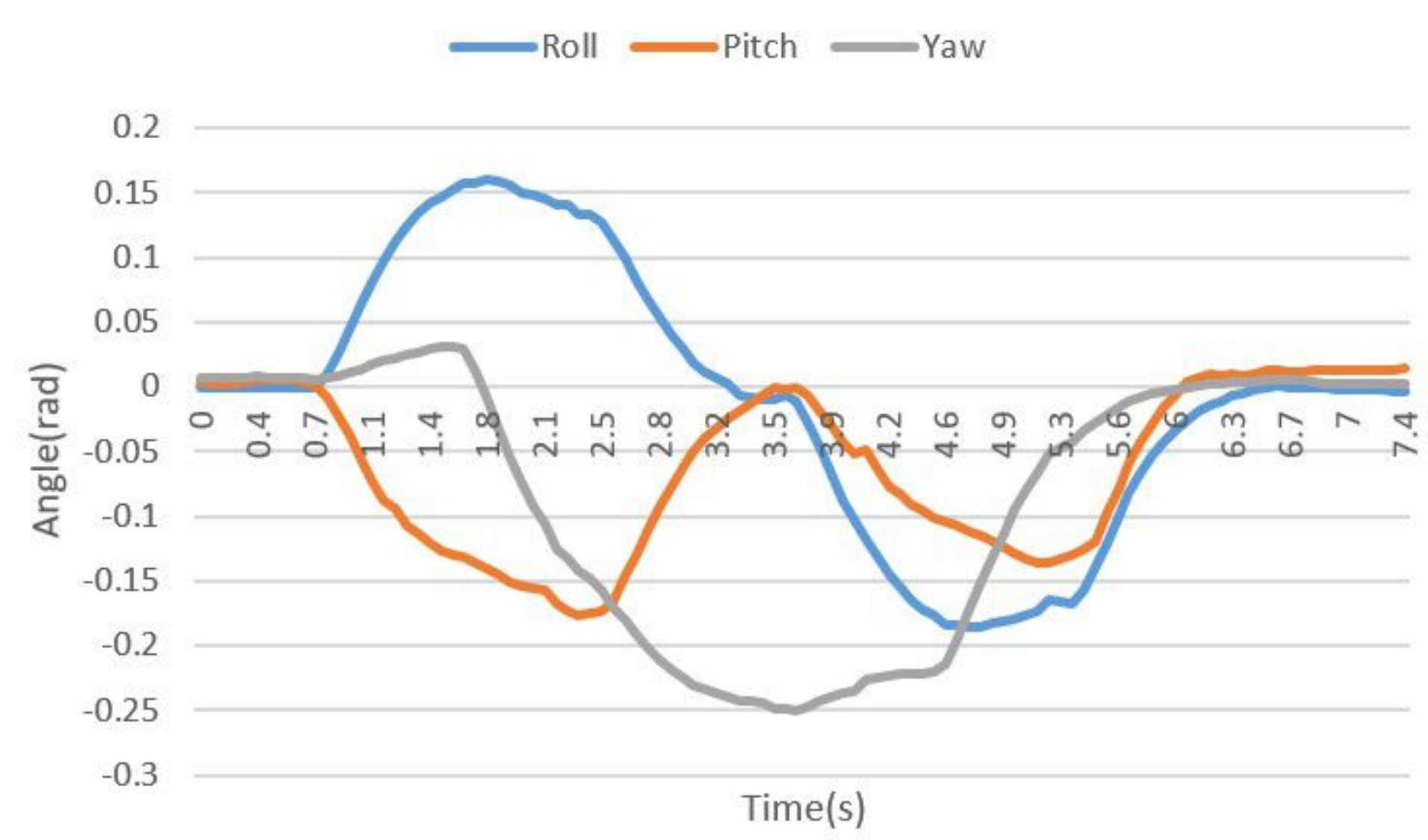}
\caption{Euler angles of the object which walks in the QS mode. At time 3.5s, both Roll and Pitch angles go to zero which indicates the QS pose.}
\label{fig:qsCam}
\end{figure}

In the experiment of the QS mode, see Fig.\ref{fig:qsCam} and Fig.\ref{fig:qs2steps}, robot starts to rotate the box around the right vertex and lift it to a SS pose (at time 2.1s in Fig.\ref{fig:qsCam} and Fig.\ref{qs2}). Then the box goes into a QS pose to change support foot (at time 3.6s in Fig.\ref{fig:qsCam} and Fig.\ref{qs3}) and moves to the target position (Fig.\ref{qs5}). 

In Table.\ref{table1}, the time for executing two QS steps takes 5.6s while the time for making two steps in DS takes 2.2s. A quicker motion of DS gait mode is achieved by selecting $t$ in the DS mode smaller than that in QS mode. The reasons to design a quicker motion of the DS gait mode are shown below: 
The QS pose requires all four bottom vertices are on the floor (see the zeros in value of roll and pitch at time 3.5s in Fig.\ref{fig:qsCam}) while in DS pose, two vertices are on the floor and two vertices are on the fly (see the values of roll and pitch at time 2.1s in Fig.\ref{fig:dsCam}). What's more, the peak value of roll angle in QS is bigger than that of DS which also slows down the motion of QS mode. The reason for designing a rather big value of roll angle in QS is that we leave some space in roll angle to absorb perturbation so that to avoid scuffing and walk stably. 
\begin{table}
\centering
\caption{Time for making two steps}
\label{table1}
\begin{tabular}{|p{35pt}|p{35pt}|p{50pt}|}
\hline
Mode & Time(s) & Step length(m) \\
\hline
DS & 2.2 & 0.085 \\
QS & 5.6 & 0.085 \\
\hline
\end{tabular}
\end{table}

\hl{In experiment 1, motions of pivoting gait based on both the DS and QS gait mode are achieved. Pivoting gait in the DS mode is faster than that of the QS mode. Though the QS mode is slow, we believe it has a potential to keep stable walking and test the potential by following experiments.}  

\subsection{Experiment 2: Uncertainty in object's mass}
\hl{Though the MPC is inherently able to cope with external disturbance, we believe selecting proper gait mode improves the robustness of the controller. We design experiments of pivot gait with uncertainty in the object's mass to check:}
\begin{itemize}
 \setlength{\parskip}{0cm} 
 \setlength{\itemsep}{0cm} 
  \item \hl{Performances of the two designed gait modes when there is disturbance.}
  \item \hl{If switching of gait mode improves the robustness of the system.}
\end{itemize}

\hl{While the weight of the box is 1.4kg, bottles in different weights are put on the box during the walking process. In experiments, the weights of bottles are 0.35kg, 0.5kg, 1kg, and 2kg, respectively.} 

\begin{figure*}
\centering     
\subfigure[]{\label{350g1}\includegraphics[width=0.16\textwidth]{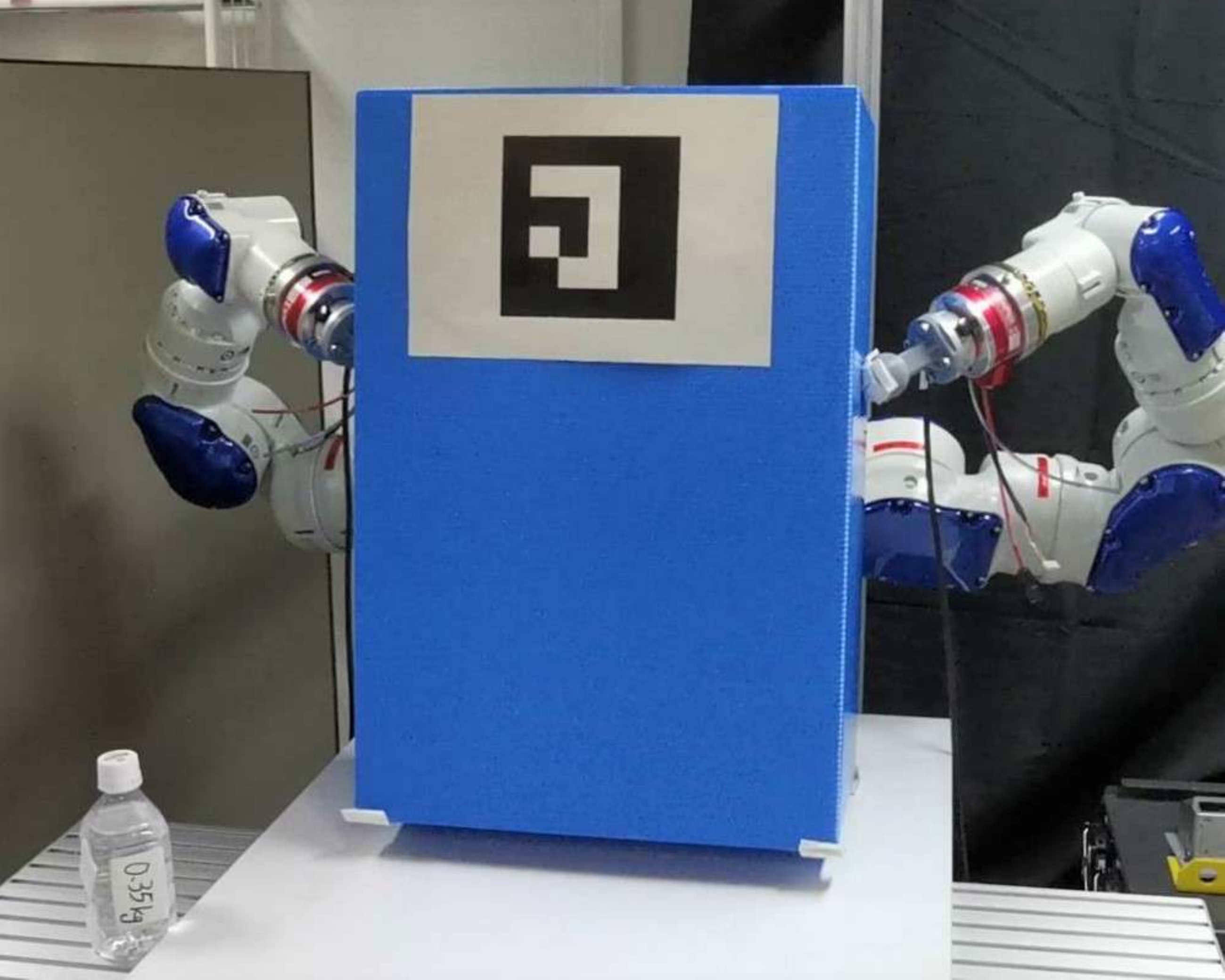}}
\subfigure[]{\label{350g2}\includegraphics[width=0.16\textwidth]{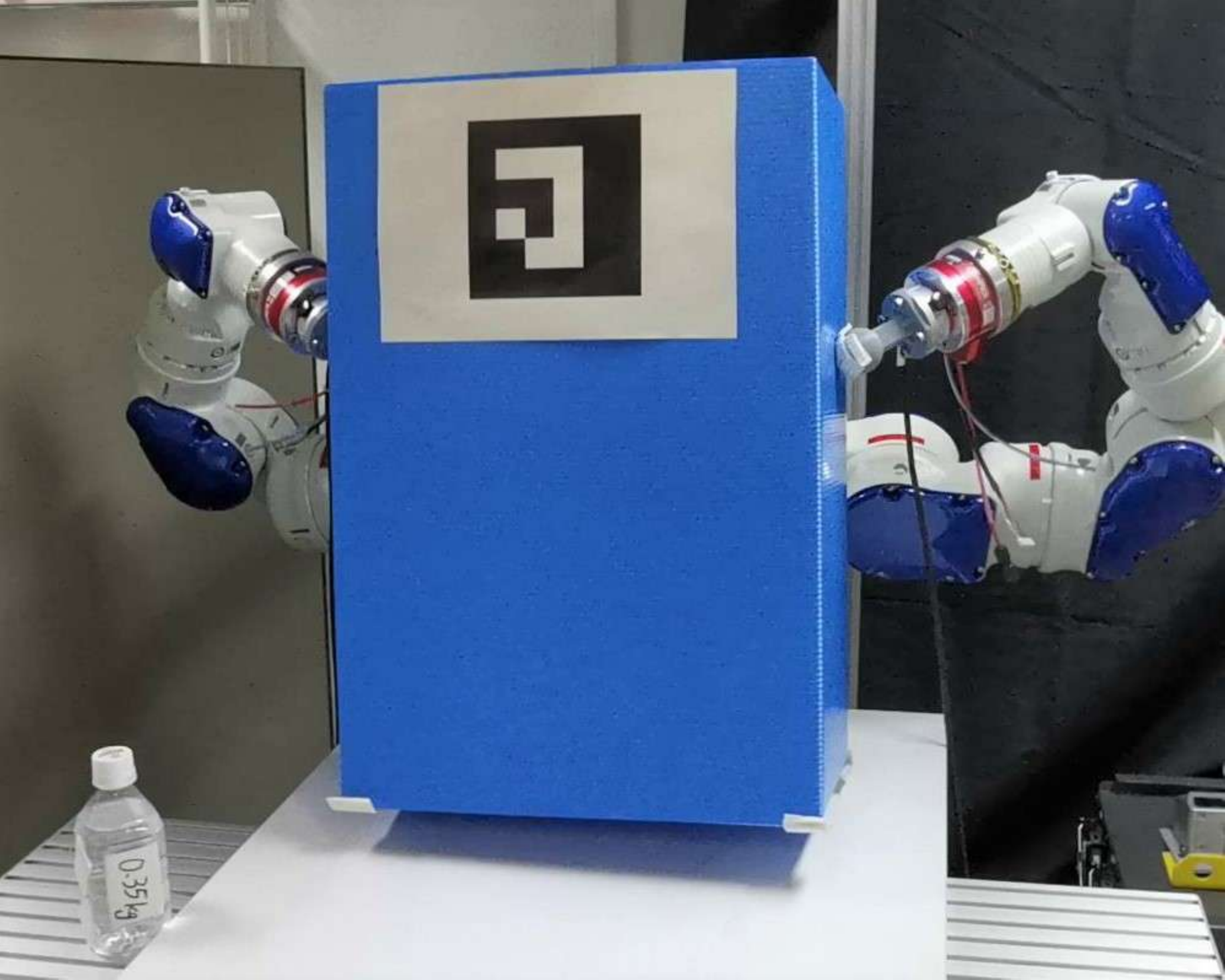}}
\subfigure[]{\label{350g3}\includegraphics[width=0.16\textwidth]{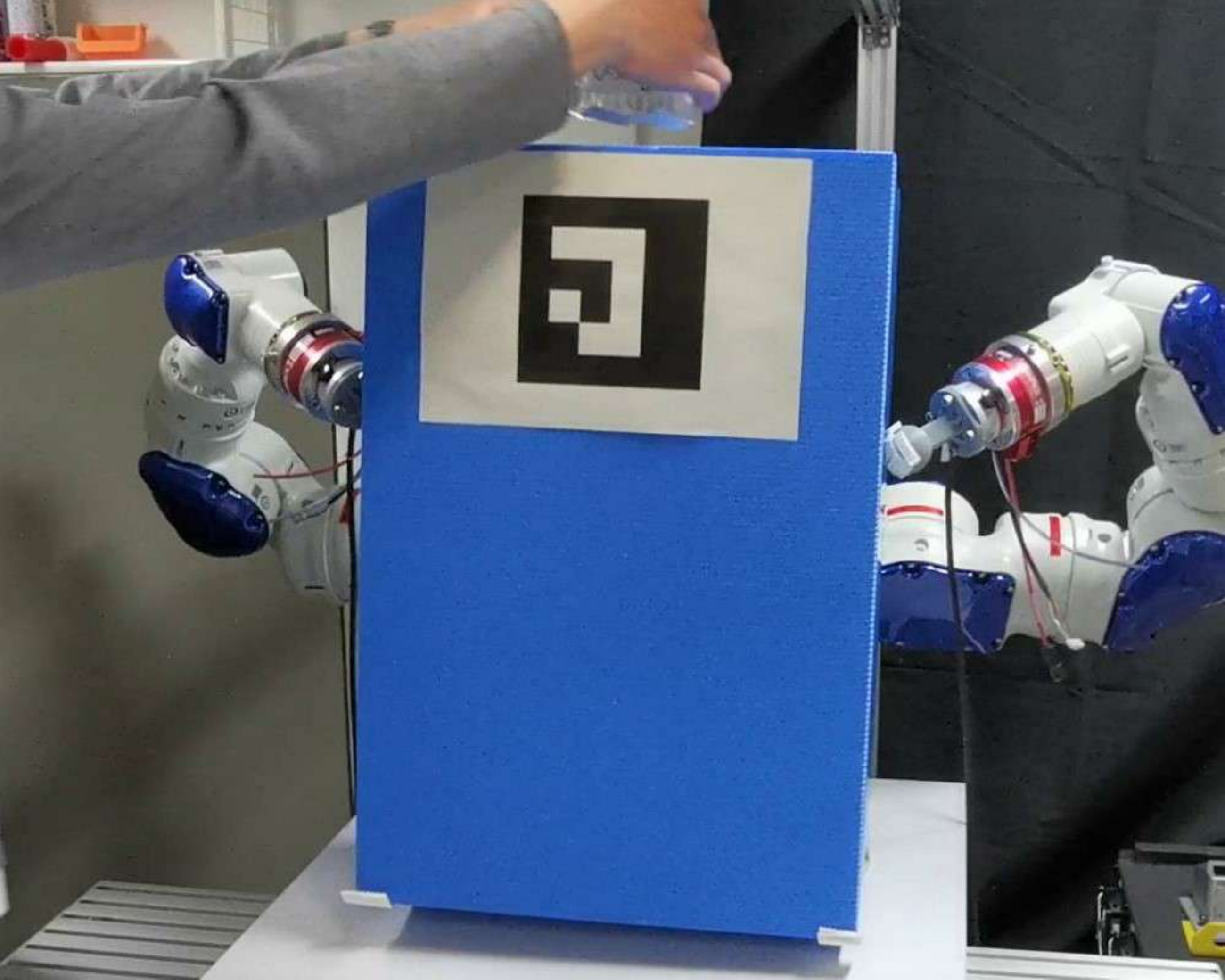}}
\subfigure[]{\label{350g4}\includegraphics[width=0.16\textwidth]{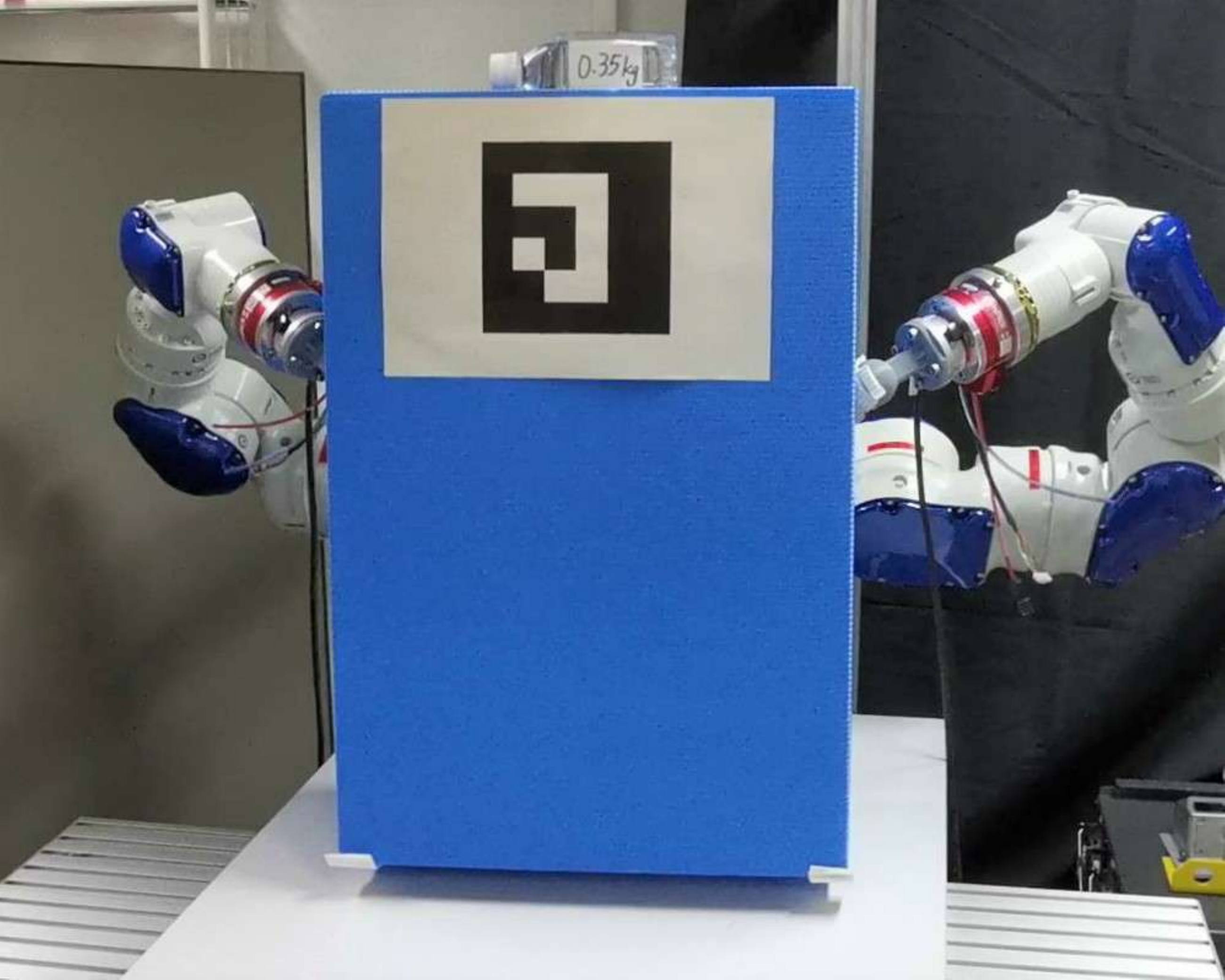}}
\subfigure[]{\label{350g5}\includegraphics[width=0.16\textwidth]{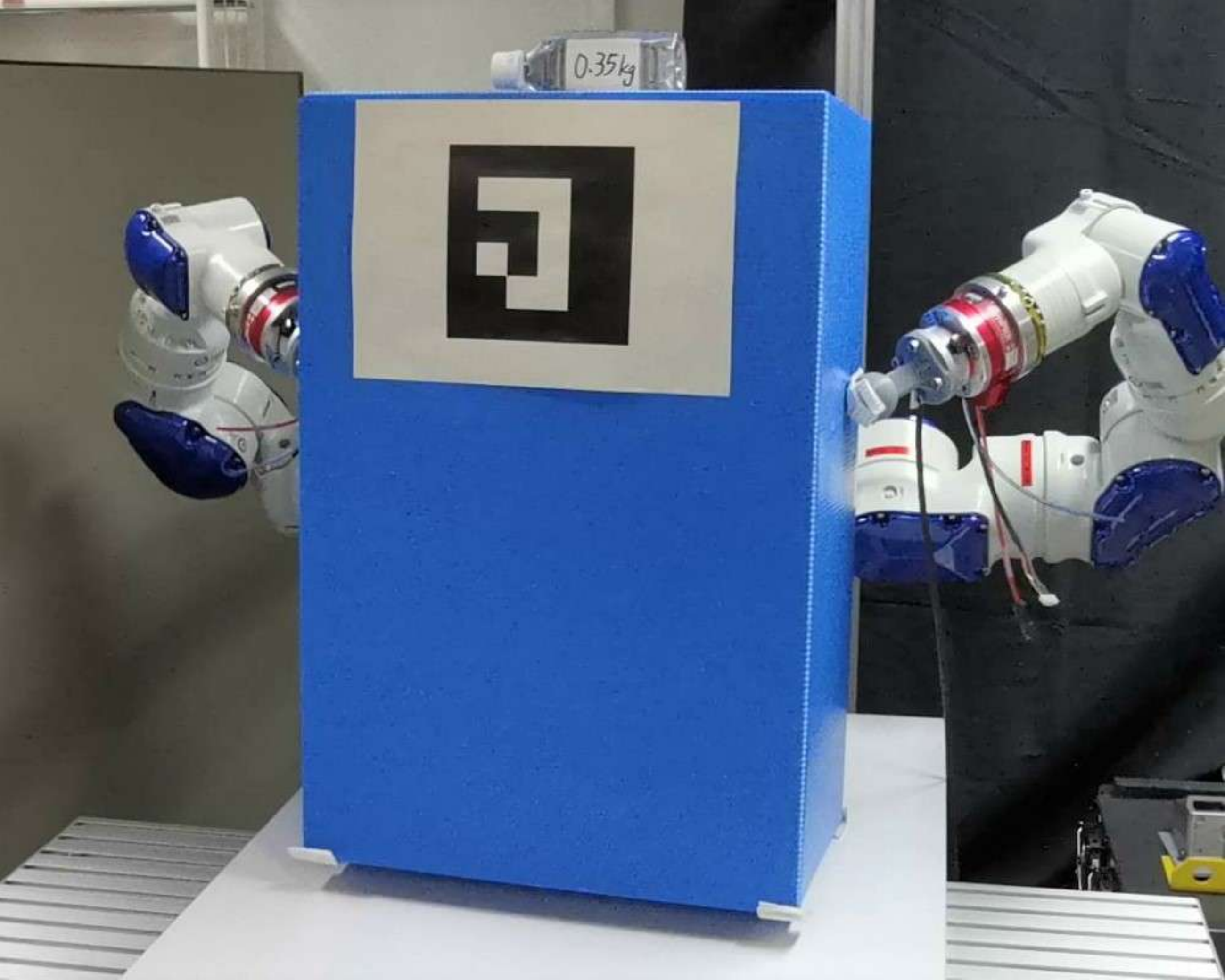}}
\subfigure[]{\label{350g6}\includegraphics[width=0.16\textwidth]{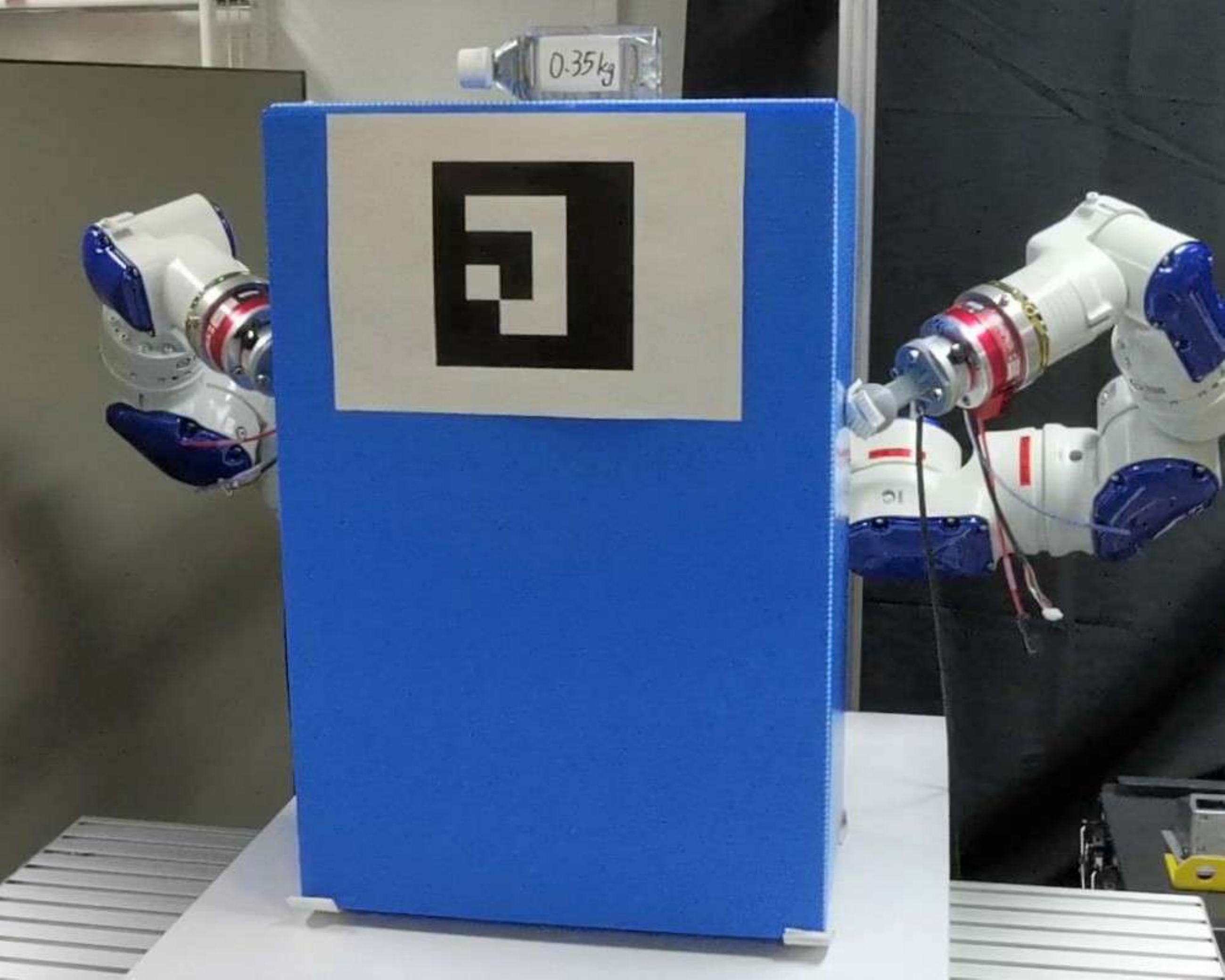}}
\caption{Experiment of placing a 0.35kg bottle on the manipulated object during the motion of pivoting, the robot successfully pivots the object in the DS mode.}
\label{fig:bottle350g}
\end{figure*}

\begin{figure*}
\centering     
\subfigure[]{\label{500g1}\includegraphics[width=0.16\textwidth]{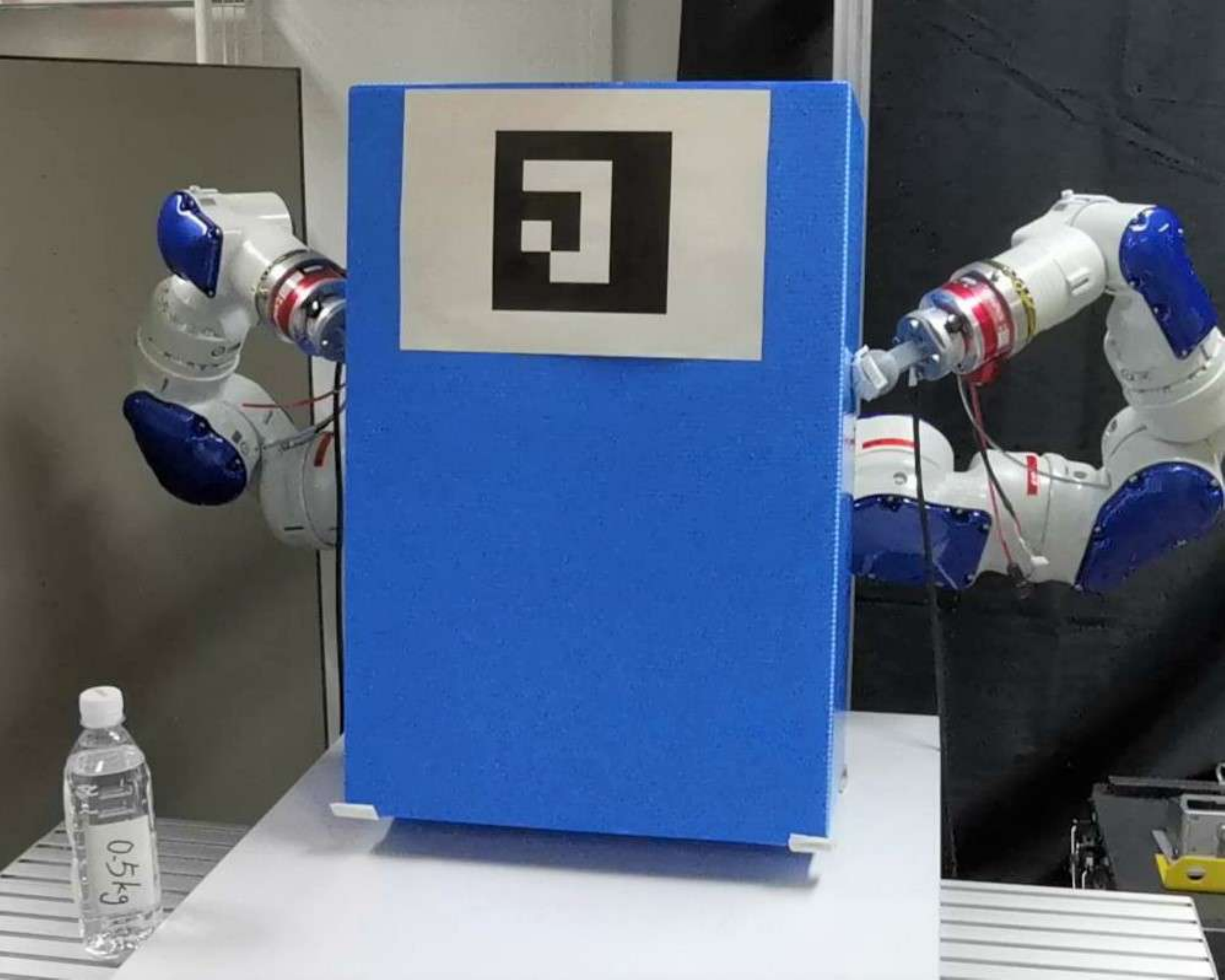}}
\subfigure[]{\label{500g2}\includegraphics[width=0.16\textwidth]{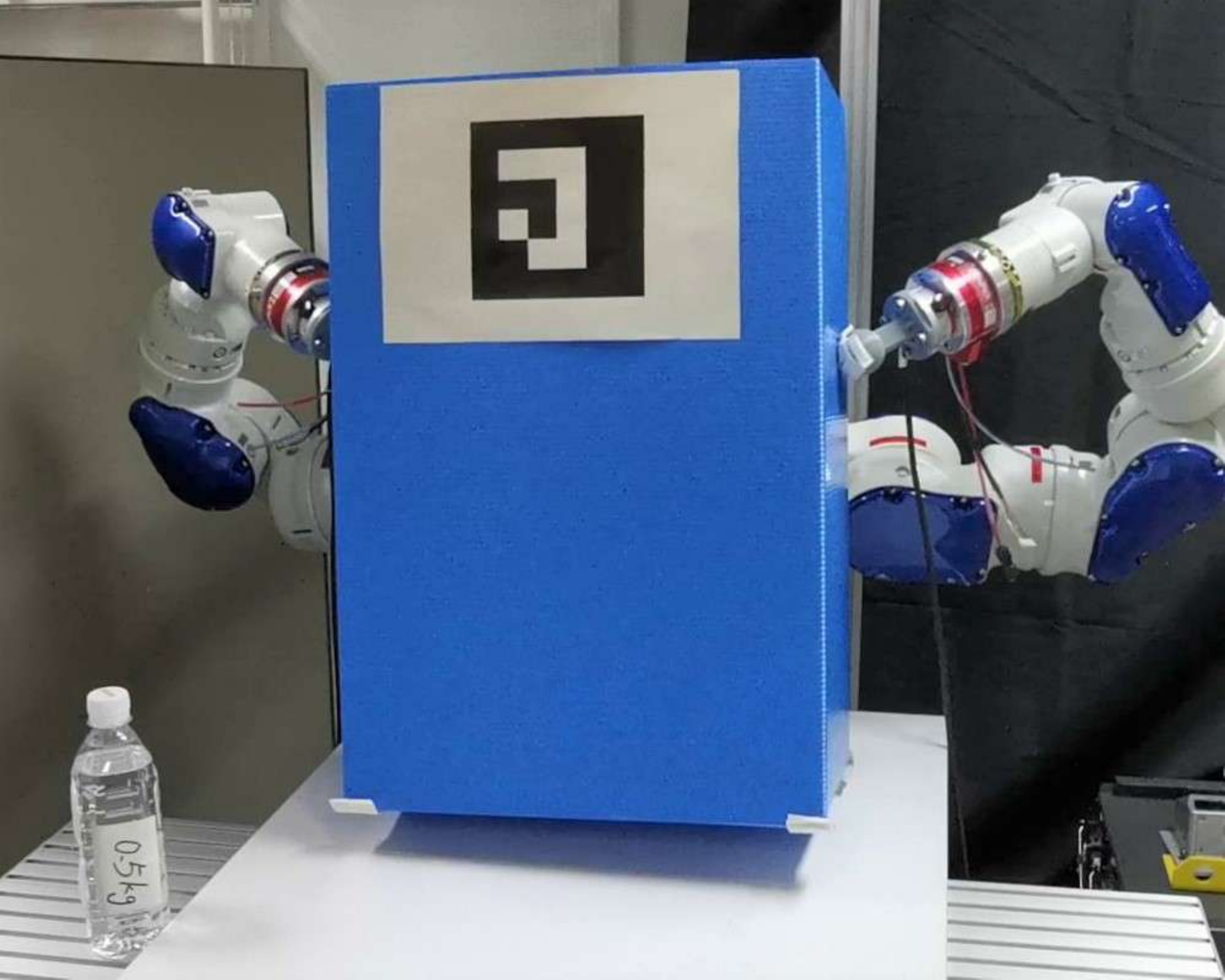}}
\subfigure[]{\label{500g3}\includegraphics[width=0.16\textwidth]{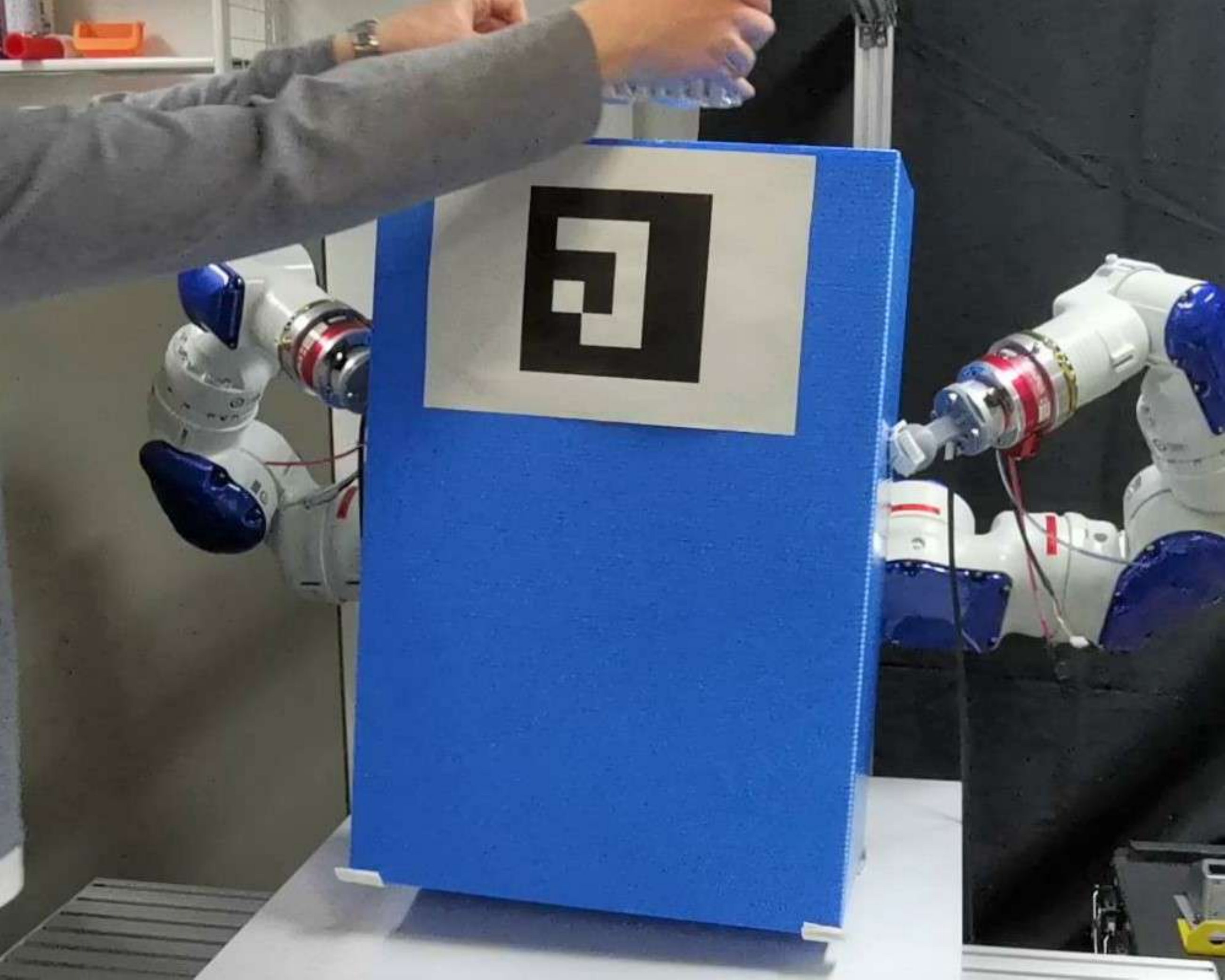}}
\subfigure[]{\label{500g4}\includegraphics[width=0.16\textwidth]{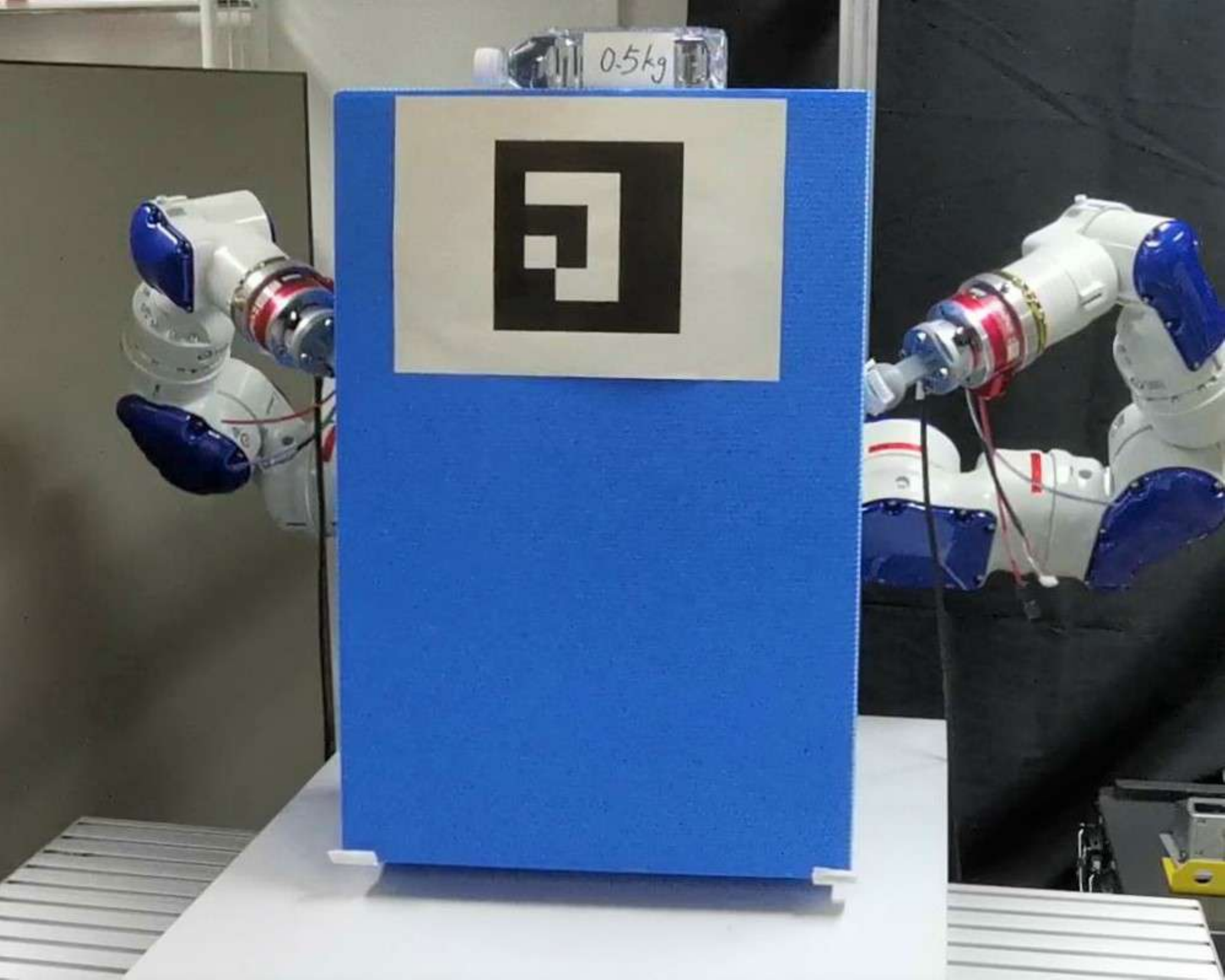}}
\subfigure[]{\label{500g5}\includegraphics[width=0.16\textwidth]{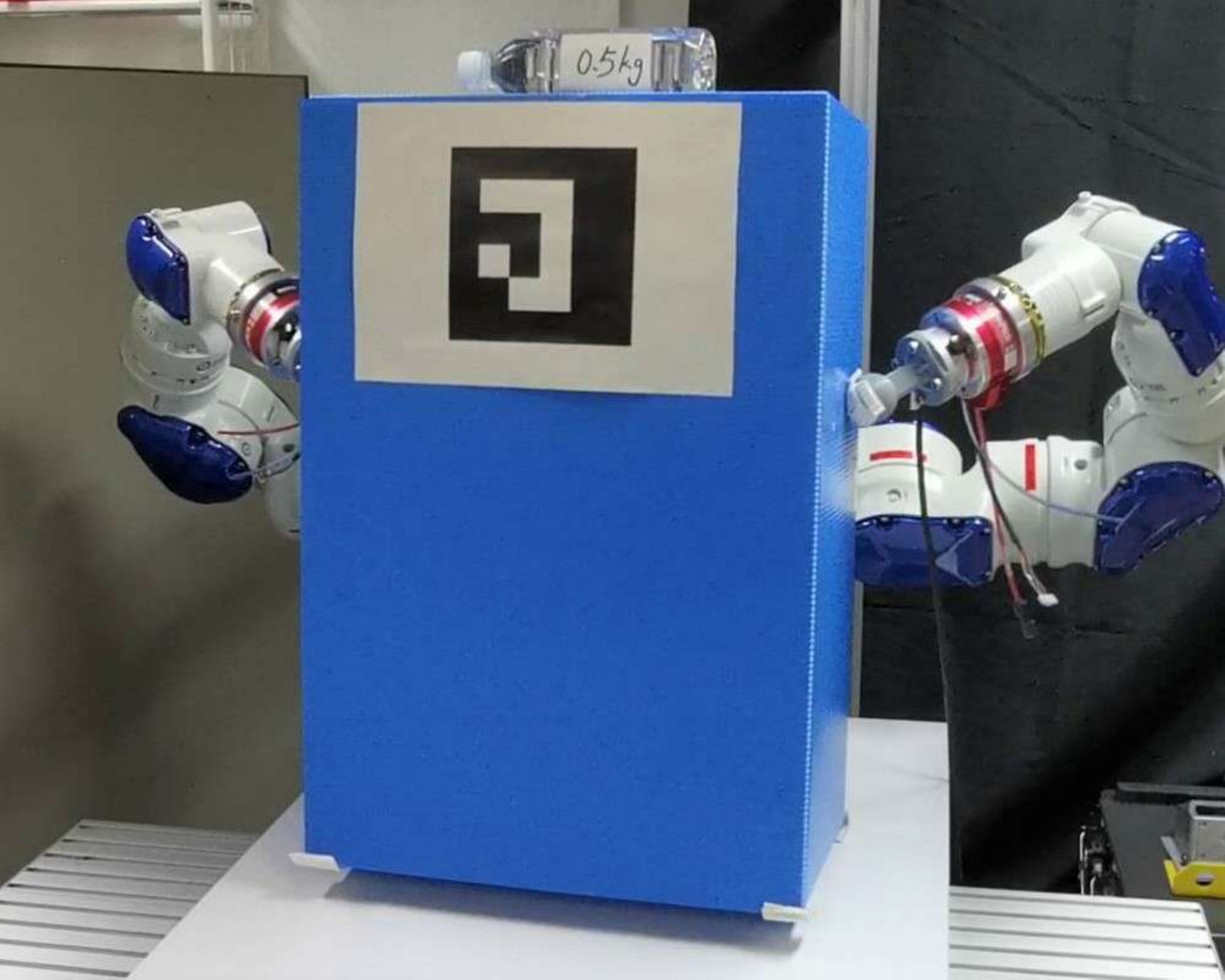}}
\subfigure[]{\label{500g6}\includegraphics[width=0.16\textwidth]{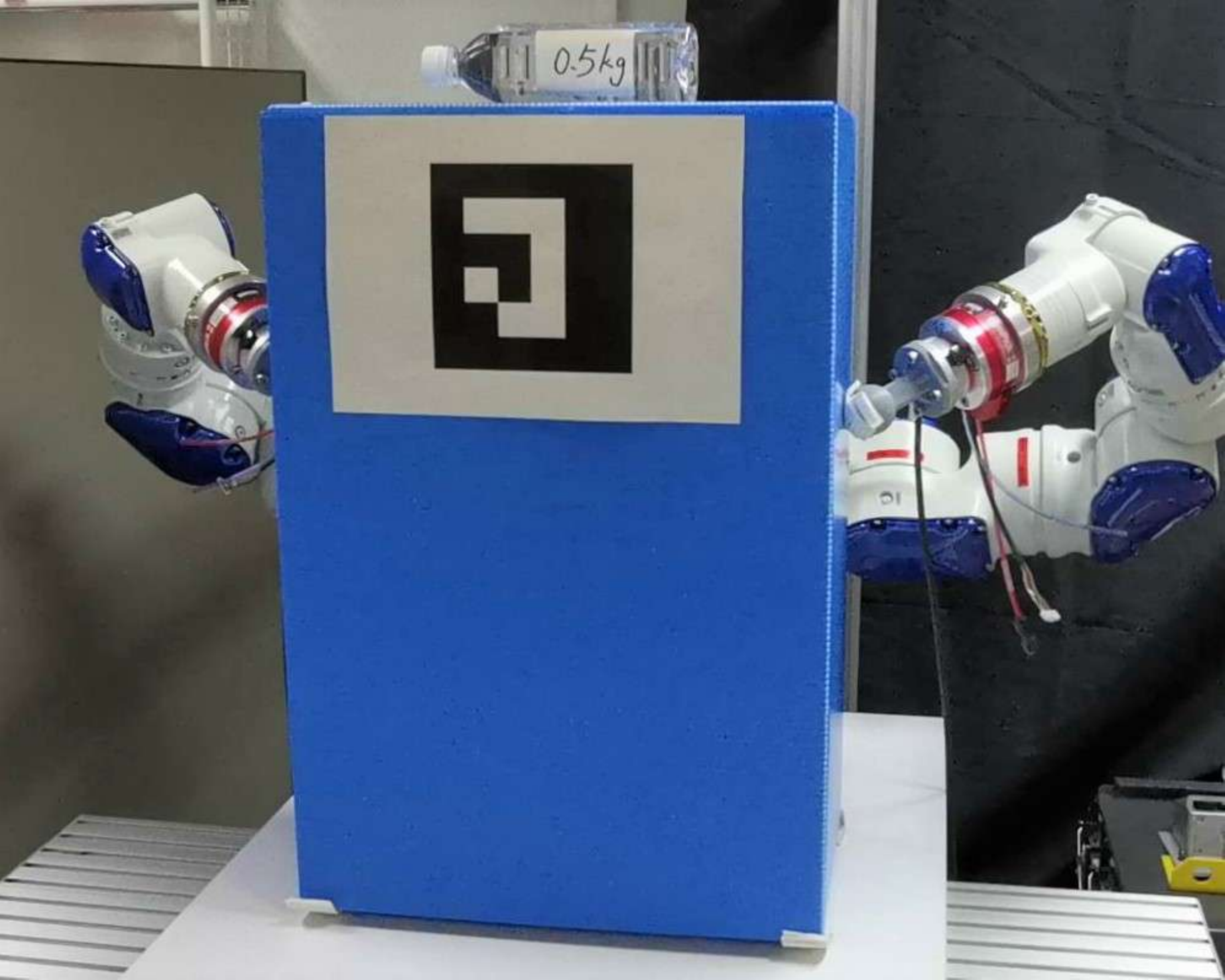}}
\caption{Experiment of placing a 0.5kg bottle on the manipulated object during the motion of pivoting, the robot successfully pivots the object in the DS mode.}
\label{fig:bottle500g}
\end{figure*}

\begin{figure*}
\centering     
\subfigure[]{\label{1k1}\includegraphics[width=0.19\textwidth]{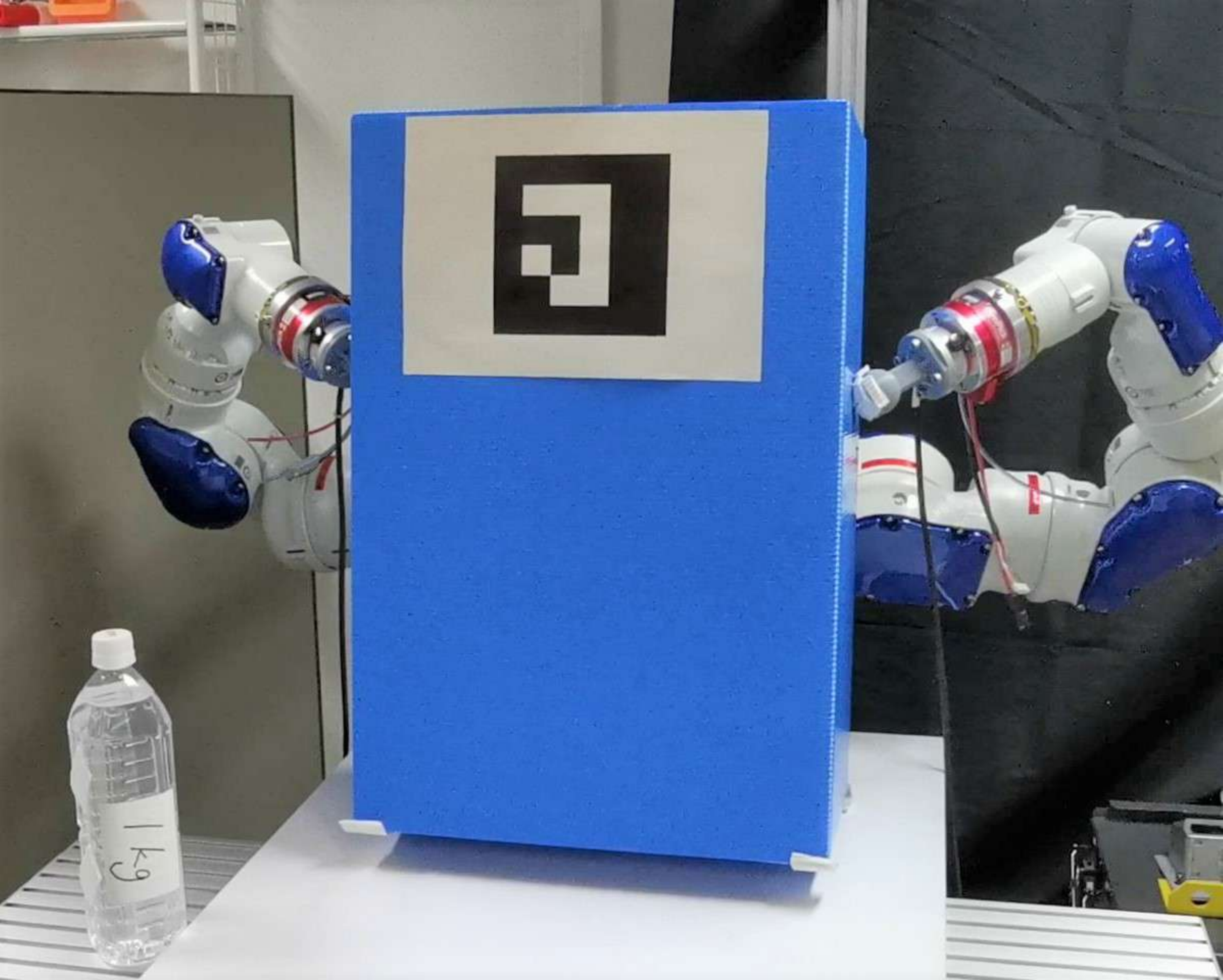}}
\subfigure[]{\label{1k2}\includegraphics[width=0.19\textwidth]{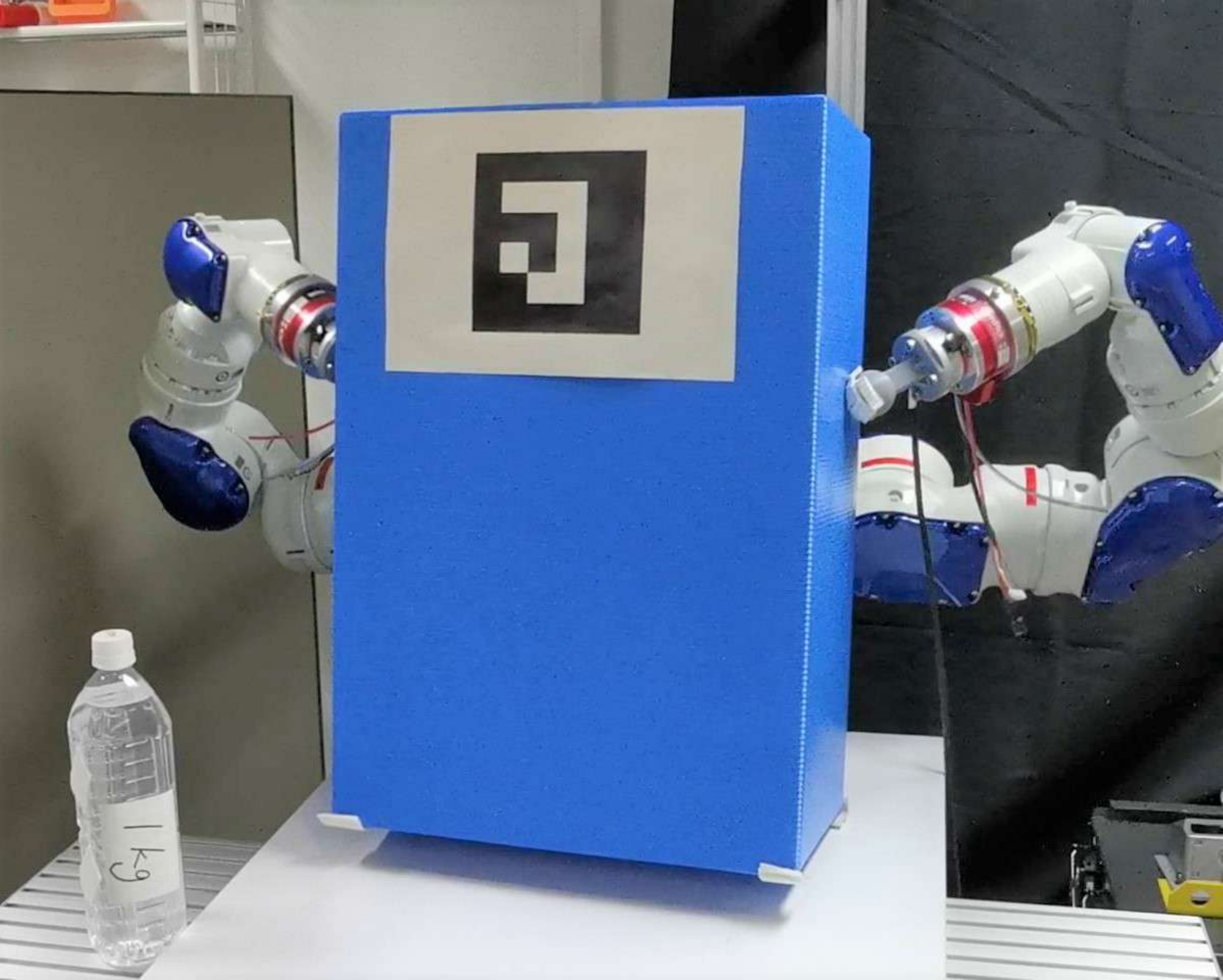}}
\subfigure[]{\label{1k3}\includegraphics[width=0.19\textwidth]{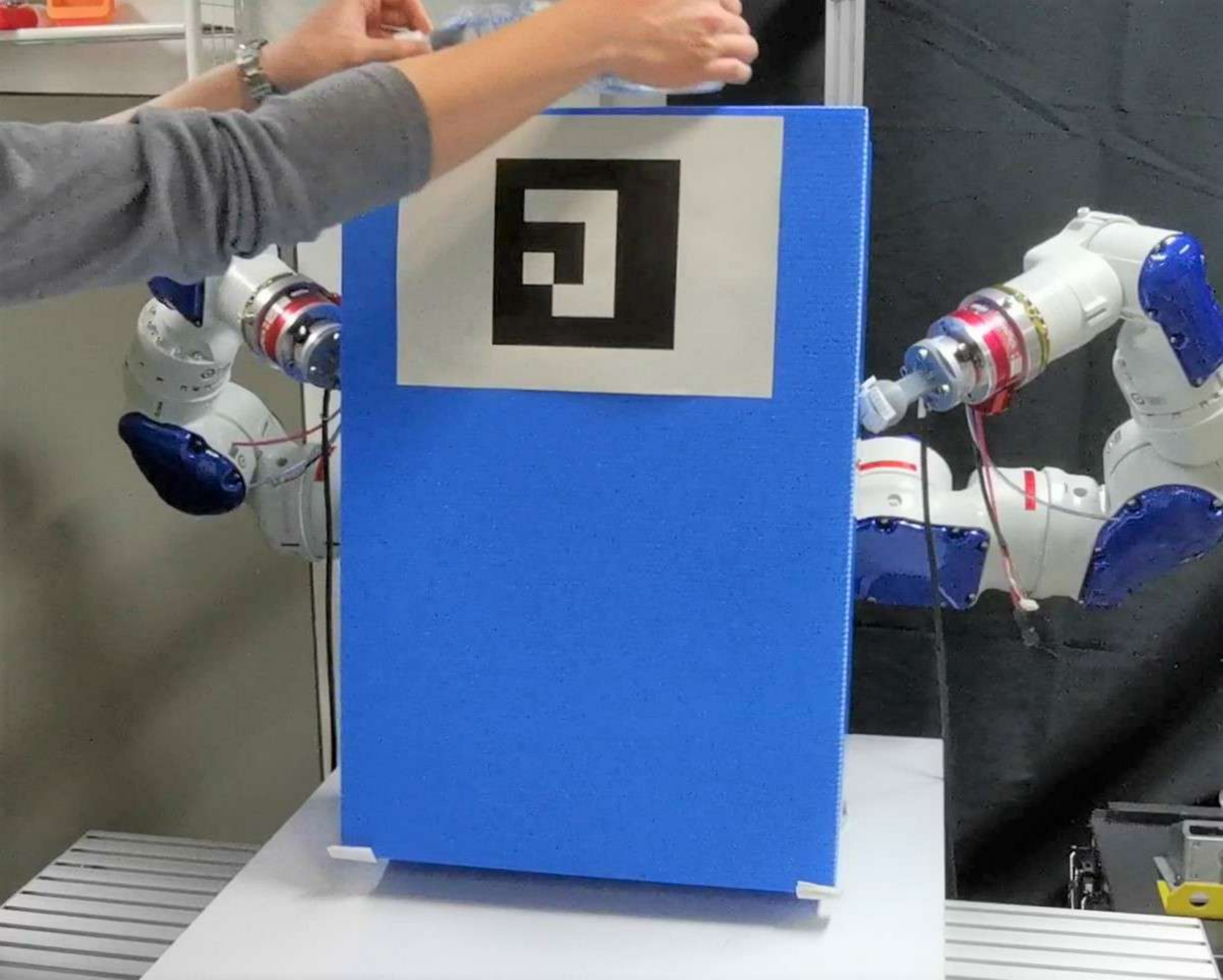}}
\subfigure[]{\label{1k4}\includegraphics[width=0.19\textwidth]{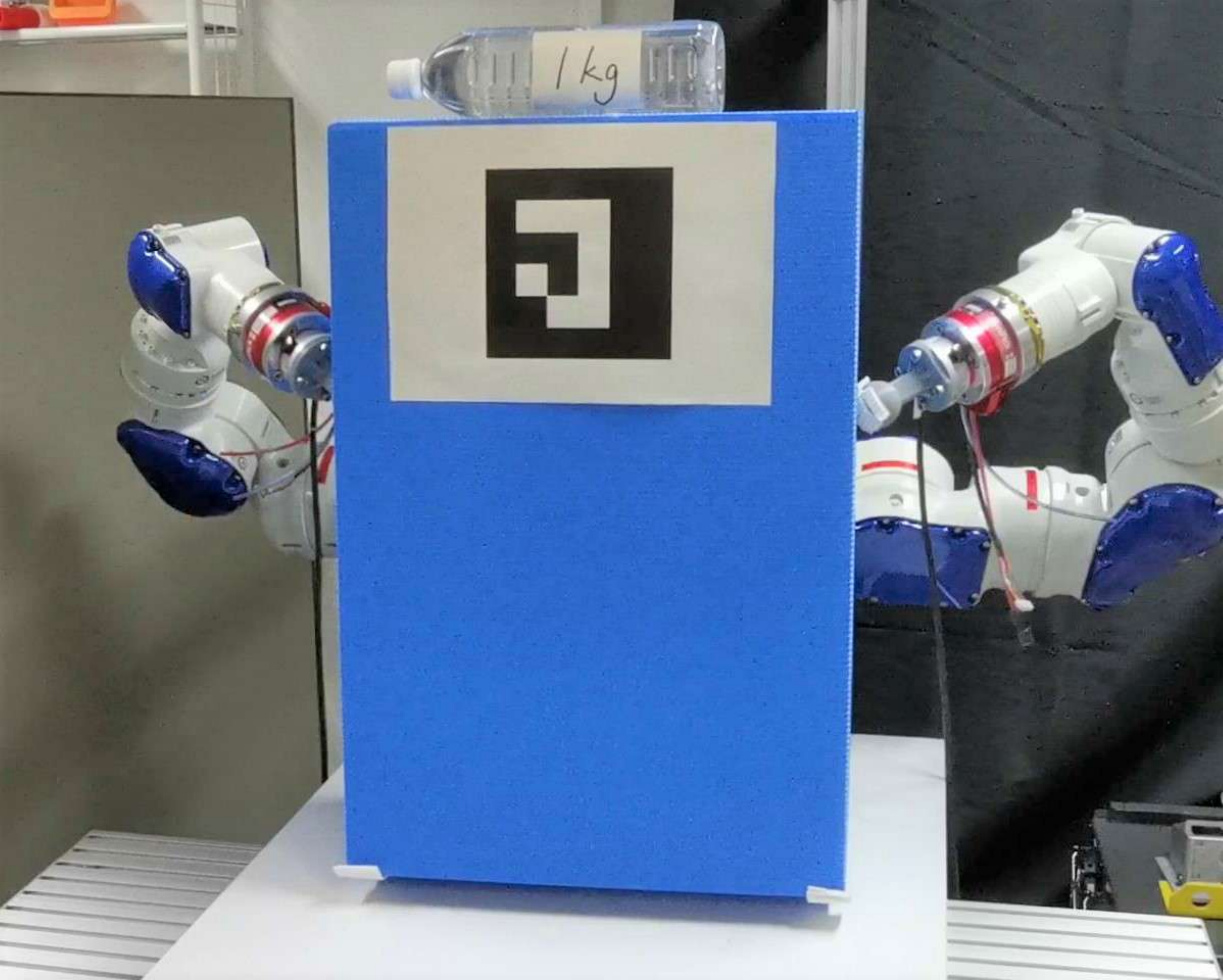}}
\subfigure[]{\label{1k5}\includegraphics[width=0.19\textwidth]{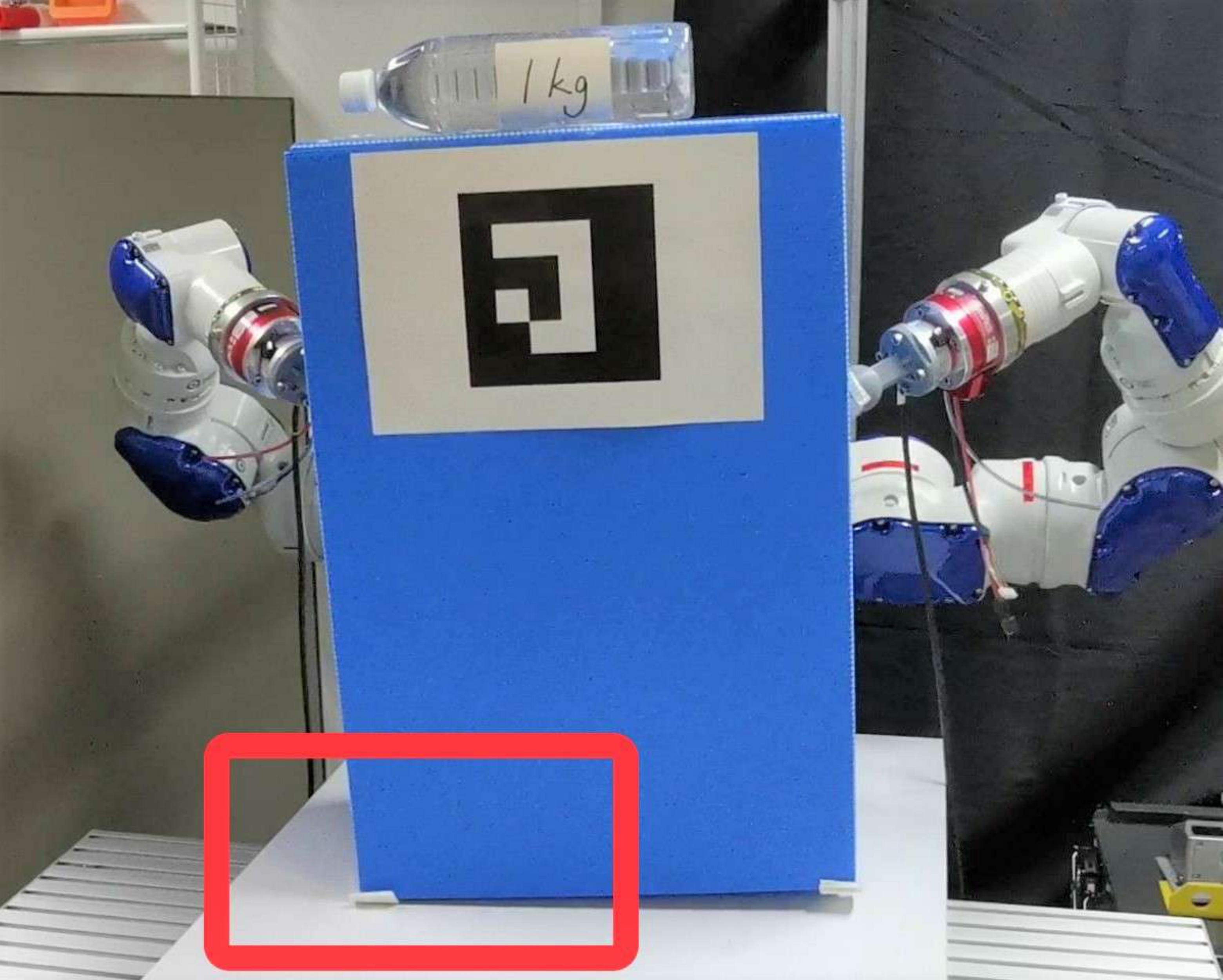}}
\caption{Experiment of placing a 1kg bottle on the manipulated object during the motion of pivoting, the robot fails to pivot the object. After the placement in Fig.\ref{1k3}, the object lands in the DS pose, see Fig.\ref{1k4}. When lifting the object, the motion fails and a collision between the table and the right front foot can be seen in the red rectangle in Fig.\ref{1k5}.}
\label{fig:bottle1kg}
\end{figure*}

\hl{We firstly test the pivoting gaits in the DS gait mode by putting a bottle on the manipulated object during motion. Bottles in weight of 0.35kg, 0.5kg and 1kg are placed,} see Fig.\ref{fig:bottle350g}, Fig.\ref{fig:bottle500g}, and Fig.\ref{fig:bottle1kg}, \hl{respectively.
In all three experiments, the robot starts to pivot the object in the DS mode,} see Fig.\ref{350g1}, Fig.\ref{500g1}, and Fig.\ref{ds1}. 
In Fig.\ref{350g3}, Fig.\ref{500g3}, and Fig.\ref{1k3}, \hl{the bottle in weight of 0.35kg, 0.5kg and 1kg is placed on the top of the box, respectively, during the walking motions which results in a change of the object mass.}
In Fig.\ref{350g4}-Fig.\ref{350g6} and Fig.\ref{500g4}-Fig.\ref{500g6}, \hl{the robot successfully pivots the object to walk after the placements of bottles in weight of 0.35kg and 0.5kg. 
In the case of placing the 1kg bottle, the object lands in a DS pose after the placement of bottle,} see Fig.\ref{1k4}. \hl{However, the robot fails to lift the object after switching the rotation vertex from the right to the left. What's more, unfortunately, a collision is found between the table and the font right foot of the object, see the red rectangle} in Fig.\ref{1k5}. 

\begin{figure*}
\centering     
\subfigure[]{\label{dt1}\includegraphics[width=0.22\textwidth]{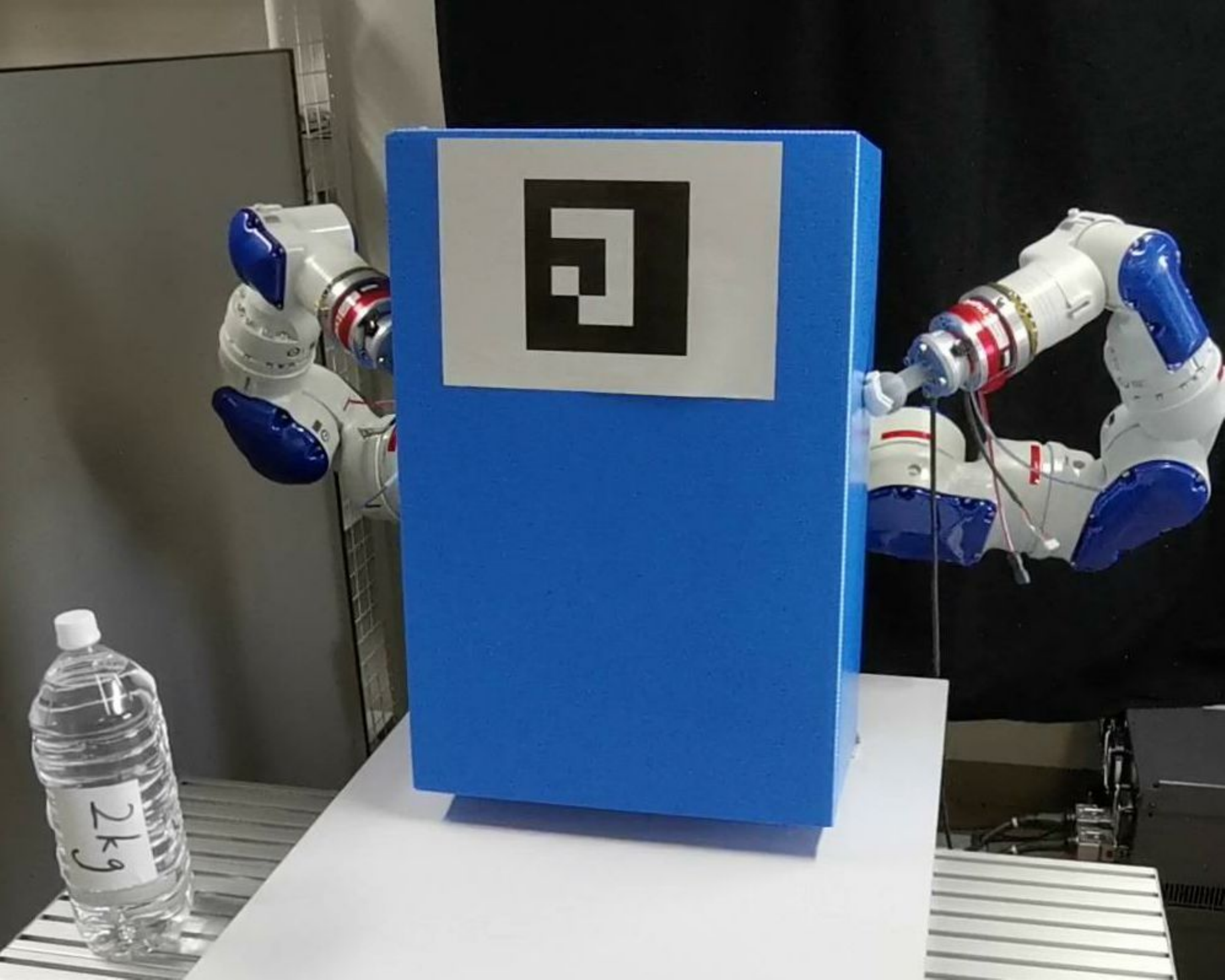}}
\subfigure[]{\label{dt2}\includegraphics[width=0.22\textwidth]{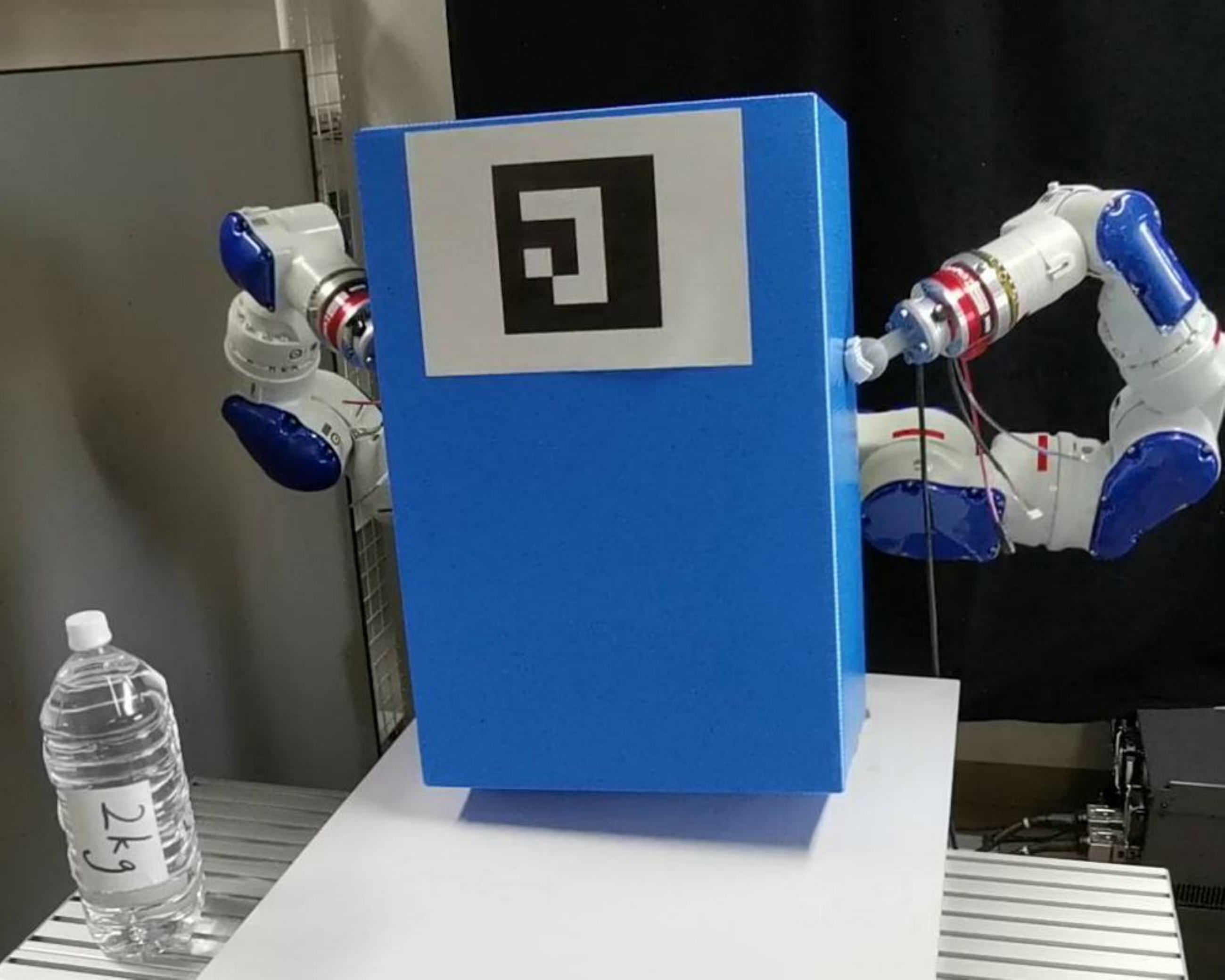}}
\subfigure[]{\label{dt3}\includegraphics[width=0.22\textwidth]{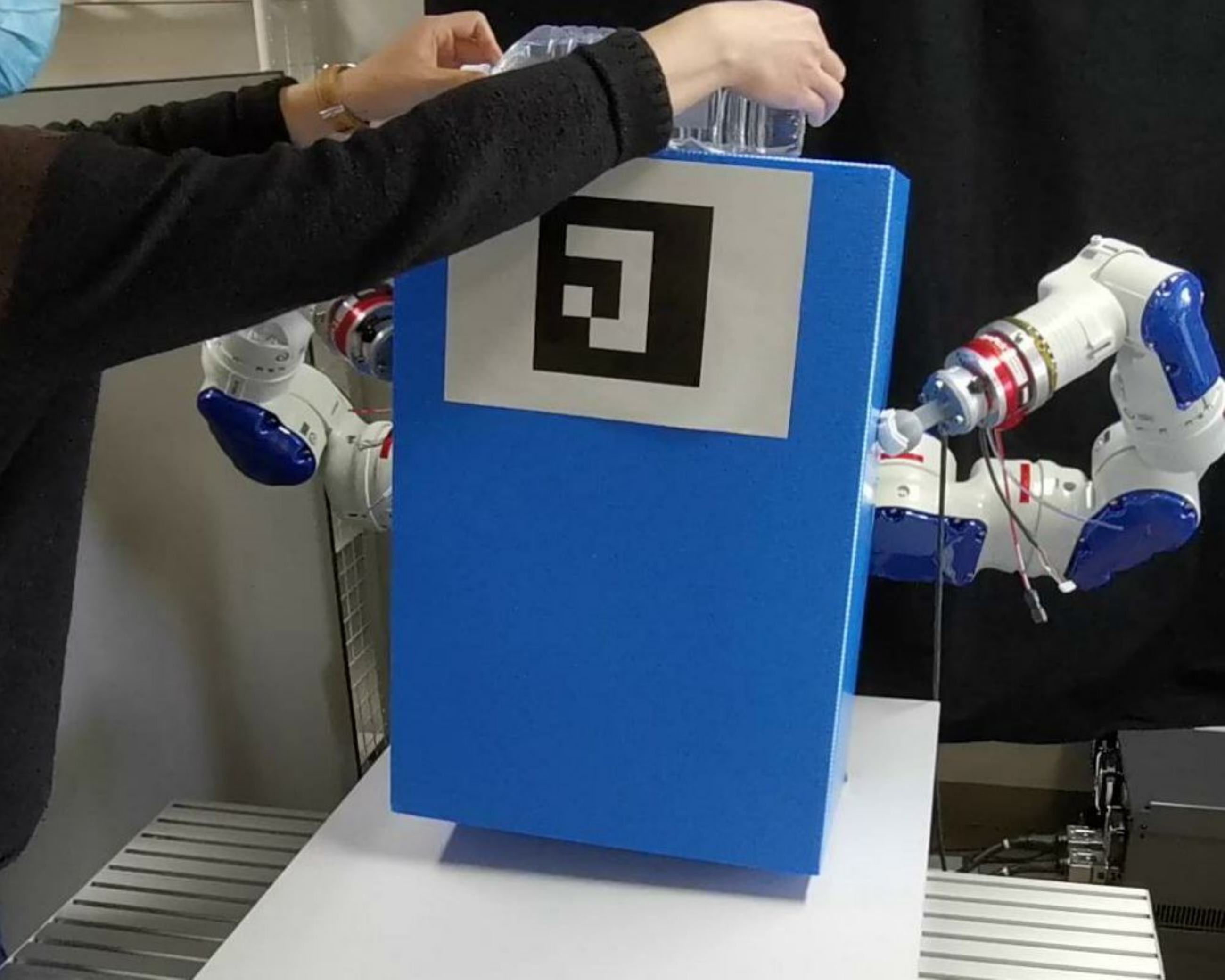}}
\subfigure[]{\label{dt4}\includegraphics[width=0.22\textwidth]{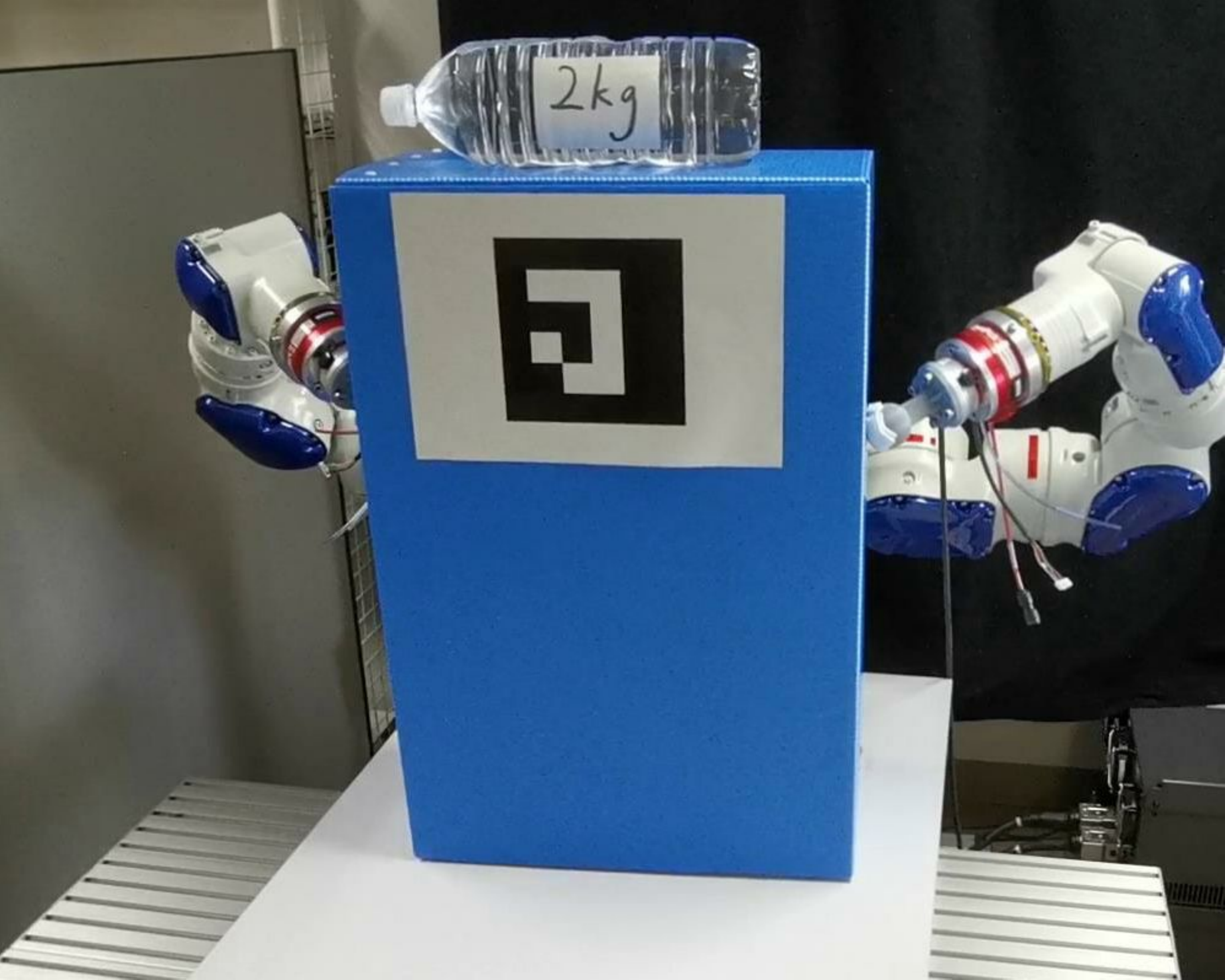}}
\subfigure[]{\label{dt5}\includegraphics[width=0.22\textwidth]{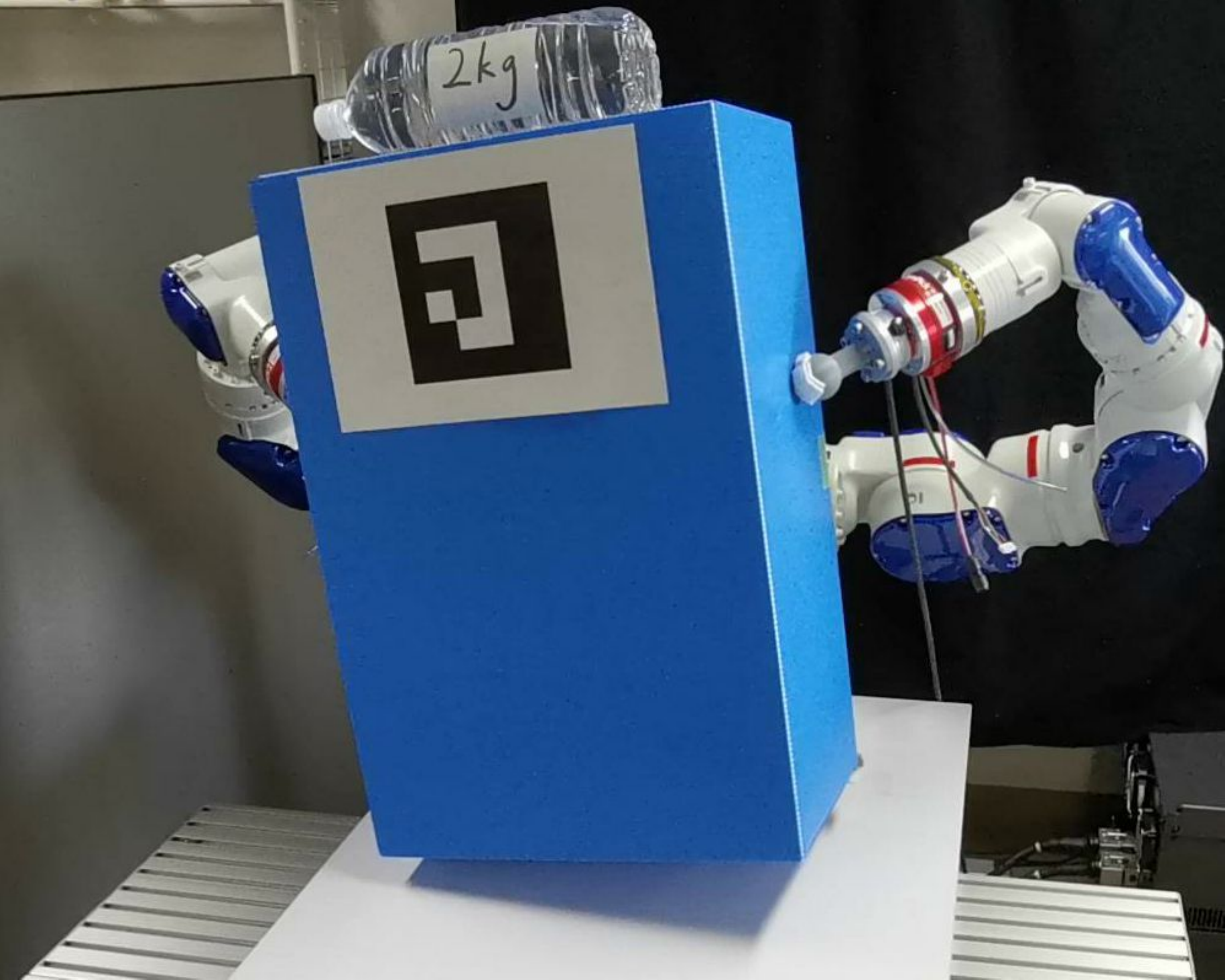}}
\subfigure[]{\label{dt6}\includegraphics[width=0.22\textwidth]{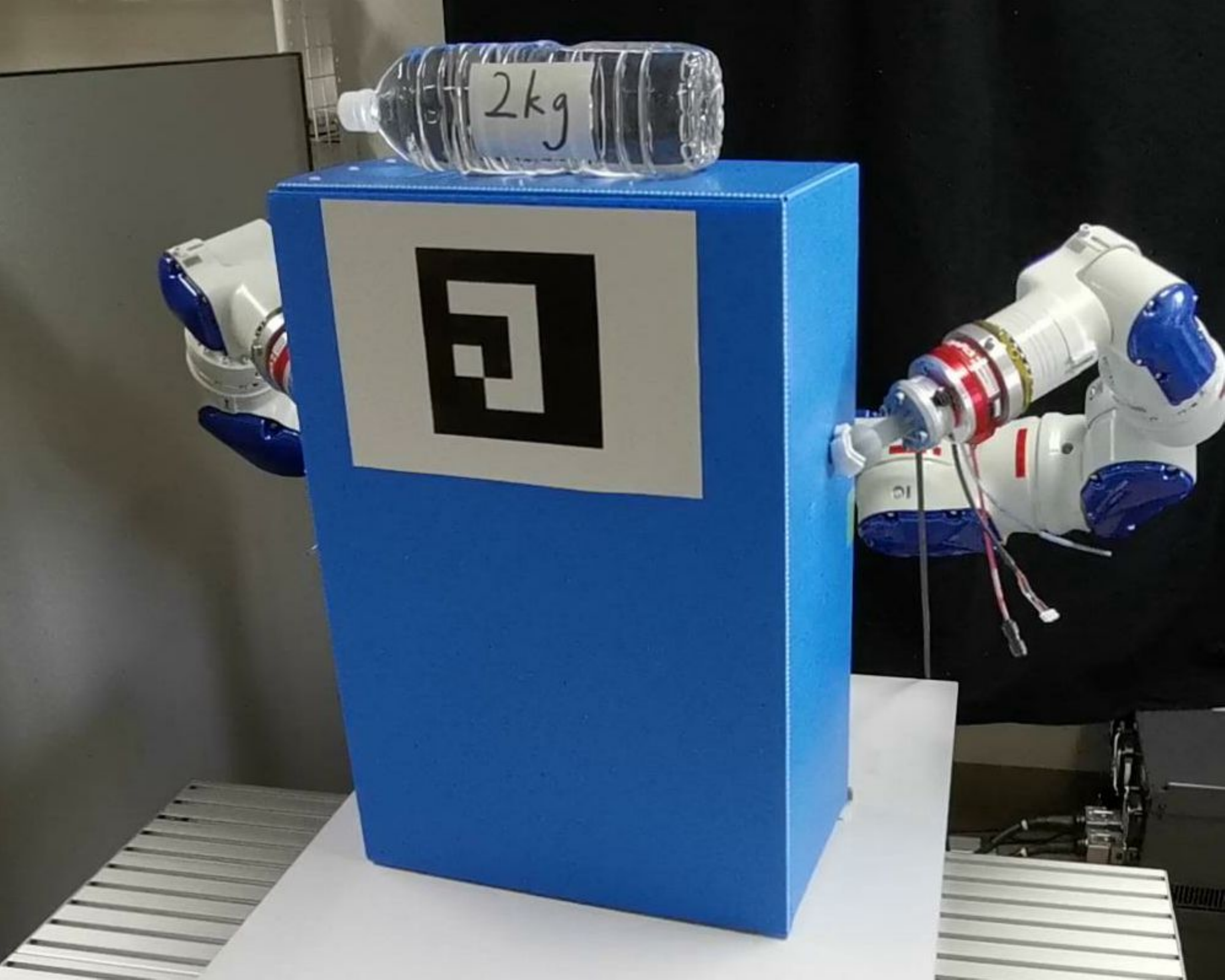}}
\subfigure[]{\label{dt7}\includegraphics[width=0.22\textwidth]{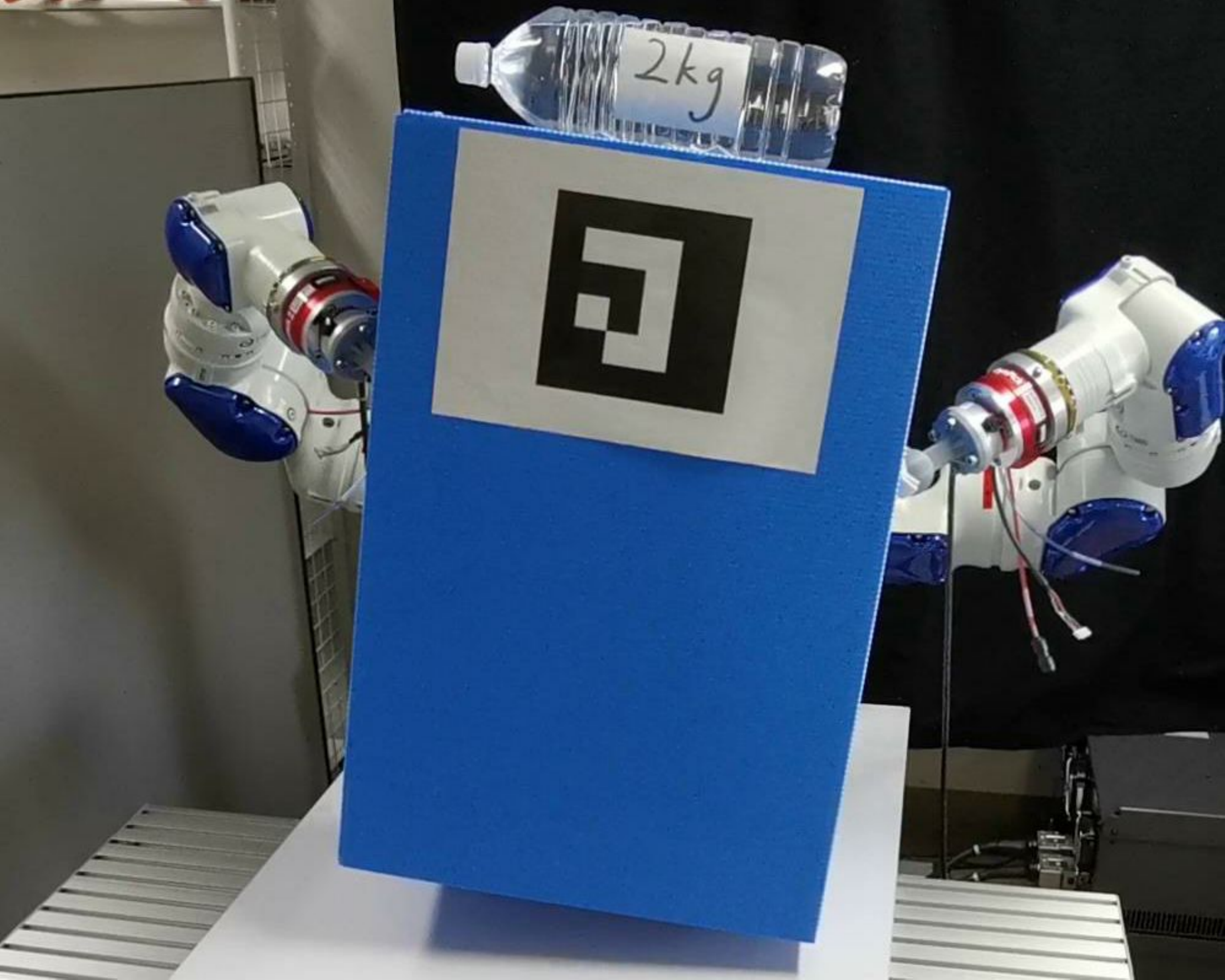}}
\subfigure[]{\label{dt8}\includegraphics[width=0.22\textwidth]{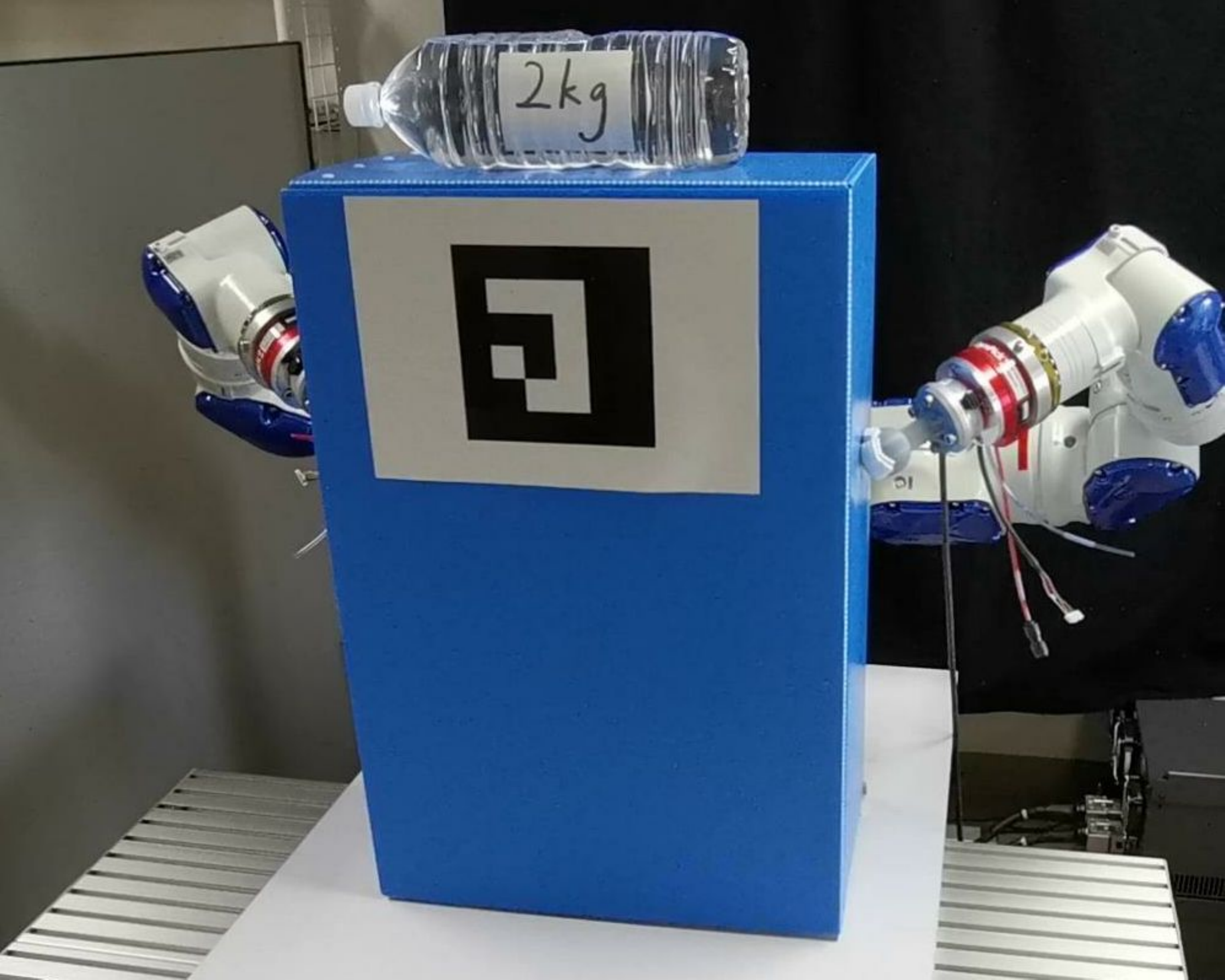}}
\caption{Experiment of placing a 2kg bottle on the manipulated object during the motion of pivoting. A 2kg bottle is put on the top of the box during walking, see Fig.\ref{dt3}. After the placement, the robot changes gait mode from DS to the QS mode(Fig.\ref{dt4}) and pivots the box to walk. }
\label{fig:bottle}
\end{figure*}

\hl{Since the failure arises during the switching of rotation vertices, we improve pivoting gait by realizing the switching gait mode to the QS mode where all four vertices of the object are on the table which leads to a firm contact between the object and the table.
With the help of graph MPC, we can change the gait mode from the DS to the QS gait mode according to the environment. 
In the experiment shown} in Fig.\ref{fig:bottle}, the robot firstly manipulates the object in the DS mode, see Fig.\ref{dt1} and Fig.\ref{dt2}. A 2kg bottle is placed on the top of the box, see Fig.\ref{dt3}. \hl{The placement is detected by the force sensors in the wrist of the robot and the graph switch the gait mode from the DS mode to the QS mode. After the placement, the object lands in QS pose in} Fig.\ref{dt4}. \hl{Then the robot pivots the object to walk for two more steps in QS mode and successfully finishes the pivoting gait, see} Fig.\ref{dt5}-\ref{dt8}. 
This experiment shows that 1. The QS mode is more stable than the DS mode when disturbance occurs. 2. Switching of gait mode improves the robustness of the control system.

\begin{table}
\centering
\caption{Placements of bottles in different weights}
\label{table2}
\begin{tabular}{|p{35pt}|p{35pt}|p{35pt}|p{40pt}|}
\hline
Weight of bottle(kg) & Gait Mode & Switching of Mode & Result \\
\hline
0.35 & DS   & No & Succeed \\
0.5  & DS   & No & Succeed \\
1.0    & DS   & No & Fail \\
2.0    & DS + QS & Yes & Succeed \\
\hline
\multicolumn{4}{p{200pt}}{\hl{Robot successfully pivots the object in the DS mode after the placements of 0.35kg and 0.5 kg bottles, respectively but fails to pivot the object with a 1 kg bottle. A switching of gait mode from the DS to the QS mode enables the robot to pivot the object after the placement of a 2kg bottle.}}\\
\end{tabular}
\end{table}

Table. \ref{table2} \hl{shows the results of experiments in this subsection. The pivoting gait in the DS mode can against little disturbances, for example, disturbances from the placement of bottles in weight of 0.35kg and 0.5kg. The pivoting gait in the DS mode fails to carry a 1kg bottle. However, after implementing graph MPC which enables the switching gait mode, the robot successfully pivots the object after the placement of a 2kg bottle by switching from the DS mode to the QS mode.}   
\hl{The experiments show that the QS mode is more stable than the DS mode and the ability to switch gait modes improves the robustness of the control system.}

\begin{figure}
\centering
\includegraphics[width=0.45\textwidth]{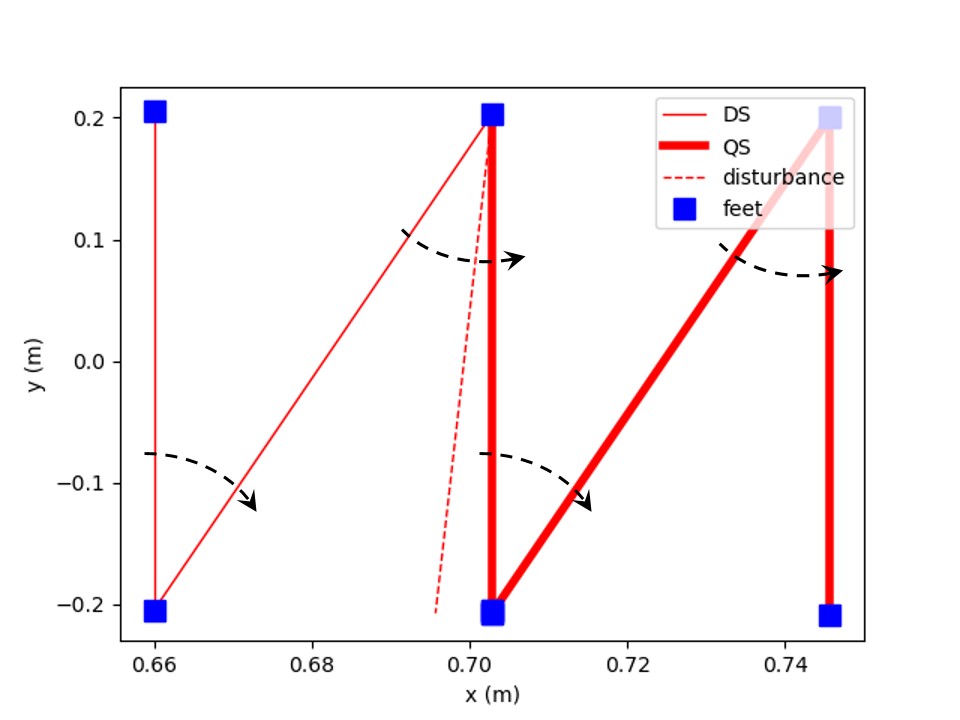}
\caption{Footprints in the experiment of placing a 2kg bottle. The placement of the bottle is detected and shown by the red dash line. After detecting the disturbance, gait mode transferred from the DS mode to the QS mode.}
\label{fig:bottlefoot}
\end{figure}

\subsection{Experiment 3: External perturbation}
\hl{In pivoting gait, if the robot continues manipulation without feedback, the error of the position of the object will accumulate. As a solution to these problem, we use visual system to watch the state of the object. 
When a perturbation causes a relative motion between the EEFs and the object, the combination of visual system and graph MPC is useful to compare the tracking data with the desired state of the object. Then the robot modifies its motion and recovers from the perturbation. }

\begin{figure}
\centering
\includegraphics[width=0.45\textwidth]{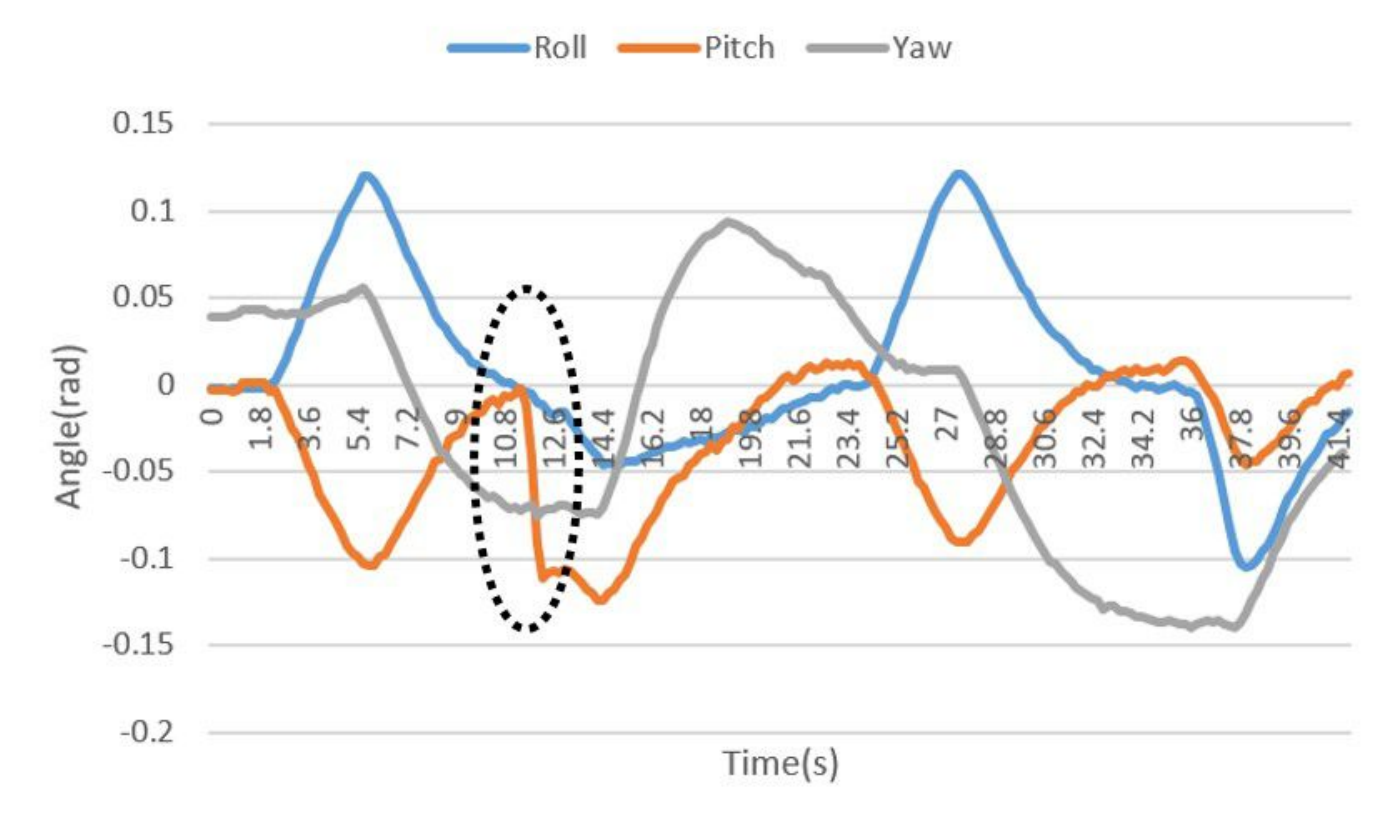}
\caption{Euler angles of the box in experiment 3. The black dashed circle shows a fluctuation in pitch angle caused by pushing and it is detected at time 11.8s. After the disturbance, the controller firstly decreases the roll and pitch angles which makes the box be in SS pose to avoid scuffing during walking. Later, the controller tunes roll and pitch angles to zeros which indicates a landing pose at time 24.2s, and then pivots the box to walk. }
\label{fig:rpyCam}
\end{figure}

\begin{figure*}
\centering     
\subfigure[Cameras in front of box]{\label{fig:1}\includegraphics[width=0.2\textwidth]{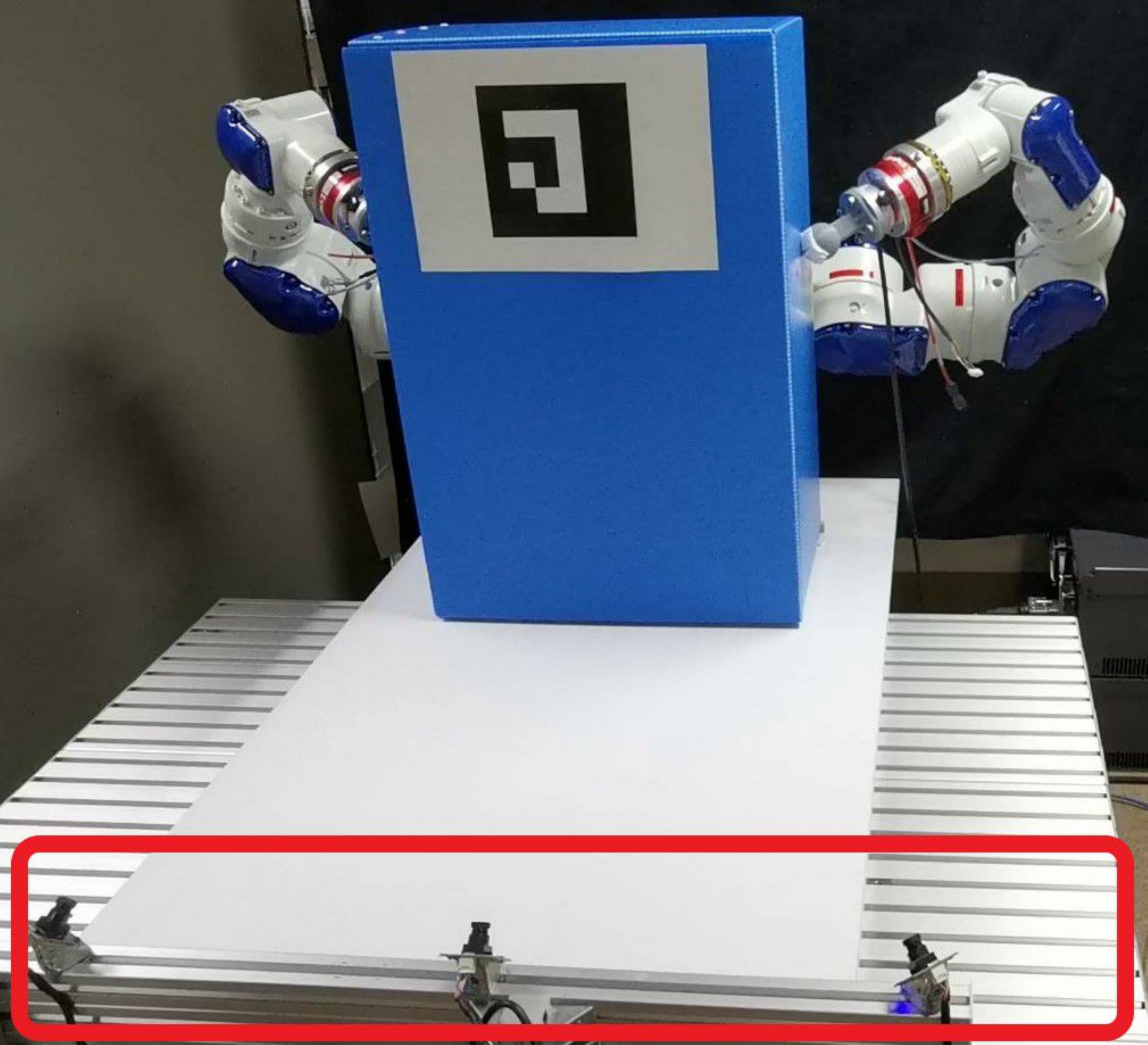}}
\subfigure[]{\label{fig:2}\includegraphics[width=0.22\textwidth]{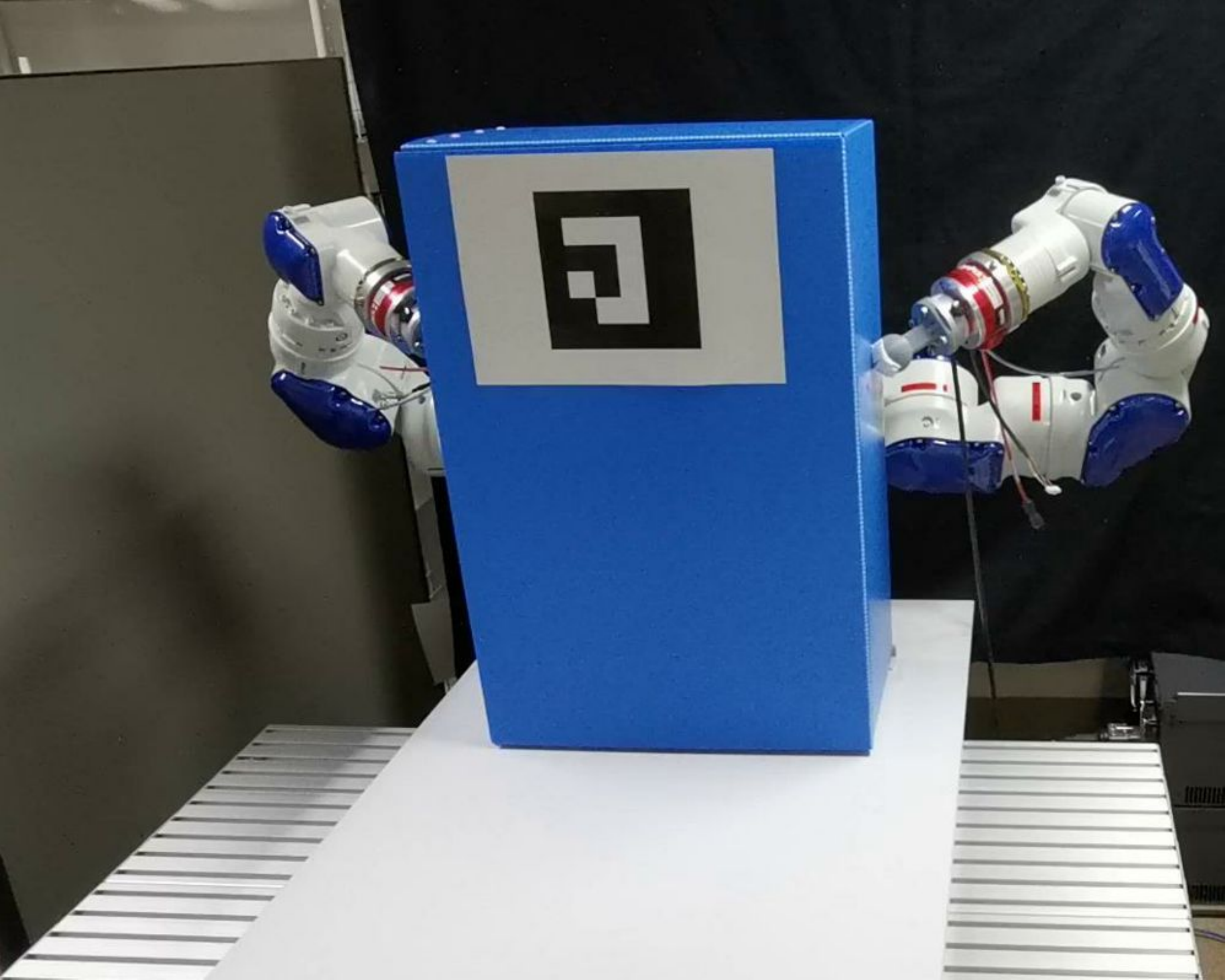}}
\subfigure[]{\label{fig:3}\includegraphics[width=0.22\textwidth]{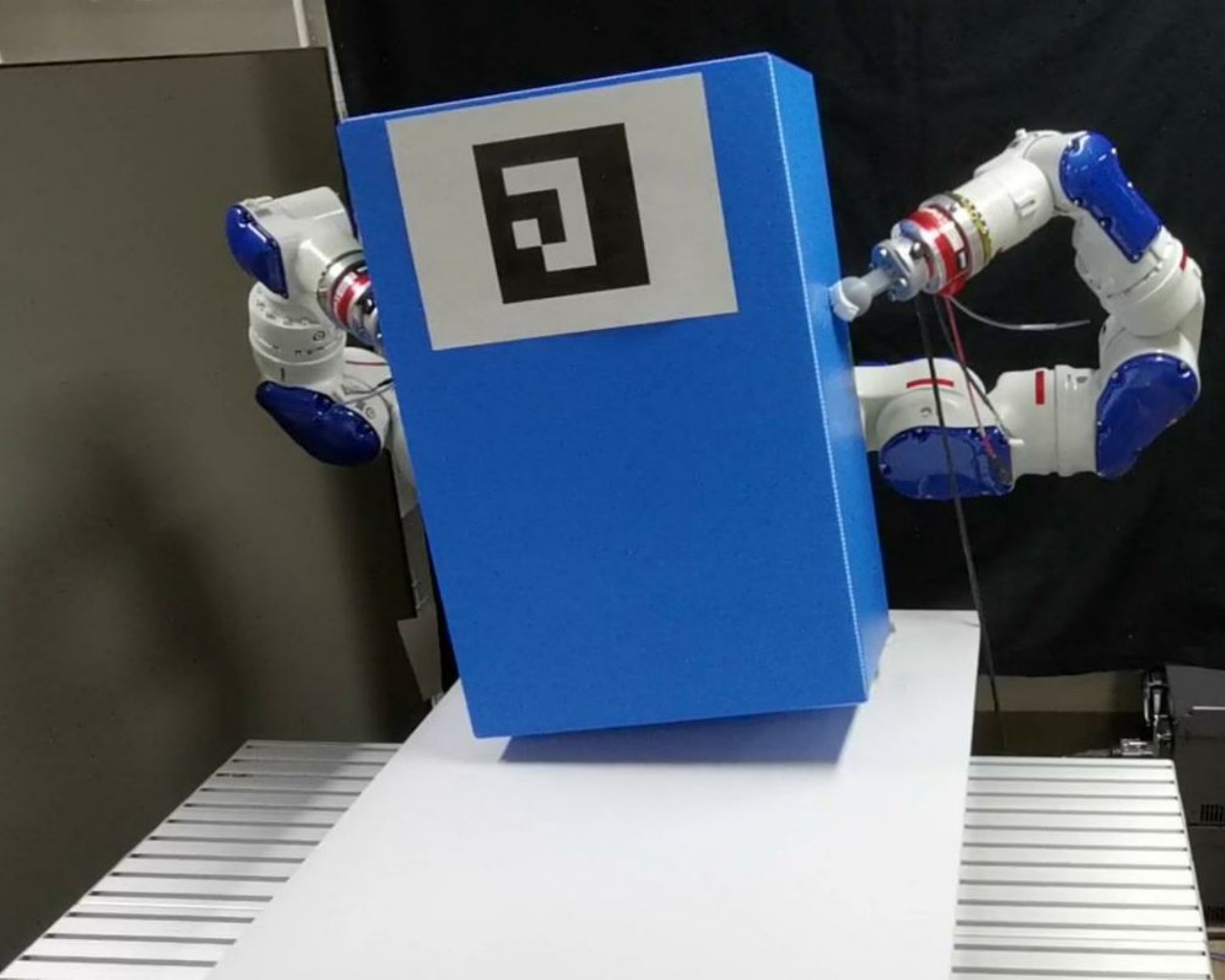}}
\subfigure[]{\label{fig:4}\includegraphics[width=0.22\textwidth]{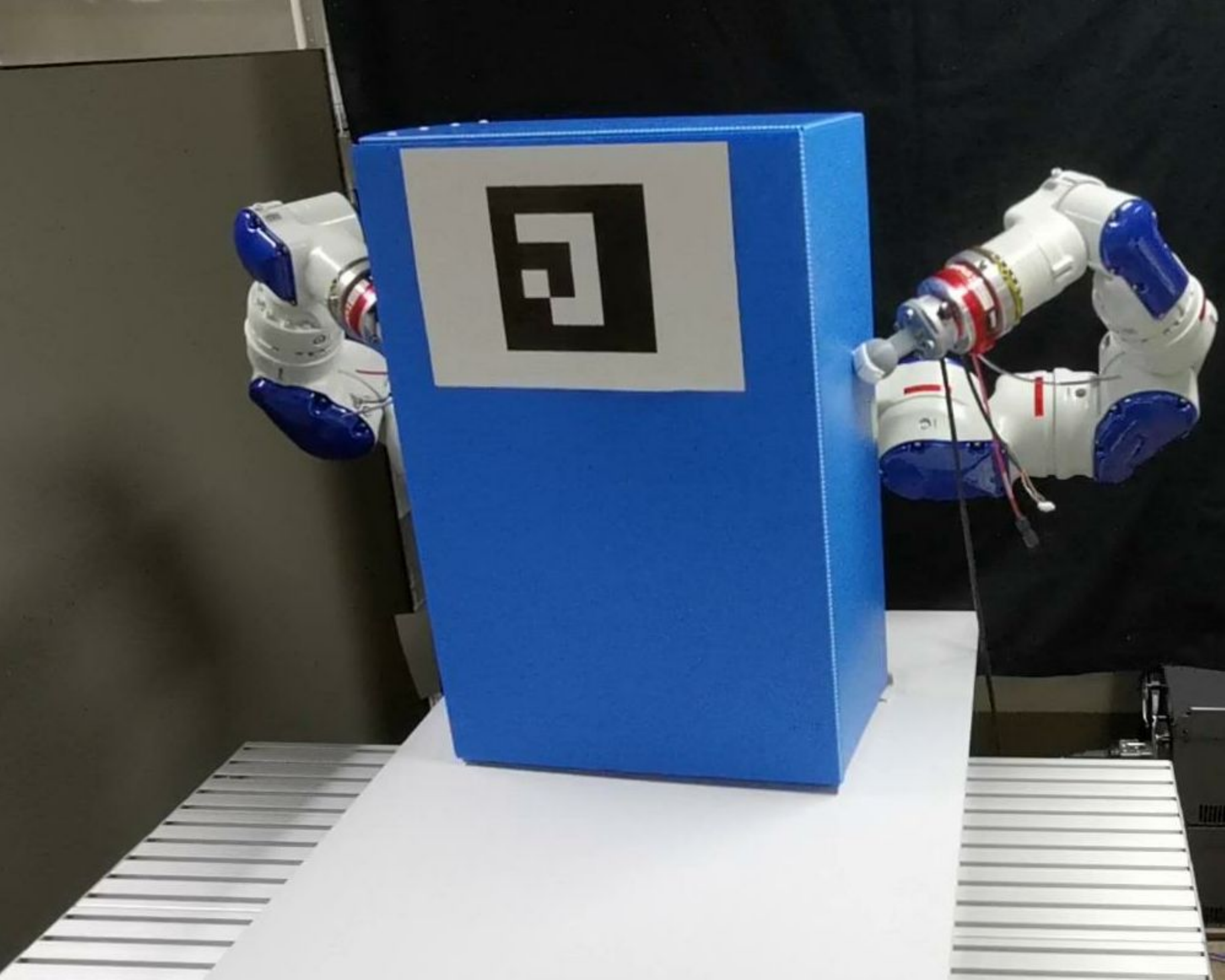}}
\subfigure[]{\label{fig:5}\includegraphics[width=0.22\textwidth]{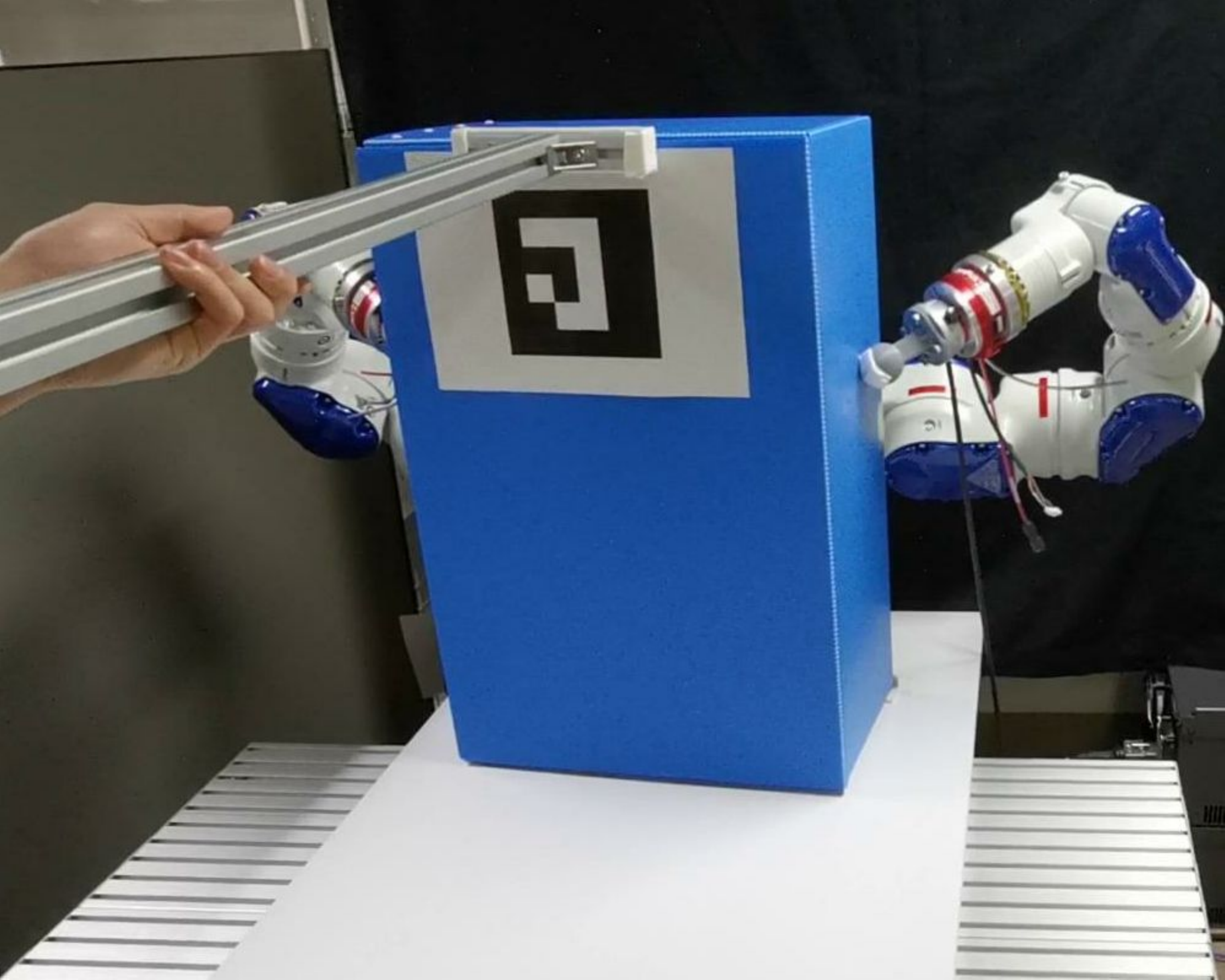}}
\subfigure[]{\label{fig:6}\includegraphics[width=0.22\textwidth]{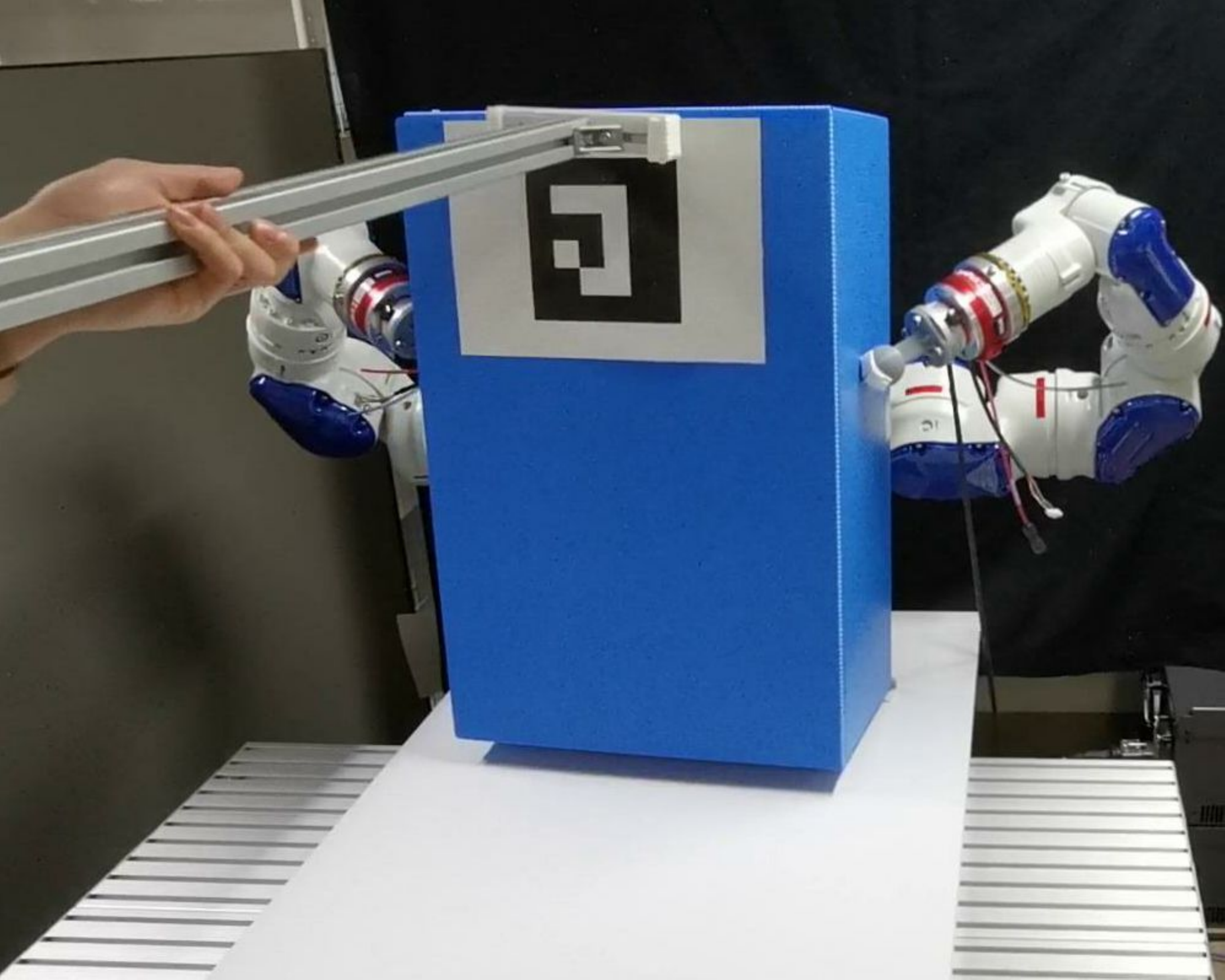}}
\subfigure[]{\label{fig:7}\includegraphics[width=0.22\textwidth]{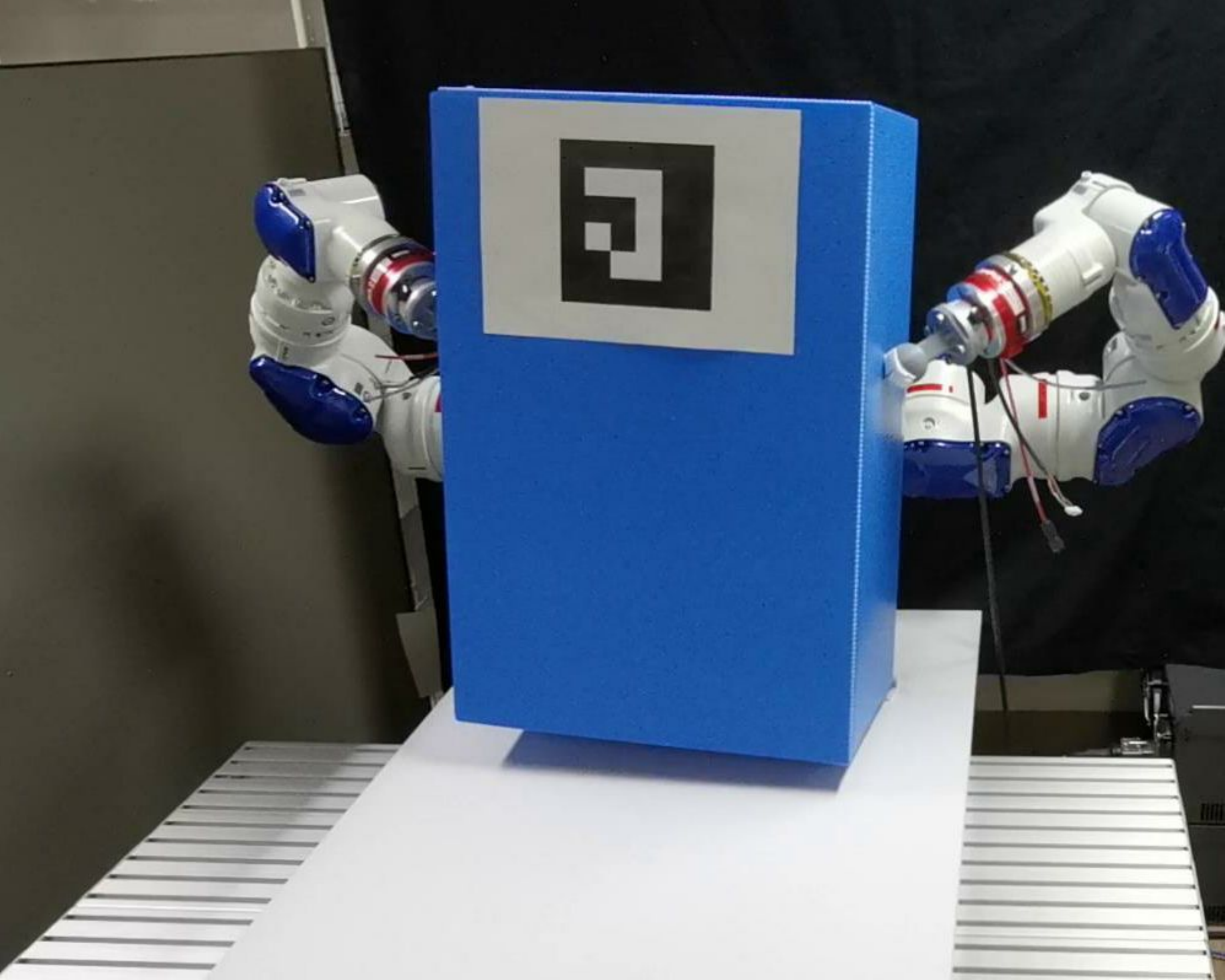}}
\subfigure[]{\label{fig:8}\includegraphics[width=0.22\textwidth]{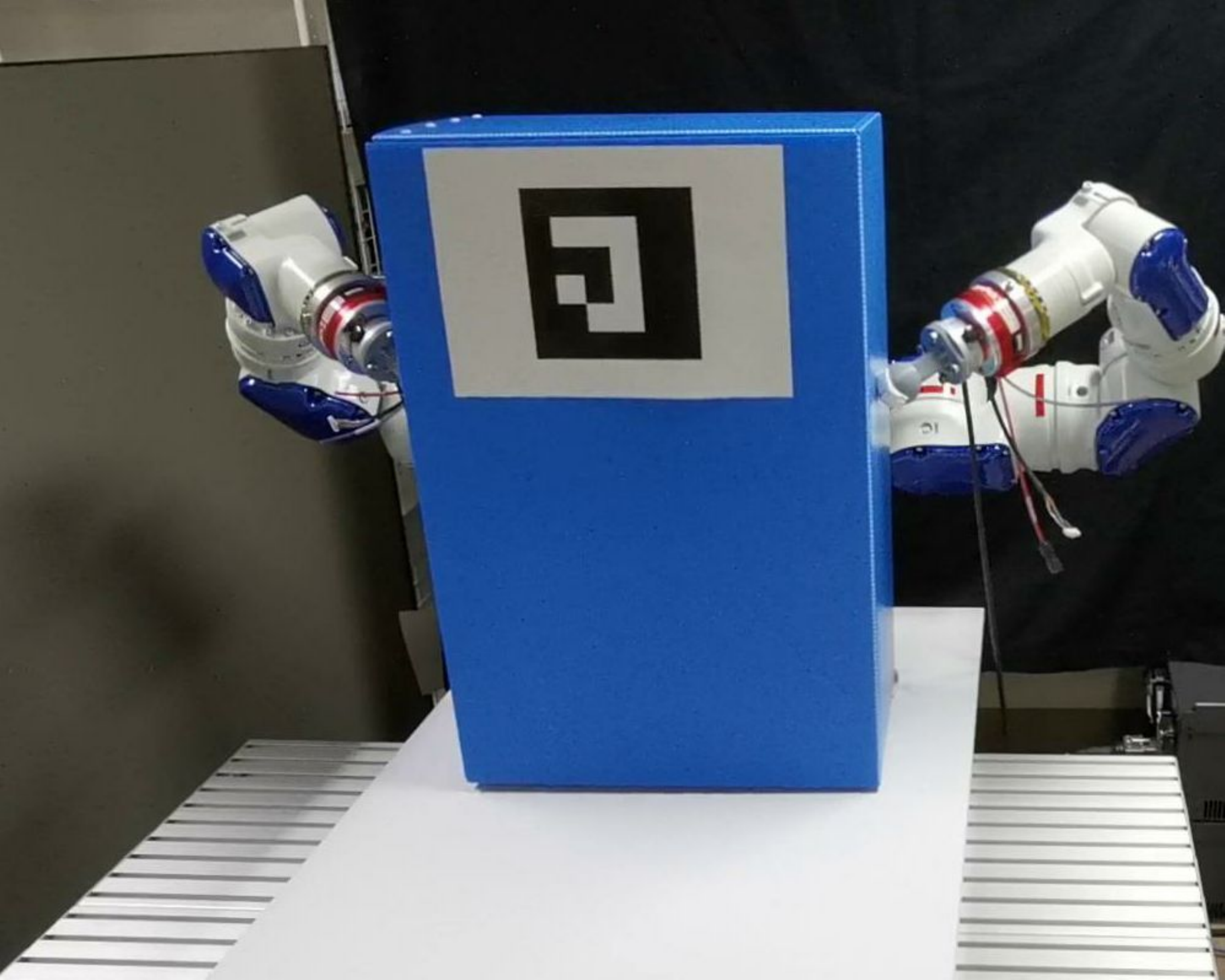}}
\subfigure[]{\label{fig:9}\includegraphics[width=0.22\textwidth]{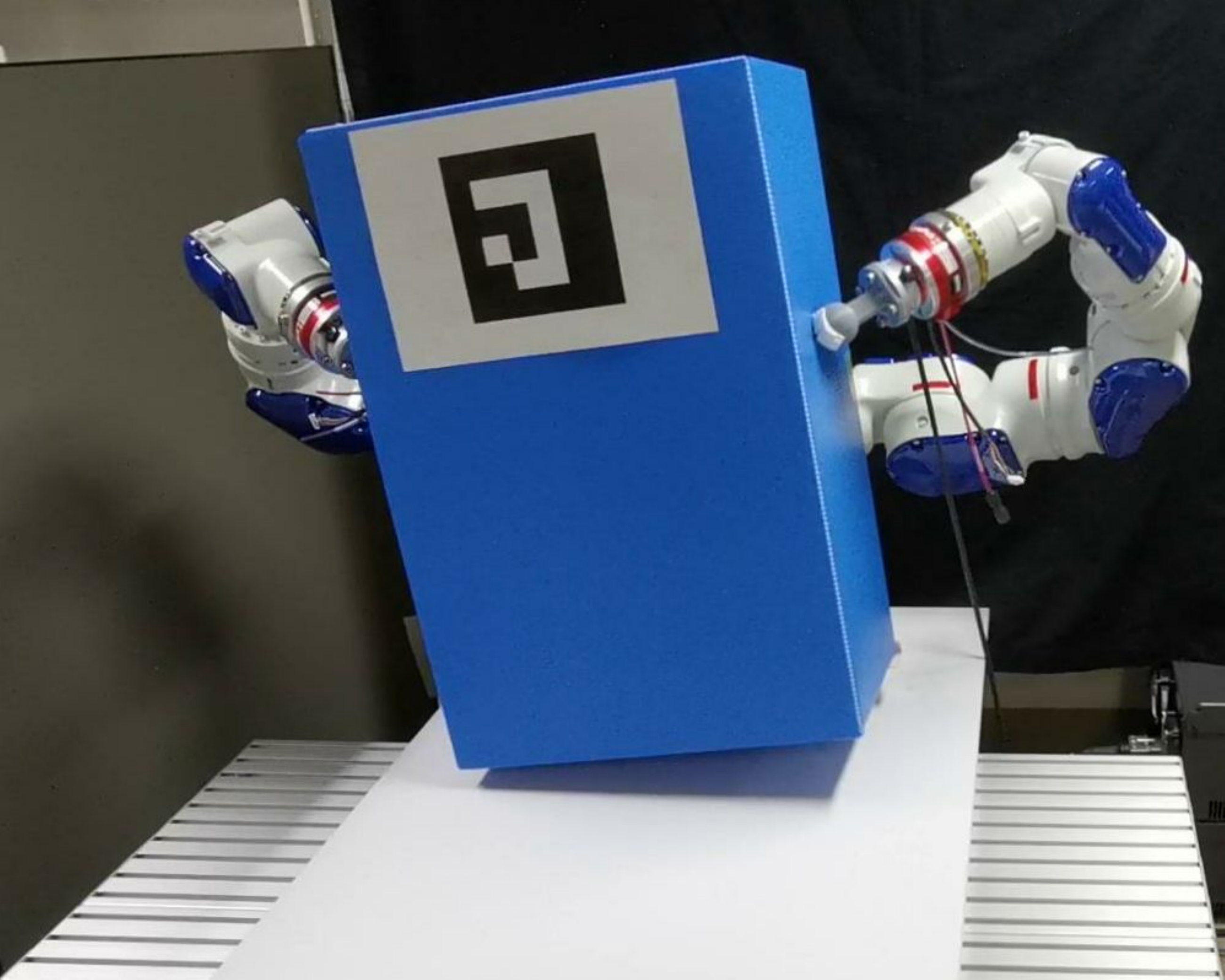}}
\subfigure[]{\label{fig:10}\includegraphics[width=0.22\textwidth]{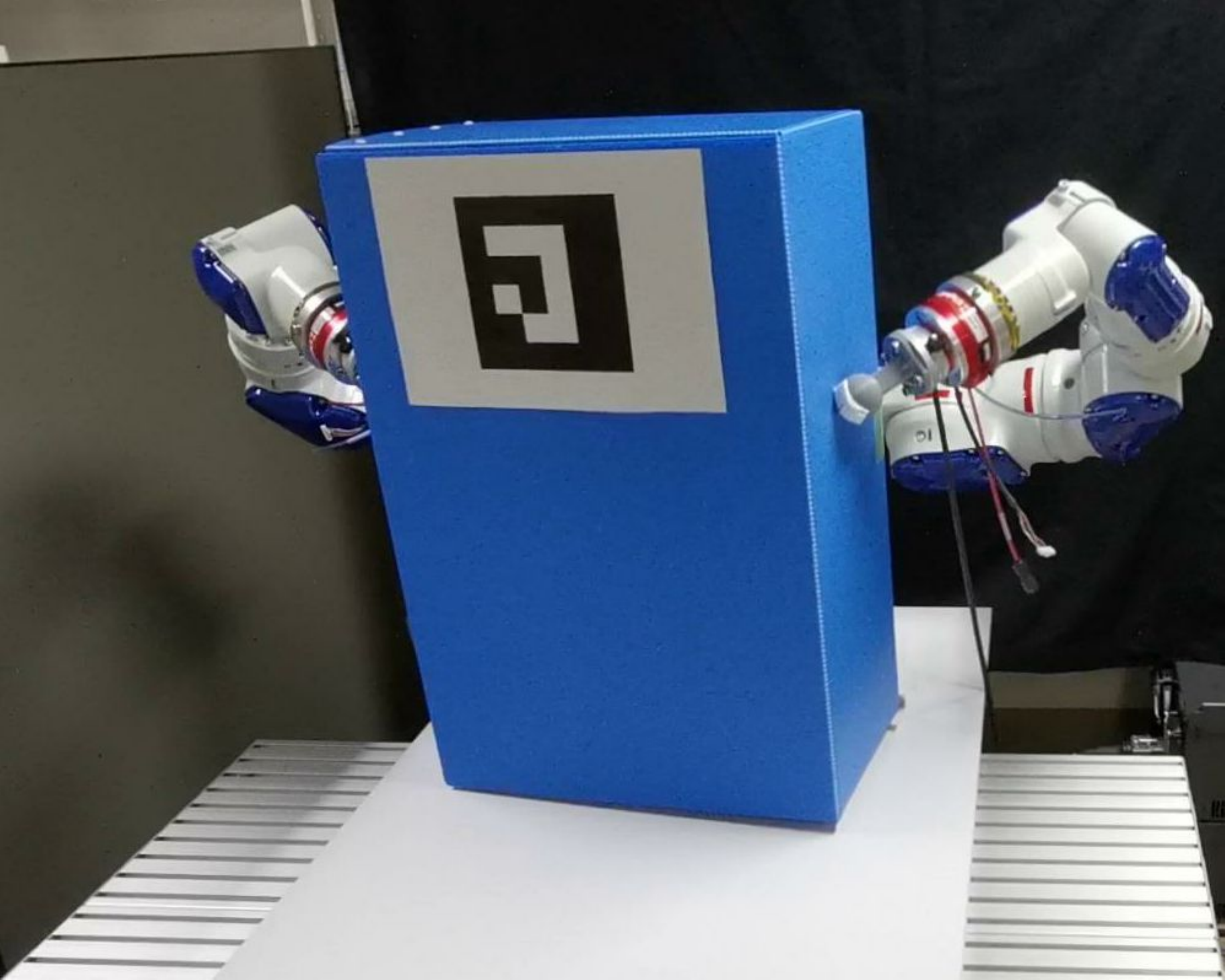}}
\subfigure[]{\label{fig:11}\includegraphics[width=0.22\textwidth]{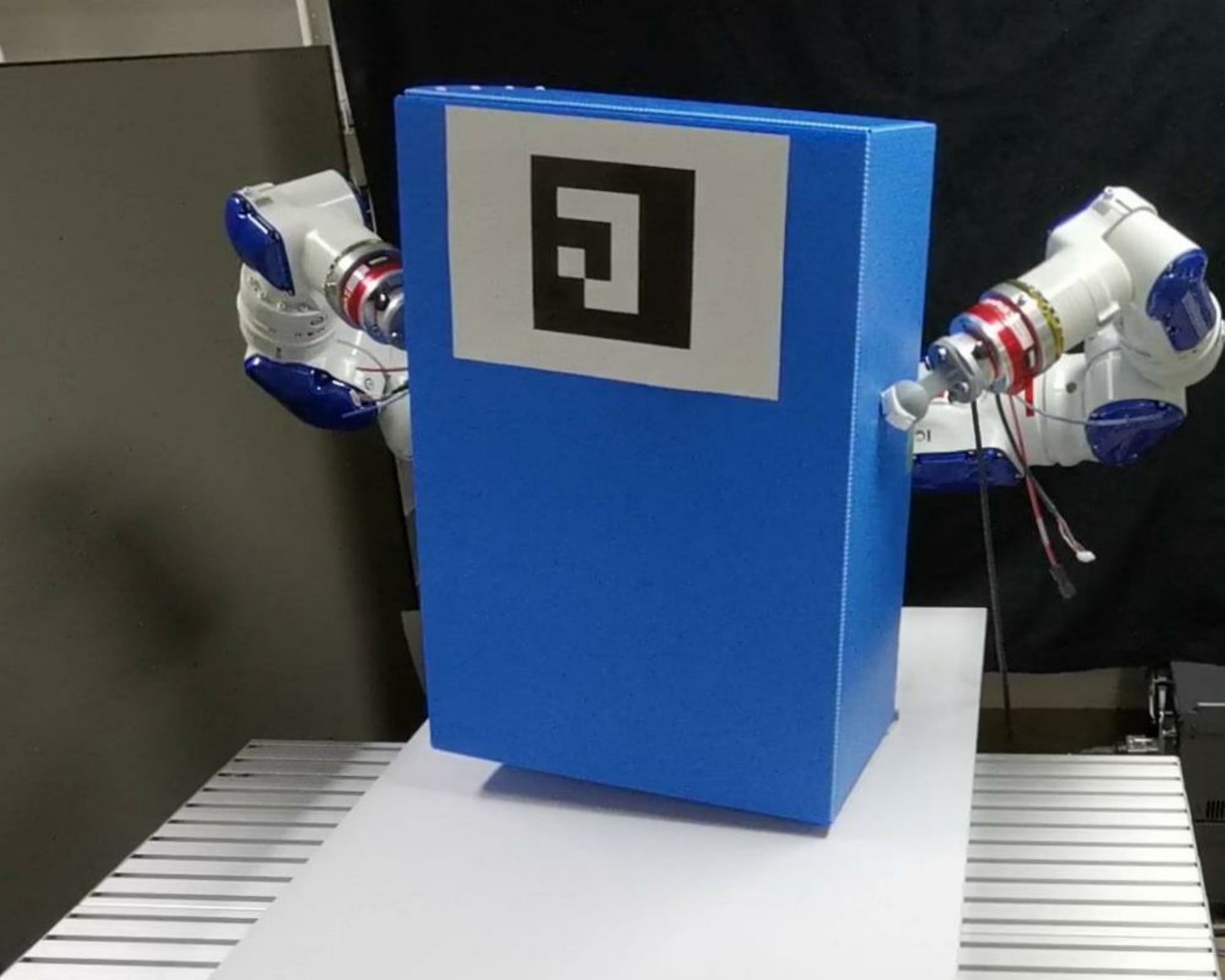}}
\subfigure[]{\label{fig:12}\includegraphics[width=0.22\textwidth]{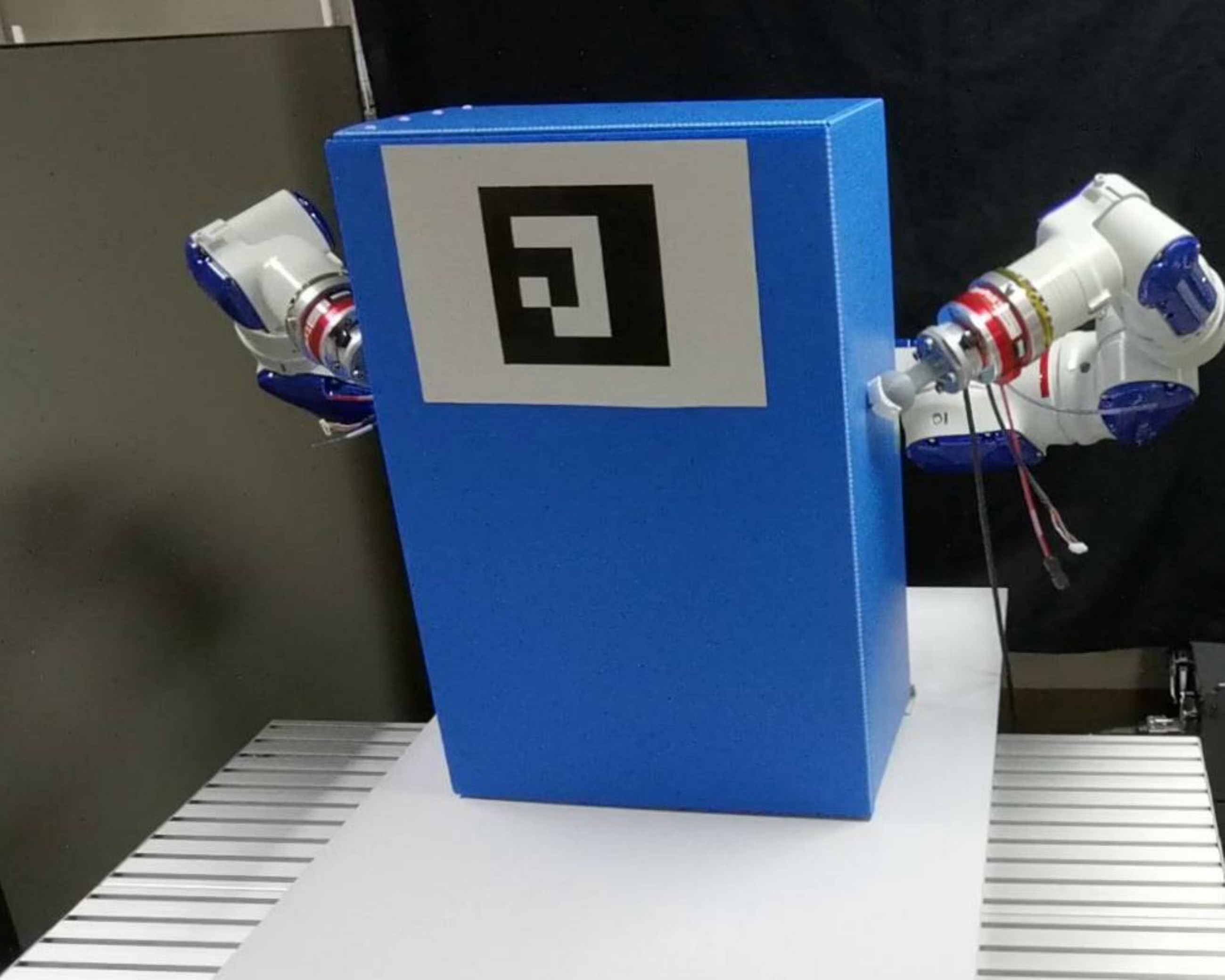}}
\caption{Cameras are used to track the motion of the box, see the red rectangle in Fig.\ref{fig:1}. An unexpected push is acted to the box at Fig.\ref{fig:5}. The robot pivots the object to walk and makes reactions to resist disturbance.}
\label{fig:cam}
\end{figure*}

Three cameras are put in front of the table, which are shown in the red rectangle in Fig.\ref{fig:1}. The cameras detect the Euler angles and the position of the marker on the box. In the experiment, the robot starts to pivot the object to walk in the QS mode, see Fig.\ref{fig:2}-\ref{fig:4}. An unexpected push is acted to the box during walking by a metal stick, see Fig.\ref{fig:5}. The states of the object are suddenly changed and captured by the vision system, see Fig.\ref{fig:rpyCam}. After the disturbance, a new motion is generated and QS mode is selected as gait mode because it provides stable walking. In the new motion, the robot firstly lifts the box into a SS pose which is to avoid the box from scuffing the table, see Fig.\ref{fig:rpyCam} from time 12.4-14.4s and Fig.\ref{fig:7}. Then, the robot lands the box into QS pose to firmly contact with the table, see Fig.\ref{fig:rpyCam} at time 24.2s and Fig.\ref{fig:8}. After landing, the robot pivots the object to walk for two more steps, Fig.\ref{fig:9}-\ref{fig:12}. 

\hl{The experiment shows that with the help of the vision system, the proposed control system is able to recover from the perturbation and track the reference trajectory.} 

\begin{figure}
\centering
\includegraphics[width=0.45\textwidth]{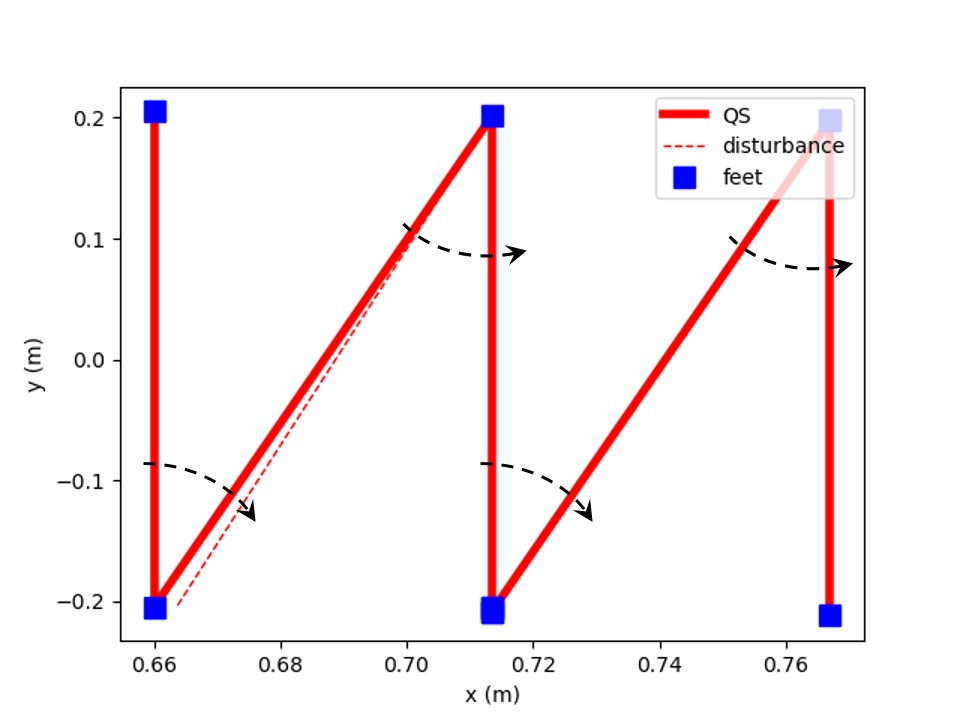}
\caption{Footprints of experiment 3. A perturbation is detected by the vision system and shown by the red dash line which is close to the first step. Though perturbation occurs, the MPC still finds a solution to follow the planned trajectory.}
\label{fig:camfoot}
\end{figure}

\section{Conclusion and Future Work}
\label{sec:6conclusion}
In this article, we use a dual-arm robot to pivot the object to walk. Two gait modes are designed for adapting to an environment where the DS mode for fast walking and the QS mode for stable walking. Vision systems and force sensors are implemented to perceive the environment. A graph MPC is proposed where the graph selects the gait mode based on the information collected from both force sensors and the vision system and outputs a reference trajectory which is tracked by the MPC. 

\hl{Experiments show that the gait mode influences the robustness of the system, especially when disturbance occurs. The DS gait mode is fast while the QS mode is stable way because it provides a form contact between the object and the table. The proposed graph MPC can select gait mode adaptively and control the pivoting gait in real-time while being free to switch gait modes, robust against external perturbations and uncertainty in the object’s weight.}

The extension is to apply our pivoting approach to objects in different shapes. It requires identifying appropriate model, selection of supporting vertices, design of grasp positions, and design of new gait modes.   

\bibliographystyle{unsrt}
\bibliography{reference.bib}

\end{document}